\documentclass[10pt,journal,compsoc]{IEEEtran}

\ifCLASSOPTIONcompsoc
  \usepackage[nocompress]{cite}
\else
  \usepackage{cite}
\fi
\ifCLASSINFOpdf
\else
\fi
\pdfoutput=1

\usepackage{url}
\usepackage{times}
\usepackage{epsfig}
\usepackage{graphicx}
\usepackage{amsmath}
\usepackage{amssymb}
\usepackage{multirow}
\usepackage{colortbl}
\usepackage{color}
\usepackage{threeparttable}
\usepackage{booktabs}
\usepackage{adjustbox}
\usepackage{subfig}
\usepackage{caption}
\usepackage[misc]{ifsym}
\usepackage{xcolor,colortbl}
\usepackage{makecell}
\newcommand*{\belowrulesepcolor}[1]{%
  \noalign{%
    \kern-\belowrulesep 
    \begingroup 
      \color{#1}%
      \hrule height\belowrulesep 
    \endgroup 
    \vspace{-0.03mm}
  }%
} 
\newcommand*{\aboverulesepcolor}[1]{%
  \noalign{%
  \vspace{-0.03mm}
    \begingroup 
      \color{#1}%
      \hrule height\aboverulesep 
    \endgroup 
    \kern-\aboverulesep 
  }%
}

\usepackage{pifont}

\usepackage{xspace}
\makeatletter
\DeclareRobustCommand\onedot{\futurelet\@let@token\@onedot}
\def\@onedot{\ifx\@let@token.\else.\null\fi\xspace}
\def\eg{\emph{e.g}\onedot} 
\def\ie{\emph{i.e}\onedot} 
 
\def\etc{\emph{etc}\onedot} 
 
\def\etal{\emph{et al}\onedot}
\def\aka{\emph{a.k.a}\onedot} 

\usepackage[linesnumbered,lined,boxed,commentsnumbered,ruled]{algorithm2e}
\usepackage[pagebackref,breaklinks=true,colorlinks,citecolor=blue,urlcolor=blue,linkcolor=blue,bookmarks=false]{hyperref}

\usepackage[edges]{forest}

\usepackage[edges]{forest}
\newsavebox{\measurebox}
\definecolor{paired-light-blue}{RGB}{198, 219, 239}
\definecolor{paired-dark-blue}{RGB}{49, 130, 188}
\definecolor{paired-light-orange}{RGB}{251, 208, 162}
\definecolor{paired-dark-orange}{RGB}{230, 85, 12}
\definecolor{paired-light-green}{RGB}{199, 233, 193}
\definecolor{paired-dark-green}{RGB}{49, 163, 83}
\definecolor{paired-light-purple}{RGB}{218, 218, 235}
\definecolor{paired-dark-purple}{RGB}{117, 107, 176}
\definecolor{paired-light-gray}{RGB}{217, 217, 217}
\definecolor{paired-dark-gray}{RGB}{99, 99, 99}
\definecolor{paired-light-pink}{RGB}{222, 158, 214}
\definecolor{paired-dark-pink}{RGB}{123, 65, 115}
\definecolor{paired-light-red}{RGB}{231, 150, 156}
\definecolor{paired-dark-red}{RGB}{131, 60, 56}
\definecolor{paired-light-yellow}{RGB}{231, 204, 100}
\definecolor{paired-dark-yellow}{RGB}{141, 209, 49}
\tikzset{%
    parent/.style =          {align=center,text width=1.2cm,rounded corners=3pt, line width=0.3mm, fill=pink!10,draw=pink!80},
    child/.style =           {align=center,text width=2.3cm,rounded corners=3pt, fill=blue!10,draw=blue!80,line width=0.3mm},
    grandchild/.style =      {align=center,text width=2cm,rounded corners=3pt},
    greatgrandchild/.style = {align=center,text width=1.5cm,rounded corners=3pt},
    greatgrandchild2/.style = {align=center,text width=1.5cm,rounded corners=3pt},    
    referenceblock/.style =  {align=center,text width=1.5cm,rounded corners=2pt},
    data/.style =           {align=center,text width=2cm,rounded corners=3pt, fill=paired-light-blue!50,draw=paired-dark-blue!65,line width=0.3mm},
    data_wide/.style =           {align=left,text width=4.45cm,rounded corners=3pt, fill=paired-light-blue!50,draw=paired-dark-blue!65,line width=0.3mm},   
    data_work/.style =           {align=center, text width=4.5cm,rounded corners=3pt, fill=paired-light-blue!50,draw=blue!0,line width=0.3mm},  
    model/.style =           {align=center,text width=2cm,rounded corners=3pt, fill=paired-light-orange!50,draw=paired-dark-orange!65,line width=0.3mm},  
    model_wide/.style =           {align=center,text width=2.5cm,rounded corners=3pt, fill= paired-light-orange!50,draw=paired-dark-orange!65,line width=0.3mm}, 
    model_more/.style =           {align=center,text width=4cm,rounded corners=3pt, fill=paired-light-orange!50,draw=paired-dark-orange!65,line width=0.3mm},
    model_more_left/.style =      {align=left,text width=4cm,rounded corners=3pt, fill=paired-light-orange!50,draw=paired-dark-orange!65,line width=0.3mm},
    model_large_left/.style =      {align=left,text width=6.45cm,rounded corners=3pt, fill=paired-light-orange!50,draw=paired-dark-orange!65,line width=0.3mm},   
    model_work/.style =           {align=center,text width=4.5cm,rounded corners=3pt, fill=paired-light-orange!50,draw=red!0,line width=0.3mm},
    model_work_left/.style =      {align=left,text width=4cm,rounded corners=3pt, fill=paired-light-orange!50,draw=red!0,line width=0.3mm}, 
    model_work_small/.style =     {align=left,text width=2cm,rounded corners=3pt, fill=paired-light-orange!50,draw=red!0,line width=0.3mm},  
    model_work_small_2/.style =     {align=left,text width=4cm,rounded corners=3pt, fill=paired-light-orange!50,draw=red!0,line width=0.3mm}, 
    pretraining/.style =           {align=center,text width=2cm,rounded corners=3pt, fill= paired-light-green!50,draw=paired-dark-green!75,line width=0.3mm}, 
    pretraining_wide/.style =           {align=center,text width=3cm,rounded corners=3pt, fill= paired-light-green!50,draw=paired-dark-green!75,line width=0.3mm}, 
    pretraining_more/.style =           {align=center,text width=4.5cm,rounded corners=3pt, fill= paired-light-green!50,draw=paired-dark-green!75,line width=0.3mm},   
    pretraining_work/.style =           {align=center,text width=4.5cm,rounded corners=3pt, fill= paired-light-green!50,draw= cyan!0,line width=0.3mm},      
    finetuning/.style =           {align=center,text width=2cm,rounded corners=3pt, fill= paired-light-purple!50,draw=paired-dark-purple!75,line width=0.3mm},   
    finetuning_wide/.style =      {align=center,text width=3.5cm,rounded corners=3pt, fill= paired-light-purple!50,draw=paired-dark-purple!75,line width=0.3mm},   
    finetuning_work/.style =      {align=center,text width=7.5cm,rounded corners=3pt, fill= paired-light-purple!50,draw= orange!0,line width=0.3mm},        
    inference/.style =           {align=center,text width=2cm,rounded corners=3pt, fill= paired-light-red!35,draw=paired-light-red!90,line width=0.3mm},           
    inference_more/.style =      {align=center,text width=4.45cm,rounded corners=3pt, fill= paired-light-red!35,draw=paired-light-red!90,line width=0.3mm},
    inference_work/.style =      {align=left,text width=4cm,rounded corners=3pt, fill= paired-light-red!35,draw= magenta!0,line width=0.3mm},      
    application/.style =           {align=center,text width=2cm,rounded corners=3pt, fill= paired-light-yellow!15,draw=paired-light-yellow!90,line width=0.3mm},           
    application_more/.style =      {align=center,text width=11.4cm,rounded corners=3pt, fill= paired-light-yellow!15,draw=paired-light-yellow!90,line width=0.3mm},
    application_work/.style =      {align=center,text width=4.5cm,rounded corners=3pt, fill= paired-light-yellow!15,draw= magenta!0,line width=0.3mm},   
}

\usepackage{color} 

\usepackage{ragged2e}  

\usepackage[capitalize]{cleveref}
\crefname{section}{Sec.}{Secs.}
\Crefname{section}{Section}{Sections}
\Crefname{table}{Table}{Tables}
\crefname{table}{Tab.}{Tabs.}
\crefname{algorithm}{Algo.}{Algos.}

\usepackage{bm}

\usepackage{orcidlink} 

\usepackage{pifont}

\usepackage{graphicx}
\usepackage{amsmath}

\usepackage{amssymb}

\usepackage{booktabs}
\usepackage{amsthm,amsmath,amssymb}
\usepackage{mathrsfs}

\usepackage{bbding}

\usepackage{amsmath}
\DeclareFontFamily{U}{mathc}{}
\DeclareFontShape{U}{mathc}{m}{it}%
{<->s*[1.03] mathc10}{}

\DeclareMathAlphabet{\mathscr}{U}{mathc}{m}{it}

\usepackage{array, tabularx, boldline}
\usepackage{graphicx}
\usepackage{cellspace}
\usepackage{color}

\usepackage{epsfig}
\usepackage{graphicx}
\usepackage{amsmath}
\usepackage{amssymb}
\usepackage{amsthm}
\usepackage{algpseudocode}
\usepackage[ruled]{algorithm2e} 

\usepackage{graphicx}
\usepackage{amsmath}
\usepackage{amssymb}
\usepackage{booktabs}
\usepackage{float}
\usepackage{pifont}
\usepackage{enumitem}
\usepackage{multirow}

\definecolor{myblue}{rgb}{0.27, 0.80, 1.0}
\definecolor{mygreen}{rgb}{0.6, 1.0, 0.6}
\definecolor{myred}{rgb}{1.0, 0.2, 0.2}

\usepackage{makecell}
\usepackage{colortbl}
\usepackage{booktabs}
\usepackage{multirow}

\usepackage{bm}

\newlength\secmargin
\newlength\subsecmargin
\newlength\paramargin
\newlength\figmargin
\newlength\eqmargin
\usepackage[capitalize]{cleveref}
\crefname{section}{Sec.}{Secs.}
\Crefname{section}{Section}{Sections}
\Crefname{table}{Table}{Tables}
\crefname{table}{Tab.}{Tabs.}
\crefname{algorithm}{Algo.}{Algos.}

\begin{document}
%

\title{Toward Visual Grounding: A Survey}

%
%
%
%

\author{{Linhui~Xiao$^{\orcidlink{0000-0003-2592-5264}}$, Xiaoshan~Yang$^{\orcidlink{0000-0001-5453-9755}}$, Xiangyuan~Lan$^{\orcidlink{0000-0001-8564-0346}}$, \\ Yaowei~Wang$^{\orcidlink{0000-0002-6110-4036}}$, \IEEEmembership{Member,~IEEE}, and~Changsheng~Xu$^{\orcidlink{0000-0001-8343-9665}}$, \IEEEmembership{Fellow,~IEEE}}
\IEEEcompsocitemizethanks{

\IEEEcompsocthanksitem Linhui Xiao is with Pengcheng Laboratory (PCL), Shenzhen 518066, China, also with Institute of Automation, Chinese Academy of Sciences (CASIA), Beijing 100190, China, and also with School of Artificial Intelligence, University of Chinese Academy of Sciences (UCAS), Beijing 100049, China (e-mail: xiaolinhui16@mails.ucas.ac.cn). 

\IEEEcompsocthanksitem Xiaoshan Yang, and Changsheng Xu are with State Key Laboratory of Multimodal Artificial Intelligence Systems (MAIS), Institute of Automation, Chinese Academy of Sciences (CASIA), Beijing 100190, China, also with Pengcheng Laboratory (PCL), Shenzhen 518066, China, and also with School of Artificial Intelligence, University of Chinese Academy of Sciences (UCAS), Beijing 100049, China (e-mail: xiaoshan.yang@nlpr.ia.ac.cn, csxu@nlpr.ia.ac.cn).

\IEEEcompsocthanksitem Xiangyuan~Lan is with Pengcheng Laboratory (PCL), Shenzhen 518066, China (e-mail: lanxy@pcl.ac.cn).

\IEEEcompsocthanksitem Yaowei~Wang is with Harbin Institute of Technology (Shenzhen), Shenzhen 518055, China, and also with Pengcheng Laboratory (PCL), Shenzhen 518066, China (e-mail: wangyw@pcl.ac.cn).

\IEEEcompsocthanksitem Corresponding author: Changsheng Xu.

\IEEEcompsocthanksitem This work was supported in part by the Major Key Project of PCL under Grant PCL2025A14, in part by the National Natural Science Foundation of China under Grants U23A20387, 62322212, 62036012, 62072455, 62536003, 62402252, in part by National Science and Technology Major Project under Grant 2021ZD0112200, and also in part by CAS Project for Young Scientists in Basic Research (YSBR-116).

\IEEEcompsocthanksitem{Digital Object Identifier \url{https://doi.org/10.1109/TPAMI.2025.3630635}}
}
}

%
%

\markboth{IEEE TRANSACTIONS ON PATTERN ANALYSIS AND MACHINE INTELLIGENCE, November 2025}
{Shell \MakeLowercase{\textit{et al.}}: Bare Advanced Demo of IEEEtran.cls for IEEE Computer Society Journals}
%

\IEEEpubid{1520-9210~\copyright~2025 IEEE. Personal use is permitted, but republication/redistribution requires IEEE permission.}



\IEEEtitleabstractindextext{
\begin{abstract}
\justifying
Visual Grounding, also known as Referring Expression Comprehension and Phrase Grounding, aims to ground the specific region(s) within the image(s) based on the given expression text. This task simulates the common referential relationships between visual and linguistic modalities, enabling machines to develop human-like multimodal comprehension capabilities. Consequently, it has extensive applications in various domains. However, since 2021, visual grounding has witnessed significant advancements, with emerging new concepts such as grounded pre-training, grounding multimodal LLMs, generalized visual grounding, and giga-pixel grounding, which have brought numerous new challenges. In this survey, we first examine the developmental history of visual grounding and provide an overview of essential background knowledge, including fundamental concepts and evaluation metrics. We systematically track and summarize the advancements, and then meticulously define and organize the various settings to standardize future research and ensure a fair comparison.  In the dataset section, we compile a comprehensive list of current relevant datasets, conduct a fair comparative analysis, and provide ultimate performance prediction to inspire the development of new standard benchmarks. Additionally, we delve into numerous applications and highlight several advanced topics. Finally, we outline the challenges confronting visual grounding and propose valuable directions for future research, which may serve as inspiration for subsequent researchers. By extracting common technical details, this survey encompasses the representative work in each subtopic over the past decade. To the best of our knowledge, this paper represents the most comprehensive overview currently available in the field of visual grounding. This survey is designed to be suitable for both beginners and experienced researchers, serving as an invaluable resource for understanding key concepts and tracking the latest research developments. We keep tracing related work at \url{https://github.com/linhuixiao/Awesome-Visual-Grounding}.


\end{abstract}

\begin{IEEEkeywords}
 Visual Grounding, Referring Expression Comprehension, Phrase Grounding, Survey
\end{IEEEkeywords}}

\maketitle

\IEEEdisplaynontitleabstractindextext

%
\IEEEpeerreviewmaketitle

\section{Introduction}
\label{sec:introduction}

\IEEEPARstart{I}{n} the field of artificial intelligence (AI) \cite{lecun2015deep, minsky1961steps, nilsson1982principles, winston1984artificial}, multimodal learning \cite{baltruvsaitis2018multimodal, xu2023multimodal} that combines visual perception and natural language understanding has emerged as a pivotal approach for achieving human-like cognition in machines. At its core lies the integration of visual and linguistic cues, intending to bridge the semantic gap between image scenes and language descriptions. Visual Grounding (VG) \cite{refcocog-google, refcocog-umd, yu2016modeling} represents such a fundamental pursuit, encompassing AI models' ability to establish intrinsic connections between linguistic expressions and corresponding visual elements.

\begin{figure}[!t]
	\centering
	\includegraphics[width=0.95\linewidth]{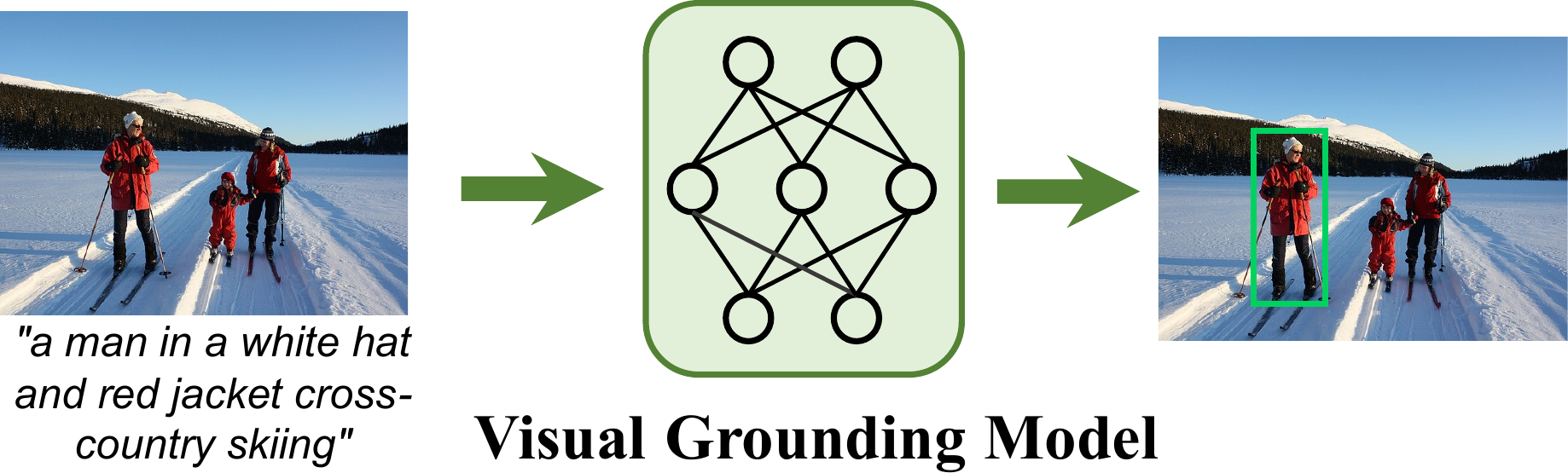}
        \vspace{-5pt}
        \caption{An illustration of visual grounding.}
        \vspace{-17pt}
	\label{fig:grounding}
\end{figure}


As depicted in \cref{fig:grounding}, visual grounding, also known as Referring Expression Comprehension (REC) and Phrase Grounding (PG), according to the classical definition \cite{deng2021transvg, yu2018mattnet, yang2020resc}, involves \textit{localizing a specific region within an image based on a given textual description}, and such a description is called ``\textit{referring expression}'' \cite{van2006building, viethen2008use, golland2010game, mitchell2010natural, mitchell2013generating, fitzgerald2013learning, kazemzadeh2014referitgame, refcocog-google}. The objective of this task is to emulate the prevalent referential relationships in social conversations, equipping machines with human-like multimodal comprehension capabilities. Consequently, it has extensive applications in visual language navigation \cite{anderson2018navigation}, human-machine dialogue \cite{das2017visual_dialog, chen2023shikra}, visual question answering \cite{fukui2016mcb, antol2015vqa}, and other related domains \cite{qiao2020referring}.

\begin{figure}[!t]
	\centering
	\includegraphics[width=0.8\linewidth]{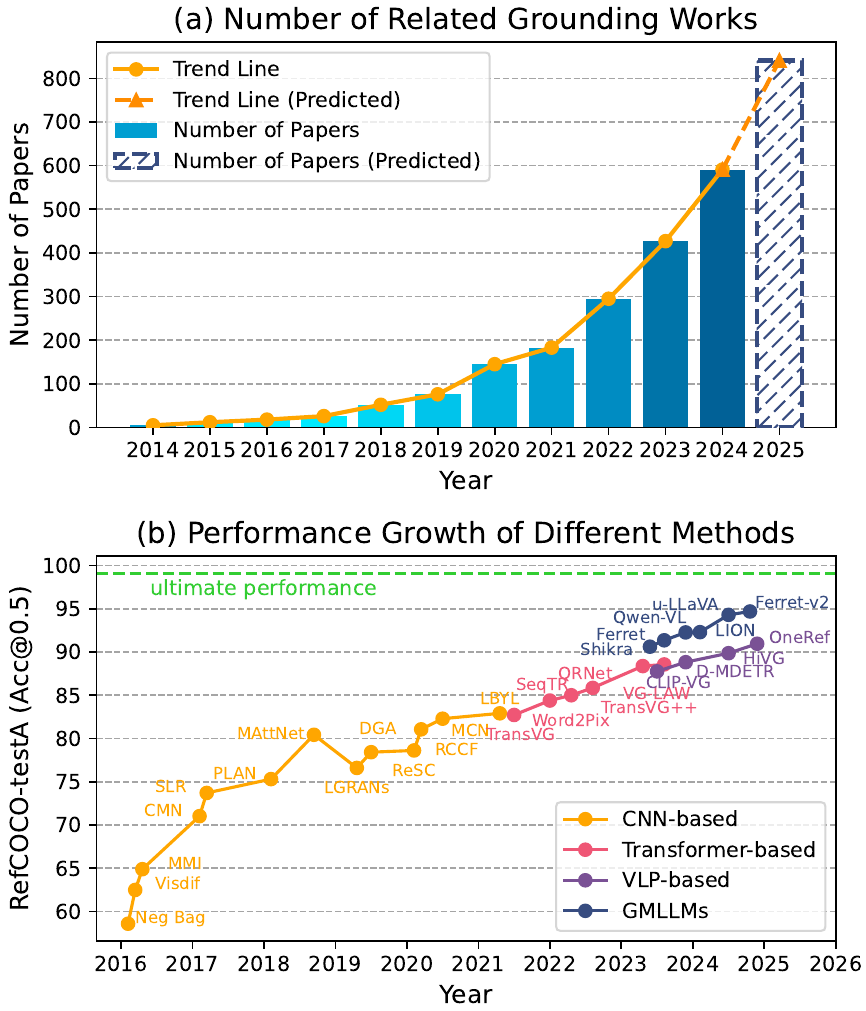}
        \vspace{-8pt}
        \caption{\small The number of papers and performance trends of visual grounding over the past decade. The data in panel (a) are derived from an exact-match lookup on Google Scholar for the term ``referring expression comprehension". The GMLLMs in (b) are the 7B version.}
        \vspace{-17pt}
	\label{fig:perf_trend}
\end{figure}

%
%

The continuous advancements in deep learning, including visual grounding, are driven by three fundamental elements: \textit{data, algorithms, and computing power} \cite{duan2019artificial}. From a \textit{data} perspective, the grounding task involves three essential types of data: \texttt{images}, \texttt{referring expressions}, and \texttt{referred bounding boxes}. However, obtaining such paired triplet data is not straightforward, despite images being more readily available among these three types. Challenges arise when acquiring expression text and corresponding bounding boxes. \textbf{\textit{Firstly}},  visual grounding heavily relies on high-quality and \textbf{\textit{unambiguous}} textual referring expression data. In 1975, Paul Grice proposed a rational principle for interactions in natural language dialogues called \textit{Gricean Maxims} \cite{grice1975logic}. This criterion reflects the requirement that when describing an object in a complex real scene, it should be \textit{informative, concise}, and \textit{unambiguous} \cite{refcocog-google, yu2016modeling}. The \textbf{\textit{unambiguity}} of referring expressions is particularly crucial due to the presence of multiple objects belonging to the same class within a real-life scene \cite{yu2016modeling, refcocog-google, xiao2023clip}. If the expression is ambiguous, valuable information cannot be effectively learned by the model and instead leads to confusion. Consequently, as shown in \cref{fig:timeline}, before 2014, a substantial amount of research \cite{van2006building, viethen2008use, golland2010game, mitchell2010natural,  fitzgerald2013learning, krahmer2012computational} primarily focused on the Referring Expression Generation (REG), while grounding received minimal attention. \textbf{\textit{Secondly}}, obtaining paired bounding boxes is also labor-intensive. In the early stage, a substantial amount of research (\eg, DT-RNN (2014)\cite{socher2014grounded}, DMSM (2015)\cite{fang2015captions}, Neg bag (2016) \cite{refcocog-umd}) was predominantly focused on weakly supervised settings due to the scarcity of available paired bounding boxes. In 2014, Kazemzadeh \textit{et al.} \cite{kazemzadeh2014referitgame} introduced the first large-scale real-world expression understanding dataset called ReferIt Game, which gradually shifted fully supervised visual grounding towards more realistic scenarios. However, due to the limited image categories and simplistic referring text in the ReferIt Game, it fails to meet the requirements of unambiguity. As a result, in 2016, Mao \etal \cite{refcocog-google} and Nagaraja \etal \cite{refcocog-umd} proposed and reorganized the RefCOCOg datasets based on the MS COCO \cite{mscoco} image dataset. Subsequently, Yu  \etal \cite{yu2016modeling}, in the same year, proposed the RefCOCO/+ \cite{yu2016modeling} datasets. These three datasets laid a solid foundation for subsequent grounding research and have become standard benchmarks over the following decade. As shown in \cref{fig:perf_trend}-(a), since then, numerous studies on visual grounding have emerged. Over time, in 2021, Kamath \etal ~\cite{kamath2021mdetr} incorporated multiple regional datasets while treating grounding as a modulated detection task, thereby significantly improving the learning of fine-grained representation. Subsequently, with the advancement of the pre-training paradigm, larger fine-grained datasets such as GRIT \cite{peng2024kosmos-2} have emerged in recent years to push visual grounding to unprecedented heights continuously.

From the perspective of \textit{algorithms} and \textit{computing power}, the research on visual grounding is constantly evolving under the influence of mainstream deep learning algorithms and increased computational capability. As shown in \cref{fig:timeline}, based on the development of deep learning algorithms, we can broadly categorize the research on visual grounding into three stages: \textbf{\textit{preliminary stage}} (before 2014), \textbf{\textit{early stage}} (2014-2020), and \textbf{\textit{surge stage}} (2021-present). \textit{Before 2014}, visual grounding was not yet systematically established; it served as a validation task to assist REG. During that time, the main method involved selecting proposals in a weakly supervised manner using language analysis tools \cite{socher2014grounded}. \textit{From 2014 to 2020}, language encoding was performed using small-scale Long Short-Term Memory (LSTM) networks \cite{hochreiter1997long} while image encoding utilized Convolutional Neural Networks (CNNs) \cite{he2016deep}. Grounding results were achieved through two-stage \cite{yu2018mattnet,cm-att-erase, hong2019rvg-tree} or one-stage \cite{yang2020resc,yang2019fast,zhou2021real} approaches. However, \textit{starting from 2021}, the LSTM and CNN methods gradually fell out of favor with the introduction of Transformer \cite{transformer}. Concurrently, driven by advancements in pre-trained models, the paradigm shifted towards \textit{``pre-training then fine-tuning''} for downstream transfer tasks. Consequently, both unimodal pre-trained models (\eg, BERT \cite{devlin2019bert}, DETR \cite{carion2020detr}, Swin Transformer \cite{liu2021swin}, DINO \cite{zhang2023dino} \etc) and Visual Language Pre-trained (VLP) models (\eg, ALBEF \cite{li2021albef}, CLIP \cite{radford2021clip}, BEiT-3 \cite{beit3}, OFA \cite{wang2022ofa}, \etc) started to be employed in grounding. This period also witnessed the emergence of various settings, including full supervision, weak supervision, zero-shot learning, and others. Furthermore, propelled by rapid advancements in computational power, both the model sizes and training data volumes have significantly expanded. This has led to the manifestation of the \textit{Scaling Law} \cite{hoffmann2022training} in deep learning that also impacts research on visual grounding. From 2023 onwards, Large Language Models (LLMs) \cite{gpt-2} and multimodal counterparts (MLLMs) \cite{gpt3} have demonstrated remarkable efficacy, leading to a proliferation of Grounding Multimodal Large Language Models (GMLLMs) \cite{you2023ferret}. Within just over one year, numerous representative methods (\eg, Shikra \cite{chen2023shikra}, LION \cite{chen2024lion}, \etc) have emerged.

\begin{figure}[t!]
	\centering
	\includegraphics[width=1.0\linewidth]{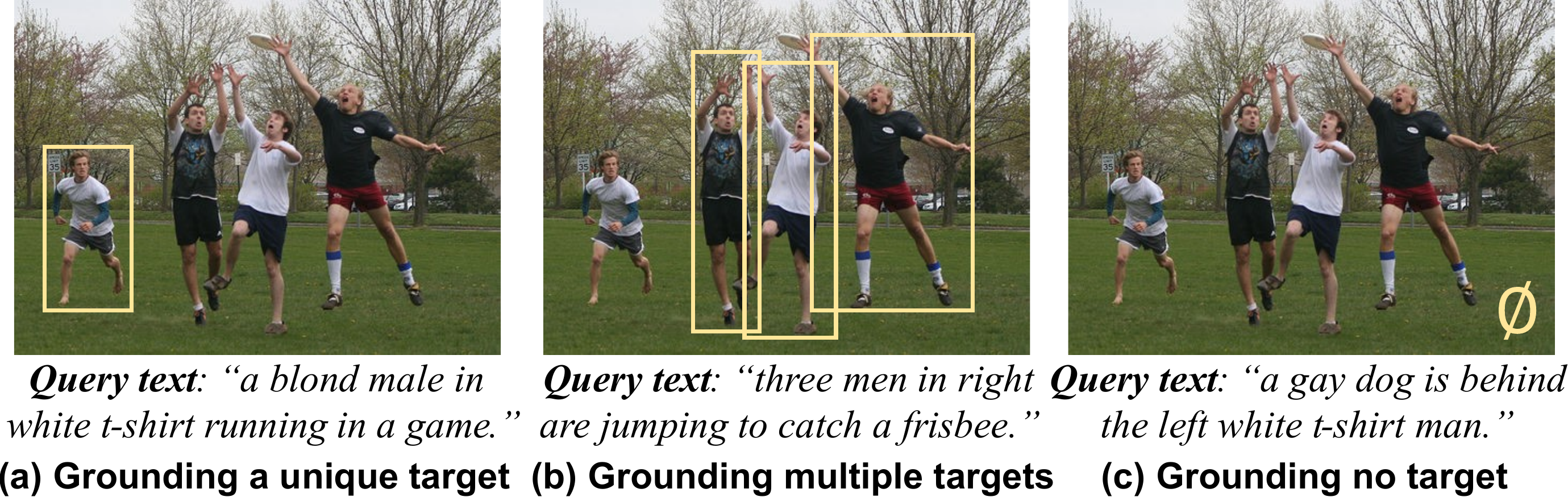}
        \vspace{-15pt}
        \caption{A future-oriented definition of generalized grounding.}
        \vspace{-10pt}
	\label{fig:definition}
\end{figure}

Although visual grounding has witnessed significant advancements over the past decade, it also leads to the accumulation of numerous challenges. \textbf{\textit{(i) Firstly}}, due to the complexity of acquiring triplet data and the availability of various pre-trained models, a wide variety of experimental settings have emerged (\eg, fully supervised \cite{deng2021transvg}, weakly supervised \cite{xiao2017weakly}, semi-supervised \cite{zhu2021utilizing, chou2022semi}, unsupervised \cite{yeh2018unsupervised, jiang2022pseudo}, zero-shot \cite{sadhu2019zsgnet, subramanian2022reclip}, and among others\cite{wang2024ovvg, xie2024described}). These settings can be confusing, often characterized by unclear boundaries and ambiguous definitions, leading to potentially unfair comparisons. For instance, direct comparisons are made between models trained on multiple datasets and fine-tuned using single datasets within the fully supervised setting (\eg, \cite{yang2022unitab, ho2023yoro}); methods that utilizing large-scale VLP models are directly compared with those using unimodal pre-trained models (\eg, \cite{shi2023dynamic}); zero-shot settings are misinterpreted as weakly supervised (\eg, \cite{zhao2024ppt}); unsupervised and weakly supervised settings suffer from vague definitions (\eg, \cite{wang2019phrase}). Nevertheless, no prior work has systematically addressed or summarized these issues so far. \textbf{\textit{(ii) Secondly}}, the datasets are limited and lack clarity in terms of future research direction. Specifically, the RefCOCO/+/g \cite{refcocog-google, refcocog-umd, yu2016modeling} datasets have been proposed for nearly ten years and continue to serve as the core evaluation benchmarks. However, as shown in \cref{fig:perf_trend}-(b), its performance gains are becoming increasingly limited. Additionally, with the emergence of LLMs, the existing datasets no longer meet the requirements of basic tasks. For instance, as shown in \cref{fig:definition}, while the current dataset focuses on grounding one specific object, according to the concept of grounding, a comprehensive dataset should encompass three conditions: \textit{(a)} grounding for one target, \textit{(b)} grounding for multiple targets, and \textit{(c)} grounding for no target. \textbf{\textit{(iii) Thirdly,}} there is a lack of systematic review that can summarize existing work and provide guidance for future research. As shown in \cref{fig:perf_trend}, due to the excessive amount of literature, many of the latest methods fail to adequately address and compare existing papers on similar ideas or settings. Among the existing surveys, the technical review by Qiao \etal \cite{qiao2020referring} primarily focuses on research conducted before 2020, whereas other reviews \cite{wang2025vg_survey_for_rs} in the grounding field offer only a limited scope. Over the past five years, the multimodal community has witnessed significant advancements, and a growing body of grounding research has emerged, marking a clear departure from the earlier landscape. Consequently, a systematic review is urgently needed to synthesize these recent developments and identify promising directions for future research.

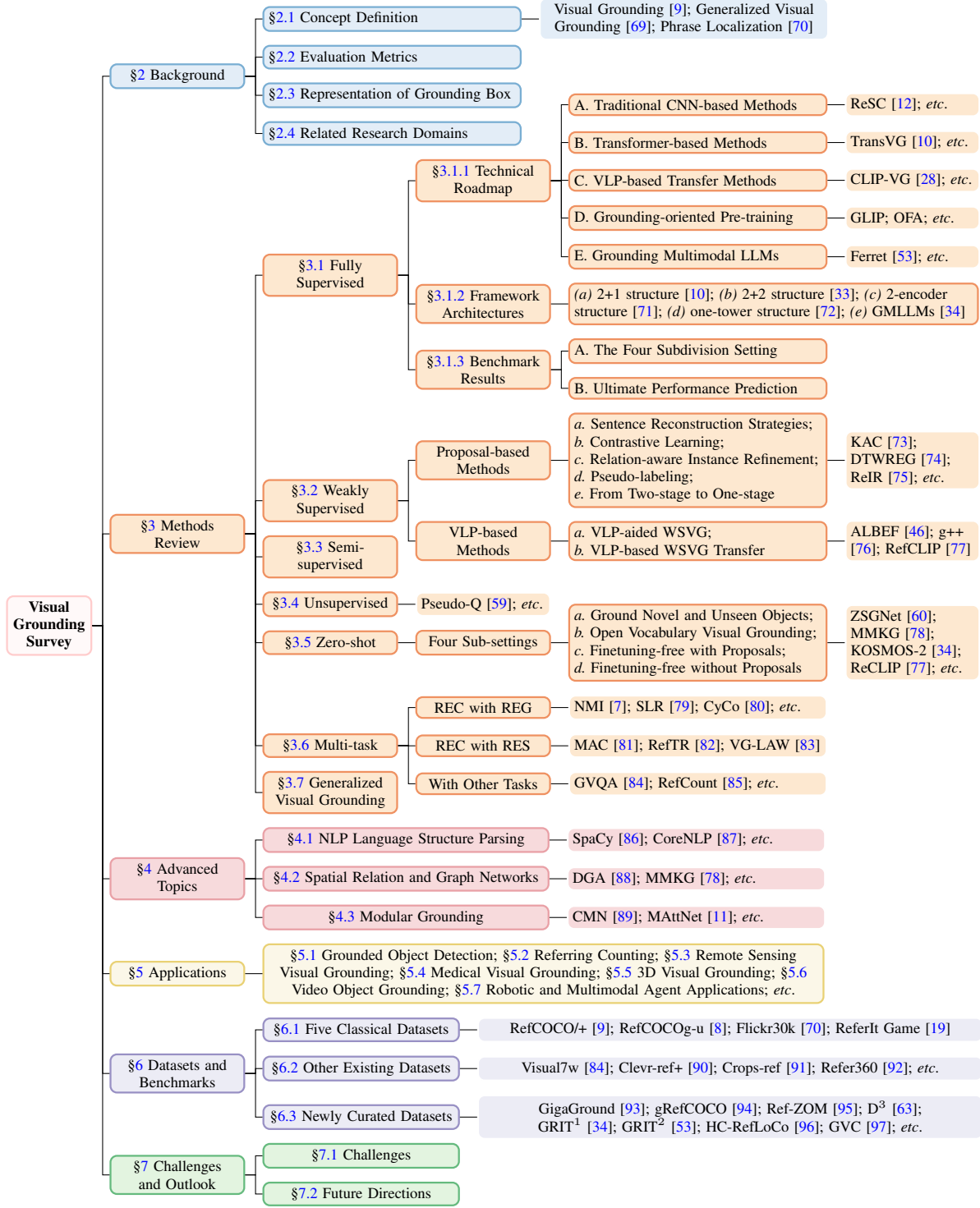
\begin{figure*}
\centering
\scriptsize
\hspace*{-30pt}
    \begin{forest}
        for tree={
            forked edges,
            grow'=0,
            draw,
            rounded corners,
            node options={align=center,},
            text width=2.7cm,
            s sep=6pt,
            calign=edge midpoint,
        },
        [\textbf{Visual} \\ \textbf{Grounding} \\ \textbf{Survey}, fill=gray!45, parent
            [\S \ref{sec:introduction} Introduction, for tree={pretraining}
                [Development History and Status, pretraining_more]
            ]
            [\S \ref{sec:background} Background, for tree={data}
                [\S \ref{subsec:definition} Concept Definition ,  data_wide  
                    [Visual Grounding \cite{yu2016modeling};  Generalized Visual Grounding \cite{he2023grec}; Phrase Localization \cite{plummer2015flickr30k}, data_work]
                ]
                [\S \ref{subsec:evaluation_metric} Evaluation Metrics, data_wide]
                [\S \ref{subsec:representation_of_box} Representation of Grounding Box,  data_wide]
                [\S \ref{subsec:realted_domain} Related Research Domains,  data_wide]
            ]
            [\S\ref{sec:method_survey} Methods Review, for tree={fill=red!45,model}
                [\S\ref{subsec:fully_sup} Fully Supervised, model
                    [\S \ref{subsubsec:road_map} Technical Roadmap, model
                        [A. Traditional CNN-based Methods, model_more_left
                        [ReSC \cite{yang2020resc}; \etc, model_work_small]
                        ]
                        [B. Transformer-based Methods, model_more_left
                        [TransVG \cite{deng2021transvg}; \etc, model_work_small]
                        ]
                        [C. VLP-based Transfer Methods, model_more_left
                        [CLIP-VG \cite{xiao2023clip}; \etc, model_work_small]
                        ]
                        [D. Grounding-oriented Pre-training, model_more_left
                        [GLIP; OFA; \etc, model_work_small]
                        ]
                        [E. Grounding Multimodal LLMs, model_more_left
                        [Ferret \cite{you2023ferret}; \etc, model_work_small]
                        ]
                    ]
                    [\S \ref{subsubsec:model_arch} Framework Architectures, model
                        [\textit{(a)} 2+1 structure \cite{deng2021transvg}; \textit{(b)} 2+2 structure \cite{kamath2021mdetr}; \textit{(c)} 2-encoder structure \cite{transvg++}; \textit{(d)} one-tower structure \cite{xiao2024oneref}; \textit{(e)} GMLLMs \cite{peng2024kosmos-2}, model_large_left]
                    ]
                    [\S \ref{subsubsec:benchmark_result} Benchmark Results, model
                        [A. The Four Subdivision Setting, model_more_left]
                        [B. Ultimate Performance Prediction, model_more_left]
                    ]
                ]
                [\S\ref{subsec:weakly_sup} Weakly Supervised, model
                    [Proposal-based Methods, model
                        [\textit{a.} Sentence Reconstruction Strategies; \\
                         \textit{b.} Contrastive Learning; \\
                         \textit{c.} Relation-aware Instance Refinement; \\
                         \textit{d.} Pseudo-labeling; \\
                         \textit{e.} From Two-stage to One-stage, model_more_left
                            [KAC \cite{chen2018knowledge}; DTWREG \cite{sun2021discriminative}; ReIR \cite{liu2021relation}; \etc, model_work_small
                            ]
                         ]
                    ]
                    [VLP-based Methods, model
                        [\textit{a.} VLP-aided WSVG; \\
                         \textit{b.} VLP-based WSVG Transfer, model_more_left
                            [ALBEF \cite{li2021albef}; g++ \cite{shaharabany2023similarity}; RefCLIP \cite{jin2023refclip}, model_work_small
                            ]
                         ]
                    ]
                ]
                [\S\ref{subsec:semi-sup_setting} Semi-supervised, model]
                [\S\ref{subsec:unsup_setting} Unsupervised, model
                    [Pseudo-Q \cite{jiang2022pseudo}; \etc, model_work_small]
                ]
                [\S\ref{subsec:zero-shot} Zero-shot, model
                    [Four Sub-settings, model
                        [\textit{a.} Ground Novel and Unseen Objects; \\
                         \textit{b.} Open Vocabulary Visual Grounding; \\
                         \textit{c.} Finetuning-free with Proposals; \\
                         \textit{d.} Finetuning-free without Proposals, model_more_left
                            [ZSGNet \cite{sadhu2019zsgnet}; MMKG \cite{shi2022mmkg}; KOSMOS-2 \cite{peng2024kosmos-2}; ReCLIP \cite{jin2023refclip}; \etc, model_work_small
                            ]                     
                        ]
                    ]
                ]
                [\S\ref{subsec:multi-task_setting} Multi-task, model
                    [REC with REG, model
                        [MMI \cite{refcocog-google}; SLR \cite{yu2017joint}; CyCo \cite{wang2024cycle}; \etc, model_work_small_2]
                    ]
                    [REC with RES, model
                        [MAC \cite{luo2020mcn}; RefTR \cite{li2021referring}; VG-LAW \cite{vg-law}, model_work_small_2]
                    ]
                    [With Other Tasks, model
                        [GVQA \cite{zhu2016visual7w}; RefCount \cite{dai2024ref-count}; \etc, model_work_small_2]
                    ]
                ]
                [\S\ref{subsec:grec} Generalized Visual Grounding, model
                ]
            ]
            [Appendix \S \textcolor{blue}{A2} Datasets and Benchmarks, for tree={finetuning}
                [\S \textcolor{blue}{A2.1} Datasets for Classical Visual Grounding, finetuning_wide
                [RefCOCO/+ \cite{yu2016modeling}; RefCOCOg-u \cite{refcocog-umd}; Flickr30k \cite{plummer2015flickr30k}; ReferIt Game \cite{kazemzadeh2014referitgame}; VLM-VG \cite{wang2025learning}; Clevr-ref+ \cite{liu2019clevr-ref+}; Crops-ref \cite{chen2020cops-ref}; Refer360 \cite{cirik2020refer360}; \etc, finetuning_work]
                ]
                [\S \textcolor{blue}{A2.2} Datasets for Generalized Visual Grounding, finetuning_wide
                [gRefCOCO \cite{liu2023gres}; Ref-ZOM \cite{hu2023beyond}; D$^3$ \cite{xie2024described}; RefDrone \cite{sun2025refdrone}; \etc, finetuning_work]
                ]
                [\S \textcolor{blue}{A2.3} Datasets and Benchmarks for GMLLMs, finetuning_wide
                [GRIT$^1$ \cite{peng2024kosmos-2}; GRIT$^2$ \cite{you2023ferret}; HC-RefLoCo \cite{wei2024hc-refloco}; HumanRef \cite{jiang2025humanref}; GVC \cite{zhang2024llava-grounding}; Ref-L4 \cite{chen2024ref-l4}; \etc, finetuning_work]
                ]                
                [\S \textcolor{blue}{A2.4} Datasets for Universal Grounding Scenarios, finetuning_wide
                [MGrounding-630K \cite{li2025migician}; MIG-Bench \cite{li2025migician}; MC-Bench \cite{xu2024mc-bench}; GigaGround \cite{ma2024visual};   \etc, finetuning_work]
                ]
            ]
            [Appendix \S \textcolor{blue}{A3} Applications, for tree={application}
                [\S \textcolor{blue}{A3.1} Grounded Object Detection; \S \textcolor{blue}{A3.2} Video Object Grounding; \S \textcolor{blue}{A3.3} Referring Counting; \S \textcolor{blue}{A3.4} Remote Sensing Visual Grounding; \S \textcolor{blue}{A3.5} Medical Visual Grounding; \S \textcolor{blue}{A3.6} 3D Visual Grounding; \S \textcolor{blue}{A3.7} Speech REC; \S \textcolor{blue}{A3.8} Robotic and Multimodal Agent Systems; \S \textcolor{blue}{A3.9} Industrial Applications; \etc, application_more]
            ]
            [Appendix \S \textcolor{blue}{A4}  Advanced \\ Topics, for tree={inference}
                [\S \textcolor{blue}{A4.1} NLP Language Structure Parsing, inference_more
                    [SpaCy \cite{honnibal2015spacy}; CoreNLP \cite{chen2014fast}; \etc, inference_work]
                ]
                [\S \textcolor{blue}{A4.2} Spatial Relation and Graph Networks, inference_more
                    [DGA \cite{yang2019dynamic}; MMKG \cite{shi2022mmkg}; \etc, inference_work]
                ]
                [\S \textcolor{blue}{A4.3} Modular Grounding, inference_more
                    [CMN \cite{hu2017modeling}; MAttNet \cite{yu2018mattnet}; \etc, inference_work]
                ]
            ]
            [\S \ref{sec:future_direction} Challenges and Outlook, for tree={pretraining}
                [\S \ref{subsec:challenge} Challenges, pretraining_more]
                [\S \ref{subsec:future_direction} Future Directions, pretraining_more]
            ]
            [\S \ref{sec:conclusion} Conclusion, for tree={pretraining}
            ]
        ]
    \end{forest}
    \vspace{-2pt}
    \caption{Overview of the paper structure, detailing Chapter \ref{sec:introduction}-\ref{sec:future_direction}, and Appendix Chapter \textcolor{blue}{A2}-\textcolor{blue}{A4}.}
    \vspace{-15pt}
    \label{fig:paper_structure}
\end{figure*}

\textbf{\textit{Survey pipeline:}} As illustrated in \cref{fig:paper_structure}, in this survey, we present a streamlined roadmap to address the afore-mentioned challenges. Firstly, \cref{sec:introduction} provides a brief overview of the historical development of visual grounding. Subsequently, \cref{sec:background} covers essential background information, encompassing definitions, evaluation criteria, and related research domains. In \cref{sec:method_survey}, we systematically review current research from seven perspectives: fully supervised, weakly supervised, semi-supervised, unsupervised, zero-shot, multi-task, and generalized grounding. The mainstream fully supervised setting will be discussed in the highlight, and the benchmark results across different settings will be compared. In \cref{sec:future_direction}, we will discuss current challenges and propose potential future directions. To provide a more comprehensive review, we will provide a detailed overview of the available datasets in Appendix Sec.\textcolor{blue}{A2}. Furthermore, we introduce the applications of grounding in Sec. \textcolor{blue}{A3} and discuss several advanced topics in Sec. \textcolor{blue}{A4} of the Appendix. Finally, a conclusion is provided in \cref{sec:conclusion}.


\textbf{\textit{Contributions:}}  \textbf{\textit{(i)}} Since the review by Qiao \etal \cite{qiao2020referring} in 2020, we are the first survey in the past five years to systematically track and summarize the development of visual grounding over the last decade. By extracting common technical details, this review encompasses the most representative work in each subtopic. \textbf{\textit{(ii)}} We meticulously organize various settings in VG and establish precise definitions for these settings to standardize future research, ensuring a fair and just comparison. \textbf{\textit{({\romannumeral 3})}} We compile datasets from recent years and provide ultimate performance prediction on five classical datasets to inspire the development of new standard benchmarks. \textbf{\textit{({\romannumeral 4})}} We consolidate current research challenges and provide valuable directions for future investigations that can enlighten subsequent researchers. \textbf{\textit{({\romannumeral 5})}} To the best of our knowledge, this survey is currently the most comprehensive review in the field of visual grounding. We aim for this article to serve as a valuable resource for beginners seeking an introduction to grounding and researchers with an established foundation, enabling them to navigate and stay up-to-date with the latest advancements.

Finally, this field is rapidly evolving, making it challenging for us to keep pace with the latest developments. We encourage researchers to share their new findings with us, ensuring that we stay updated. These new methods will be incorporated and discussed in the revised version and tracked in our project repository. 


\section{Background}
\label{sec:background}

\indent \textbf{\textit{Overview:}} In this section, we will provide a comprehensive definition of visual grounding and present an in-depth discussion on the corresponding evaluation metrics. Furthermore, we will introduce several closely related research domains.

\subsection{Concept Definition}
\label{subsec:definition}

We provide three grounding-related concept definitions.

\vspace{2pt}
\noindent $\bullet$ \textit{\textbf{Classical Visual Grounding.}}  Based on the literature from the past decade, we provide a widely accepted and dataset-related narrow definition. Specifically, \textbf{\textit{\textit{Visual Grounding} (VG) or \textit{Referring Expression Comprehension} (REC), involves localizing a specific region within an image based on a given textual description}}. When the descriptive text consists of only a few short words, it is referred as Phrase Grounding (PG). Current literature \cite{deng2021transvg} commonly associates PG with ReferIt Game \cite{kazemzadeh2014referitgame} and Flickr30k Entities \cite{plummer2015flickr30k} datasets, while it is termed REC when related to RefCOCO/+/g \cite{yu2016modeling, refcocog-umd} datasets. 


\vspace{+2pt}
\noindent $\bullet$ \textit{\textbf{Generalized Visual Grounding.}} The traditional VG is built based on a strong assumption that there must be only one object described by a sentence within an image, which is not applicable to real-world scenarios. Consequently, previous models fail when dealing with expressions referring to multiple or no objects. To overcome these limitations, several methods \cite{he2023grec, liu2023gres, xie2024described, hu2023beyond} proposed a similar concept in 2023. Following He \etal \cite{he2023grec}, we nominate such tasks as \textit{\textbf{Generalized Visual Grounding (GVG)}} or \textit{\textbf{Generalized Referring Expression Comprehension (GREC)},  which involve grounding (\textit{a}) one, (\textit{b}) multiple, or even (\textit{c}) no objects described by textual input within an image} (as depicted in \cref{fig:definition}). This concept is also referred to as \textit{\textbf{Described Object Detection (DOD)}} in Xie \etal's work \cite{xie2024described}. It is worth noting that GREC tasks are more suitable for real-world scenarios and possess significant societal application value. For instance, performing a simple query like \textit{``individuals without safety helmet"} in a camera video stream can have wide-ranging uses in engineering construction and traffic safety domains. However, since it encompasses three cases, traditional REC and Open-Vocabulary Detection (OVD) \cite{zareian2021ovrcnn} approaches cannot address it adequately (\ie, OVD can only detect \textit{``individuals"} and \textit{``helmet"} while REC fails to detect multi-target and no-target cases). Conversely, GREC requires models to have a comprehensive understanding of each instance. We will discuss the evaluation metrics, research status, and corresponding dataset of GREC tasks in \cref{subsec:evaluation_metric}, \cref{subsec:grec}, and Sec. \textcolor{blue}{A2.2}, respectively.

\begin{figure*}[t!]
	\centering
	\includegraphics[width=1.0\linewidth]{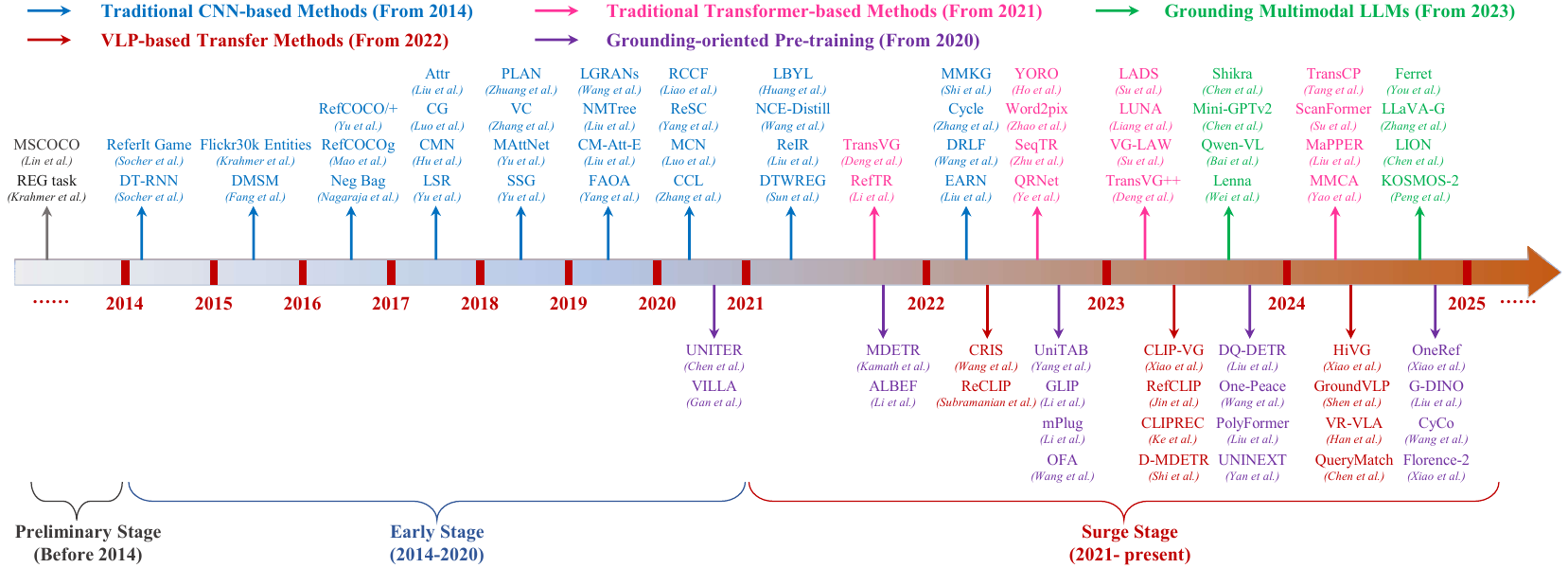}
        \vspace{-16pt}
        \caption{A chronological overview of the representative research progress in fully supervised visual grounding from the perspective of the technical roadmap (\cref{subsubsec:road_map}). The corresponding citations for abbreviated methods can be found in the main text.}
        \vspace{-14pt}
	\label{fig:timeline}
\end{figure*}

\vspace{2pt}
\noindent $\bullet$ \textit{\textbf{Phrase Localization.}} \textbf{\textit{Phrase Localization (PL), also known as Phrase Grounding (PG),}} According to the early literature \cite{plummer2015flickr30k, wang2016structured, plummer2017phrase}, \textbf{\textit{is defined as identifying and localizing all entities mentioned in a textual phrase within an image}}. PL was initially introduced as an application task in the Flickr30k Entities \cite{plummer2015flickr30k} dataset in 2015. Unlike REC, PL requires parsing and extracting noun chunks from the textual phrase using an NLP parser (Sec. \textcolor{blue}{A3.1}), and generating proposals by detectors, followed by scoring, ranking, and pairing these image regions with the corresponding noun entities \cite{plummer2015flickr30k}. This process is not conducive to end-to-end training and makes it challenging to model unique grounding for objects. Consequently, subsequent research \cite{deng2021transvg} has gradually shifted away from this task setting, focusing instead on grounding only the subjects in phrases. The region-to-phrase correspondences established by PL have a direct positive impact on grounded language image pre-training (\eg, MDETR \cite{kamath2021mdetr}, GLIP \cite{li2022glip}), which emerged in 2021. Considering the relatively limited number of PL studies \cite{plummer2015flickr30k, chen2017query}, this survey does not specifically differentiate PL from VG.

\subsection{Evaluation Metrics}
\label{subsec:evaluation_metric}

We denote the learned grounding model as $\mathcal{M}_{g}$. For any given image $\mathcal{I}\in\mathbb{R}^{3\times H\times W}$ and text $\mathcal{T}\in\mathbb{R}^{L_t}$ pairs, a set of predicted bounding box $\hat{\bm{B}}=\{\hat{\mathscr{B}}_i\}_{i=0}^k$ can be obtained through the reasoning of the grounding model:
\begin{equation}
\hat{\bm{B}}= \mathcal{M}_g(\mathcal{I}, \mathcal{T}),
\label{eq:model}
\vspace{-3pt}
\end{equation}
where $H$ and $W$ denote the height and width of the image, $L_t$ represents the length of the text tokens, $\hat{\mathscr{B}}_i = (\hat{x}_i,\hat{y}_i,\hat{w}_i,\hat{h}_i)$ denote the coordinates of each predicted box, and $k = {0, 1, 2, ...}$ is the number of target objects. Specifically, when $k = 1$, it belongs to the classical grounding; when $k = 0$, $\hat{\bm{B}}$ is an empty set.

\noindent $\bullet$ \textit{\textbf{Classical Visual Grounding.}} At the individual sample level, the commonly employed evaluation criterion in visual grounding is the \textit{Intersection over Union} (IoU, \aka, Jaccard overlap) \cite{giou} between the model-predicted grounding box $\hat{\mathcal{B}}$ and the ground truth bounding box $\mathcal{B}=(x,y,w,h)$.
At the dataset level, the performance indicator is typically determined by calculating the proportion of predicted results in all test samples with an IoU value greater than 0.5 (\ie, IoU@0.5($\%$)).

\noindent $\bullet$ \textit{\textbf{Generalized Visual Grounding.}} Under the GVG cases, evaluating becomes challenging. When multiple targets are involved, using the IoU of the mixed region as an evaluation metric may lead to inaccuracies, as larger bounding boxes can obscure smaller ones. Currently, there is no authoritative evaluation scheme in the research community. He \etal \cite{he2023grec} recommended using ``\textit{Precision@(F1=1,IoU$\geq$0.5)}" and ``\textit{N-acc}" as criteria for multi-object and no-object grounding respectively. Specifically, ``\textit{Precision@(F1=1, IoU$\geq$0.5)}" calculates the percentage of samples with an F1 score equal to 1.0 and an IoU threshold set at 0.5. This scheme is relatively reasonable since grounding can be essentially considered a binary classification of target boxes where TP (\textit{true positive}), TN (\textit{true negative}), FP (\textit{false positive}), and FN (\textit{false negative}) are possible outcomes. The F1 score of a sample is calculated as \textit{F1=}$\frac{2TP}{2TP+FN+FP}$. A sample with \textit{F1=1.0} is considered successfully predicted. ``\textit{Precision@(F1=1, IoU$\geq$0.5)}" represents the ratio of successfully predicted samples based on this criterion (refer to \cite{he2023grec} for detailed explanations). Additionally, ``\textit{N-acc}" (\textit{No-target accuracy}) evaluates the model's proficiency in no-target grounding scenarios. In this case, predictions without any bounding boxes are considered TP; otherwise, they are regarded as FN. Therefore, ``\textit{N-acc}" is defined as \textit{N-acc.=}$\frac{TP}{TP+FN}$. Subsequent researchers are expected to explore more reasonable evaluation criteria.


\begin{figure}[!t]
        \vspace{5pt}
	\centering
	\includegraphics[width=0.7\linewidth]{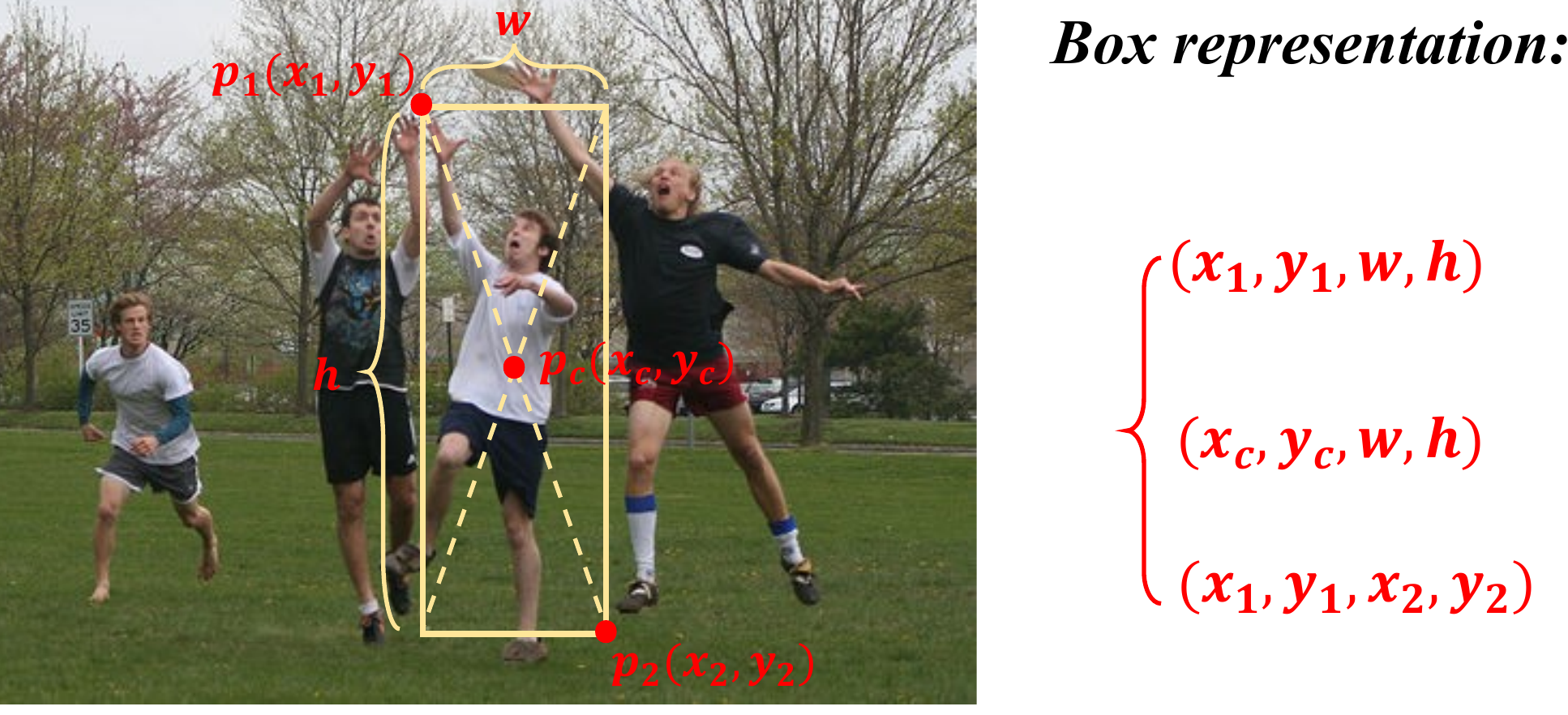}
        \vspace{-7pt}
        \caption{The representations of the bounding box in grounding.}
        \vspace{-10pt}
	\label{fig:box}
\end{figure}

\vspace{-3pt}
\subsection{Representation of the Grounding Box}
\label{subsec:representation_of_box}

The representation of grounding boxes in dataset storage, data preprocessing, and model result output exhibits significant variations. As depicted in \cref{fig:box}, multiple representations are commonly employed, including $(x_{1},y_{1},w,h)$, $(x_{c},y_{c},w,h)$, and $(x_1,y_1,x_2,y_2)$ formats. The prevailing approach for representing the output box is often through the normalized $(x_1,y_1,x_2,y_2)$ format, \ie, $\mathcal{B}_{norm}=(x_1/W,y_1/H,x_2/W,y_2/H)$.

\begin{figure*}[t!]
	\centering
	\includegraphics[width=1.0\linewidth]{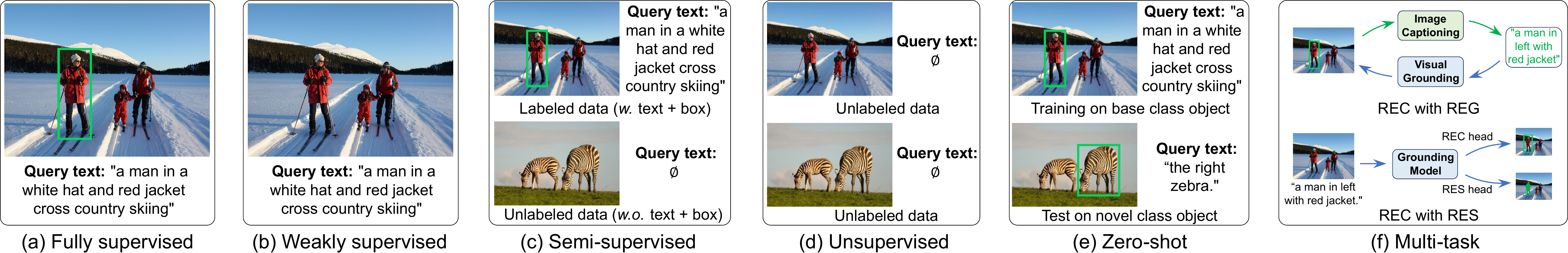}
        \vspace{-17pt}
        \caption{Mainstream settings in visual grounding. Specific definitions of each setting are provided in \cref{sec:method_survey}.}
        \vspace{-12pt}
	\label{fig:setting}
\end{figure*}

In addition, the output of the grounding coordinates is a highly regarded technique, encompassing various position paradigms. The early anchor-based method (\eg, Fast R-CNN-based methods \cite{yang2019fast}) utilizes a predefined sliding window and candidate regions for classification, selecting the proposal with the highest similarity to output the grounding coordinates. Conversely, the current end-to-end approach (\eg, TransVG\cite{deng2021transvg}, \etc) directly regresses the bounding box coordinates using four numerical values. Pix2seq \cite{chen2022pix2seq} treats detection as a sequence generation task by representing spatial positions in discrete bins and utilizing an equal number of tokens for representation, enabling auto-regressive output generation. Building upon this concept, several studies (\eg, OFA \cite{wang2022ofa}, Unified-IO \cite{lu2022unified-io}, UniTAB \cite{yang2022unitab}, GIT \cite{wang2022git}, VisionLLM \cite{wang2023visionllm}, \etc) introduce similar coordinate vocabularies to unify grounding and generation tasks. Furthermore, current MLLM-based methods (\eg, Ferret \cite{you2023ferret}, Shikra \cite{chen2023shikra}, \etc) consider treating coordinate numbers as textual vocabularies.

\subsection{Related Research Domains}
\label{subsec:realted_domain}

The field of visual grounding encompasses several interconnected research domains, for which we will provide a concise overview.

\noindent $\bullet$ \textit{\textbf{Referring Expression Generation (REG).}} REG \cite{winograd1972understanding, krahmer2012computational} is the most closely related task, with its influence deeply ingrained in the development of visual grounding as highlighted in \cref{sec:introduction}. Initially, visual grounding served as an auxiliary task for REG. However, recent years have witnessed a shift towards utilizing REG for generating pseudo-labels and implementing cycle consistency training to facilitate advancements in visual grounding research, which will be discussed in \cref{subsec:rec_and_reg}.

%
%



\noindent$\bullet$ \textit{\textbf{Referring Expression Segmentation (RES).}} RES \cite{hu2016segmentation}, also known as Referring Image Segmentation (RIS), distinguishes itself from REC by necessitating a more intricate and irregular mask area instead of a regular rectangular box. In certain contexts, REC and RES are collectively discussed, while the concurrent implementation of both is termed as multi-task visual grounding, which will be discussed in \cref{subsec:rec_and_res}. However, due to the need for finer-grained regions, extensive research on RES \cite{ji2024survey} has been conducted independently from REC.

\section{Methods: A Survey}
\label{sec:method_survey}

\indent \textbf{\textit{Overview:}} To better facilitate the understanding of the current research status of grounding, in this section, we systematically classify and review existing methods according to their experimental settings, with particular emphasis on those developed within the past five years. \cref{fig:setting} illustrates a concise definition of the commonly used settings. These settings pertain to the types of data or learning approaches employed during model training. Specifically:

\noindent$\bullet$ \textit{\textbf{Fully Supervised Setting.}} This setting involves training or fine-tuning the grounding model using triplets, which consist of data pairs (\ie, image, query text) along with corresponding grounding boxes. It is currently one of the most extensively studied settings.

\noindent$\bullet$ \textit{\textbf{Weakly Supervised Setting.}} As shown in \cref{fig:setting}-(b), in this setting, the grounding model is trained using only image-query text pairs without explicit grounding box annotations. This approach is typically complemented by the use of additional detectors.


\noindent$\bullet$ \textit{\textbf{Semi-supervised Setting.}} As shown in \cref{fig:setting}-(c), the semi-supervised setting refers to utilizing complete labeled triplet data and incomplete image-only data during the training process.


\noindent$\bullet$ \textit{\textbf{Unsupervised Setting.}}  As shown in \cref{fig:setting}-(d), unsupervised grounding is learned solely from unlabeled images while leveraging assisted models such as detectors.

\noindent$\bullet$ \textit{\textbf{Zero-shot Setting.}} There are two typical branches in the zero-shot setting. \textit{(i)} The first branch involves learning grounding ability in the base class and testing its performance in the novel class \cite{sadhu2019zsgnet}. \textit{(ii)} The second branch refers to using pre-trained models from other tasks, particularly pre-training tasks, to evaluate the grounding ability without specific fine-tuning \cite{subramanian2022reclip}. 

\noindent$\bullet$ \textit{\textbf{Multi-task Setting.}} This configuration encompasses various forms where grounding is learned concurrently with other downstream tasks like REG or RES \etc.

\noindent$\bullet$ \textit{\textbf{Generalized Visual Grounding.}} GVG is the newly curated concept as introduced in \cref{subsec:definition} and \cref{fig:definition}.

In the following sections, we will detail each of these settings.

\subsection{Fully Supervised Setting}
\label{subsec:fully_sup}

The Fully Supervised Visual Grounding (FSVG) is currently the most extensively researched domain, which has undergone a decade of development and witnessed the emergence of numerous branches. In this section, we will delve into the technical road map, the classification of the framework architecture, and the benchmark results under four subdivision settings.

\subsubsection{The Technical Roadmap}
\label{subsubsec:road_map}

As depicted in \cref{fig:timeline}, the advancement of visual grounding is intricately linked to the progression of deep learning algorithms and exhibits significant paradigm-shifting stages. We categorize the predominant approaches into five technical routes, namely traditional CNN-based methods, Transformer-based methods, Visual Language Pre-training (VLP)-based methods, and multimodal large language model-based methods.

\begin{figure}[!t]
	\centering
	\includegraphics[width=1.0\linewidth]{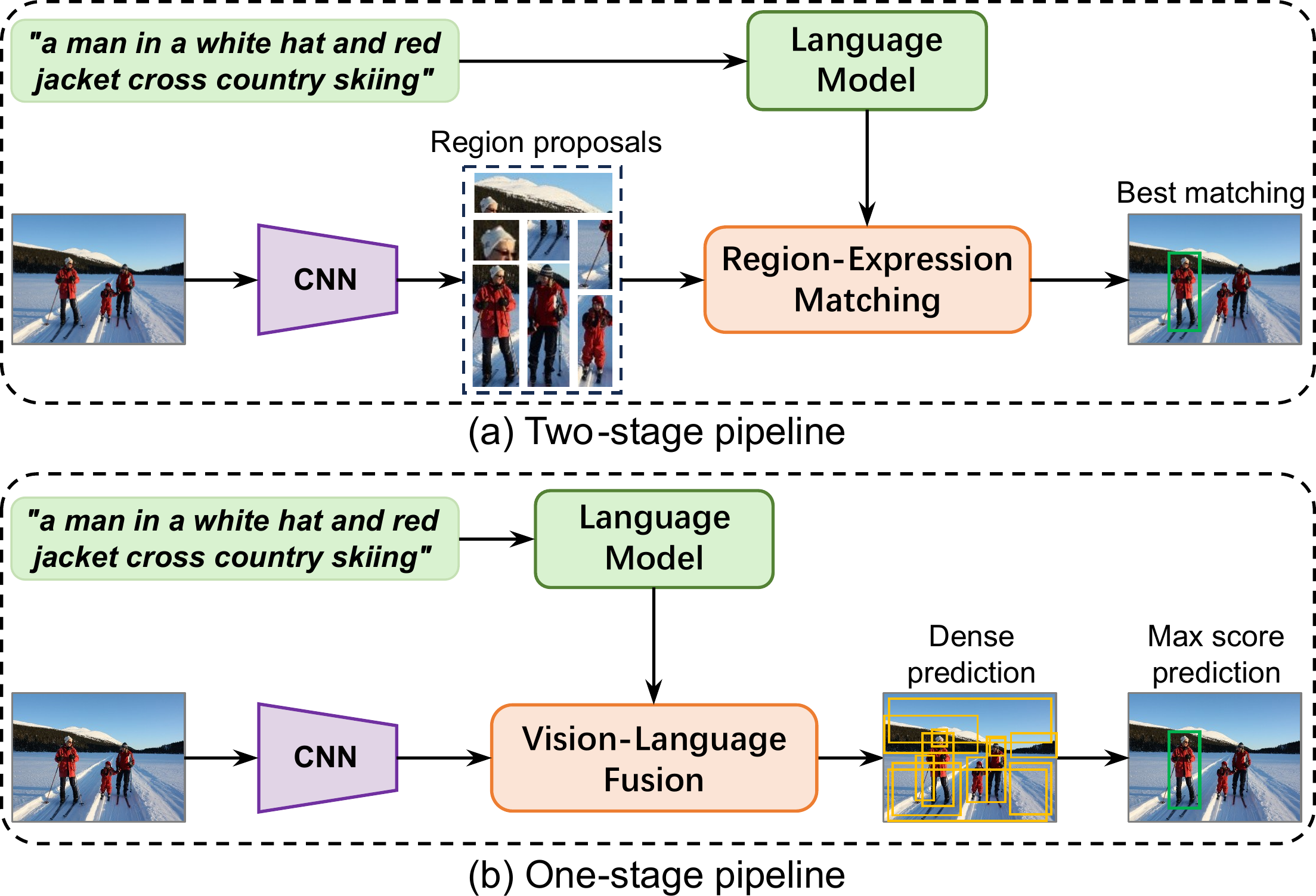}
        \vspace{-19pt}
        \caption{A comparison of two-stage and one-stage pipeline.}
        \vspace{-12pt}
	\label{fig:one_or_two_stage}
\end{figure}

\vspace{4pt}
\noindent \textit{\textbf{A. Traditional CNN-based Methods (From 2014)}}

Intuitively, the initial step in visual grounding involves encoding both the image and referring expression text into a shared vector space, followed by identifying the corresponding visual region based on linguistic cues. In the early stages, CNNs \cite{he2016deep, girshick2015fast} have been dominant for processing images. By embedding input images into fixed-length vectors, CNNs can generate comprehensive image representations suitable for various visual tasks, such as object detection \cite{ren2015faster, liu2016ssd, liu2023foregroundness} and image classification \cite{Simonyan15vgg}. Similarly, Recurrent Neural Networks (RNNs) such as Gated Recurrent Unit (GRU) \cite{chung2014empirical} and LSTM \cite{hochreiter1997long} are commonly employed to encode sentences and exhibit commendable performance in sequence modeling tasks. As encoding techniques continue to advance for the modalities, visual grounding also demonstrates two clear trends of technical evolution.

\textit{\textbf{(a) From two-stage to one-stage.}} In the vision branch, as shown in \cref{fig:one_or_two_stage}, influenced by advancements in object detectors, this period's methods can be typically categorized into two categories. \textbf{\textit{(i) Two-stage methods.}} Due to the limitations imposed by early detectors' technologies such as non-maximum suppression (NMS) \cite{neubeck2006nms}, RoI pooling \cite{yang2019fast, ren2015faster} \etc, the two-stage method initially generates a set of region proposals and subsequently employs region-text matching to identify the proposal with the highest confidence. As depicted in \cref{tab:two_and_one_stage_result}, a substantial number of two-stage methods emerged during this period. Simultaneously, the phrase grounding exhibits similar paradigm (\eg, MCB'16 \cite{fukui2016mcb}, Sim Net'18 \cite{wang2018learning}, CITE'18 \cite{plummer2018cite}, DDPN'18 \cite{yu2018rethinking}, PIRC'19 \cite{kovvuri2019pirc}, CMCC'20 \cite{liu2020cmcc}, \etc). However, these approaches encounter several significant challenges. \textit{Firstly}, generating dense proposals in the first stage necessitates substantial additional computation, thereby degrading computational efficiency. \textit{Secondly}, the final grounding performance is directly influenced by the quality of region proposals obtained in the first stage. \textit{Lastly}, integrating language-guided information into extracted proposals proves challenging. Consequently, with the introduction of the single-stage detectors (\eg, YOLO \cite{redmon2016yolo}, SSD \cite{liu2016ssd}) and end-to-end detectors (\eg, YOLOv3 (DarkNet) \cite{redmon2018yolov3}, Faster-RCNN \cite{ren2015faster}), subsequent research has gradually shifted towards a one-stage approach. 
\textbf{\textit{(ii) The One-stage methods}} eliminates the need for proposal extraction and integrates language information within an intermediate layer of the object detector while outputting the box with a maximum score from pre-defined dense anchors. The representative methods of this period are summarized in \cref{tab:two_and_one_stage_result}.


\begin{table}[t!]
  \setlength\tabcolsep{5pt}
    \footnotesize
    \caption{Summary of one-stage and two-stage methods during the early stage. The results are derived from base model.}
    \vspace{-8pt}
    \centering
    \resizebox{1.0\columnwidth}{!}{%
    \begin{tabular}{c|c|c|c|ccc}
    \toprule
    \multirow{2}[2]{*}{Methods} & \multirow{2}[2]{*}{Venue} & Visual   & Language  & \multicolumn{3}{c}{RefCOCO} \\
                                &                           & branch  &  branch  &   val  & testA & testB       \\
    \midrule    
    \multicolumn{7}{l}{\textbf{\textit{a. Two-stage methods.}}}        \\
    MMI \cite{refcocog-google}   &  CVPR'16   &  VGG16\cite{Simonyan15vgg}  & LSTM\cite{greff2016lstm}   &  --   &  64.90  &  54.51  \\
    Neg Bag\cite{refcocog-umd}   &  ECCV'16   &  VGG16\cite{Simonyan15vgg}  & LSTM\cite{greff2016lstm}   &  --   &  58.60  &  56.40  \\
    Visdif \cite{yu2016modeling}   &  ECCV'16   &  VGG16\cite{Simonyan15vgg}  & LSTM\cite{greff2016lstm}   &  --   &  67.64  &  55.16  \\
    Attr \cite{liu2017attribute}   &  ICCV'17 &  VGG16\cite{Simonyan15vgg}  & LSTM\cite{greff2016lstm}   &  --  &  72.08  & 57.29   \\
    CG \cite{Luo_Shakhnarovich_2017} &  CVPR'17  &  VGG16\cite{Simonyan15vgg}  & LSTM\cite{greff2016lstm} &  --  &  67.94  &  55.18  \\
    CMN \cite{hu2017modeling}    & CVPR'17    &  VGG16\cite{Simonyan15vgg}  & LSTM\cite{greff2016lstm}   &  --   &  71.03  & 65.77   \\
    SLR \cite{yu2017joint}       &  CVPR'17   &  VGG16\cite{Simonyan15vgg}  & LSTM\cite{greff2016lstm}   & 69.48 &  73.71  & 64.96   \\
    PLAN \cite{zhuang2018parallel} &  CVPR'18 &  VGG16\cite{Simonyan15vgg}  & LSTM\cite{greff2016lstm}   &  --   &  75.31  &  65.22  \\
    VC \cite{zhang2018grounding}   &  CVPR'18 &  VGG16\cite{Simonyan15vgg}  &  BiLSTM\cite{schuster1997bilstm}    &  --  & 73.33  &  67.44  \\
    LGRANs\cite{wang2019neighbourhood} &  CVPR'19  &  VGG16\cite{Simonyan15vgg}   & BiLSTM\cite{schuster1997bilstm}  &  --  & 76.60 & 66.40   \\
    MAttNet \cite{yu2018mattnet} & CVPR'18  & Faster RCNN   &  BiLSTM\cite{schuster1997bilstm}   &  76.40  & 80.43   & 69.28   \\
    DGA\cite{yang2019dynamic}   & ICCV'19    & Faster RCNN   &  BiLSTM\cite{schuster1997bilstm} &  --   & 78.42   & 65.53  \\
    NMTree \cite{liu2019learning}  & ICCV'19   & Faster RCNN   &  BiLSTM\cite{schuster1997bilstm}  & 71.65   &  74.81  & 67.34   \\
    RVGTree \cite{hong2019rvg-tree} & TPAMI'19    &  Faster RCNN   &  BiLSTM\cite{schuster1997bilstm} &  71.59  & 76.05   &  68.03  \\
    CM-Att-E \cite{cm-att-erase}    & CVPR'19    &  Faster RCNN   &  BiLSTM\cite{schuster1997bilstm} &  78.35  & 83.14   &  71.32  \\
    
    \midrule    
    \multicolumn{7}{l}{\textbf{\textit{a. One-stage methods.}}}        \\
    SSG \cite{chen2018real}  & ArXiv'18   & YOLOv3 \cite{redmon2018yolov3}  & BiLSTM\cite{schuster1997bilstm}    & --   & 76.51   &  67.50   \\
    FAOA \cite{yang2019fast} & ICCV'19    & YOLOv3 \cite{redmon2018yolov3}   & BERT \cite{devlin2019bert}  & 72.05  & 74.81   & 67.59  \\
    RCCF \cite{liao2020rccf} & CVPR'20    & DLA-34\cite{yu2018dla}   & BiLSTM\cite{schuster1997bilstm}     &  --  & 81.06   &  71.85   \\
    ReSC \cite{yang2020resc} & ECCV'20    & DarkNet \cite{redmon2018yolov3}   & BERT \cite{devlin2019bert}  & 76.59  & 78.22   & 73.25   \\
    MCN \cite{luo2020mcn} & CVPR'20    & DarkNet \cite{redmon2018yolov3}   &  BiGRU \cite{chung2014bigru} & 80.08   & 82.29   &  74.98  \\
    RealGIN \cite{zhou2021real}   & TNNLS'21  & DarkNet \cite{redmon2018yolov3}   &  BiGRU \cite{chung2014bigru} & 77.25  & 78.70 &  72.10  \\
    LBYL \cite{huang2021lbyl}  & CVPR'21    & DarkNet \cite{redmon2018yolov3}   &  BERT \cite{devlin2019bert}  &  79.67  & 82.91   & 74.15   \\
    \bottomrule
    \end{tabular}%
    }
    \label{tab:two_and_one_stage_result}%
    \vspace{-10pt}    
\end{table}%



\textit{\textbf{(b) From GRU/LSTM to attention mechanism.}} In the language branch, as well as the cross-modal fusion branch also shows a clear technical shift. 
\textbf{\textit{(i) CNN-GRU/LSTM period.}} As the original approaches, the majority of two-stage methods (\eg, MMI \cite{refcocog-google}, Visdif \cite{yu2016modeling}, VC \cite{zhang2018grounding}, SCRC \cite{hu2016natural}, CG \cite{Luo_Shakhnarovich_2017}, Attribute \cite{liu2017attribute}, SLR \cite{yu2017joint}, \etc) adopt the CNN-LSTM framework due to its simplicity and effectiveness. However, these methods are constrained by a singular vector representation and overlook the intricate contextual structures present in both languages and images. When dealing with complex query text, they encode it sequentially while disregarding semantic dependencies within the textual expression. 
\textbf{\textit{(ii) CNN-Attention mechanism period.}} The attention mechanism was initially employed in Neural Machine Translation (NMT) in 2014 \cite{mnih2014recurrent, bahdanau2015neural}, followed by the introduction of self-attention in 2016 \cite{cheng2016long}. Subsequently, Multi-Head Self-Attention (MHSA) was proposed within the Transformer framework \cite{transformer}. The effectiveness of utilizing attention mechanisms has been empirically validated for visual and multimodal tasks (\eg, Up-down \cite{anderson2018bottom}, DANs \cite{nam2017dual}, \etc). Consequently, researchers are increasingly applying it to language modules and cross-modal fusion modules in visual grounding (\eg, MCB \cite{fukui2016mcb}, CMN \cite{hu2017modeling}, MattNet \cite{yu2018mattnet}, DGA\cite{yang2019dynamic}, PLAN \cite{zhuang2018parallel}, A-ATT \cite{deng2018visual}, KPRN \cite{liu2019knowledge}, PLAN \cite{zhuang2018parallel}, CM-Att-E \cite{cm-att-erase}, \etc). By employing this technique, token-wise connections can be established between image and language information, facilitating the integration of specific and selective visual and textual features during the encoding process, thereby resulting in semantically enriched cross-modal representations.

\vspace{3pt}
\noindent \textit{\textbf{B. Traditional Transformer-based Methods (From 2021)}}

As mentioned above, the attention mechanism has become an increasingly effective technique in the 2010s \cite{cheng2016long}. The introduction of Transformer \cite{transformer} sparked a revolutionary breakthrough in NLP. In 2018, BERT \cite{devlin2019bert} proposed a self-supervised pre-training paradigm named Next Sentence Prediction (NSP), which enabled the model to learn general language representations. Its success has gradually influenced the Computer Vision (CV) field, leading to the proposal of ViT \cite{dosovitskiy2020image} and DETR \cite{carion2020detr}, which allow Transformer to be used as the visual backbone for grounding tasks. Compared with previous work, a core symbol of research during this period is the use of Transformer as a visual encoding or cross-modal fusion module in grounding frameworks. We summarize these methods from this new era in \cref{tab:full_sota_main} and Appendix Tab. \textcolor{blue}{A1}.

\textit{\textbf{(a) ViT as the vision backbone.}} In 2021, TransVG \cite{deng2021transvg} becomes the pioneering Transformer-based grounding framework to incorporate the encoder from DETR. Since the Transformer architecture no longer requires previous detector-based technologies \cite{ren2015faster} such as RPN, Proposal, NMS, and ROI pooling, realizing grounding in such a framework becomes challenging. TransVG proposes to reformulate visual grounding as a regression problem by utilizing a learnable [\texttt{Region}] token, thereby achieving decoupling from traditional detection tasks. 

\textit{\textbf{(b) Language-guided visual grounding.}} The vision backbone is typically pre-trained on detection and segmentation tasks. Therefore, during grounding learning, additional fusion modules are often required to integrate visual and linguistic features. Such architectural design intuitively reveals a potential flaw: the local visual information may be treated independently during encoding, leading to possible loss of information and resulting in irrelevant visual features for referring text. To address this issue, researchers propose numerous language-guided visual grounding techniques \cite{chen2017query}, such as QRNet \cite{qrnet}, language-conditioned adapter \cite{transvg++}, language prompt \cite{transvg++}, multi-layer adaptive cross-modal bridge (MACB) \cite{xiao2024hivg}, language adaptive dynamic subnet (LADS) \cite{su2023lads}, adaptive weight generation (VG-LAW) \cite{vg-law}, cross-modal attention \cite{liu2023dqdetr}, multimodal conditional adaptation (MMCA) \cite{yao2024mmca} \etc.

\begin{table}[t!]
  \setlength\tabcolsep{3pt}
    \footnotesize
    \caption{A comparison of selected representative work (Base version) during the surge stage under the fully supervised setting (\cref{subsubsec:benchmark_result}). \textbf{The full table is provided in Appendix Table \textcolor{blue}{A1}.}}
    \vspace{-8pt}
    \centering
    \resizebox{1.0\columnwidth}{!}{%
    \begin{tabular}{c|c|c|ccc}
    \toprule
    \multirow{2}[2]{*}{Methods} & \multirow{2}[2]{*}{Venue} & Visual   / Language  & \multicolumn{3}{c}{RefCOCO} \\
                                &                           & branch   branch  &   val  & testA & testB       \\
    \midrule    
    \multicolumn{6}{l}{\textbf{\textit{a. Single-dataset fine-tuning w. unimodal pre-trained close-set detector.}}}        \\
    TransVG \cite{deng2021transvg}  & ICCV'21    &  RN101+DETR   /  BERT-B   &  81.02  &  82.72  & 78.35   \\
    RefTR \cite{li2021referring} &  NeurIPS'21    & RN101+DETR   /  BERT-B  &  82.23  & 85.59   &  76.57  \\
    QRNet \cite{qrnet}  & CVPR'22    & Swin-S \cite{liu2021swin}   / BERT-B    & 84.01   &  85.85  & 82.34   \\
    VG-LAW \cite{vg-law} & CVPR'23    & ViT-Det \cite{vit-det} / BERT-B  & 86.06  & 88.56  & 82.87  \\
    TransVG++ \cite{transvg++} & TPAMI'23    &  ViT-Det \cite{vit-det} / BERT-B  & 86.28  & 88.37  & 80.97  \\
    \midrule    
    \multicolumn{6}{l}{\textbf{\textit{b. Single-dataset fine-tuning setting w. self-supervised VLP models.}}}        \\
    CLIP-VG \cite{xiao2023clip}   &   TMM'23      & CLIP-B / CLIP-B   &  84.29  & 87.76  & 78.43  \\
    D-MDETR \cite{shi2023dynamic} & TPAMI'23    & CLIP-B  / CLIP-B   & 85.97  & 88.82  & 80.12  \\
    HiVG \cite{xiao2024hivg}  &   ACMMM'24      & CLIP-B / CLIP-B   &   87.32    &   89.86    &   83.27  \\
    OneRef  \cite{xiao2024oneref}    &   NeurIPS'24    & BEiT3-B  /  BEiT3-B   &   88.75    &   90.95    &   85.34  \\
    \midrule    
    \multicolumn{6}{l}{\textbf{\textit{c. Dataset-mixed intermediate pre-training setting.}}}        \\
    MDETR$^\dagger$ \cite{kamath2021mdetr}     & ICCV'21   & RN101+DETR  / RoBERT-B    & 86.75  & 89.58  & 81.41  \\
    G-DINO-B$^\dagger$  \cite{liu2023grounding}   & ECCV'24  & Swin-T   /  BERT-B   &  89.19  & 91.86  & 85.99  \\
    HiVG-B* \cite{xiao2024hivg}        & ACMMM'24  & CLIP-B   /  CLIP-B     & 90.56  & 92.55  & 87.23 \\
    OneRef-B* \cite{xiao2024oneref}   & NeurIPS'24  & BEiT3-B /  BEiT3-B    &   91.89 &  94.31 &  88.58 \\
    OFA-B$^\ddagger$ \cite{wang2022ofa}         & ICML'22   & OFA-B   / OFA-B     & 88.48  & 90.67  & 83.30 \\
    CyCo$^\ddagger$ \cite{wang2024cycle}      & AAAI'24   & ViT\cite{dosovitskiy2020image} / BERT-B & 89.47  & 91.87  & 85.33  \\
    \rowcolor{red!07}
    \multicolumn{3}{l|}{~~~\textit{\textbf{Predicted ultimate performance (based on OneRef-B):}}} &  98.69  &  99.08 & 98.57   \\
    \midrule    
    \multicolumn{6}{l}{\textbf{\textit{d. Fine-tuning setting w. grounding multimodal LLMs. (GMLLMs)}}}        \\
    Shikra-7B \cite{chen2023shikra}   &   arXiv'23     & CLIP-L / Vicuna-7B\cite{chiang2023vicuna}   & 87.01 & 90.61 & 80.24  \\
    Ferret-7B \cite{you2023ferret}  &   ICLR'24     & CLIP-L / Vicuna-7B\cite{chiang2023vicuna}  & 87.49 & 91.35 & 82.45   \\
    G-GPT \cite{li2024groundinggpt}  &   ACL'24     & CLIP-L / Vicuna-7B\cite{chiang2023vicuna}   & 88.02   & 91.55 & 82.47   \\
    LION-4B \cite{chen2024lion}  &   CVPR'24     & EVA-G\cite{fang2023eva}/FlanT5-3B   & 89.73 & 92.29 & 84.82  \\
    \bottomrule
    \end{tabular}%
    }
    \label{tab:full_sota_main}%
    \vspace{-10pt}    
\end{table}%

\vspace{4pt}
\noindent \textit{\textbf{C. VLP-based Transfer Methods (From 2021)}}


Under the traditional paradigm, the two modalities of visual grounding are encoded separately by backbones based on detection and language tasks \cite{deng2021transvg}. The features learned by such backbone networks are naturally not aligned across modalities, resulting in a significant gap in the fusion of visual and language representations. In 2021, Radford \etal \cite{radford2021clip} proposed to utilize self-supervised Contrastive Language Image Pre-training (CLIP) to train on large-scale web image-text data pairs. CLIP achieves comparable performance on image classification tasks in a zero-shot setting compared to previous fully supervised methods, thereby unleashing a surge of multimodal pre-training \cite{peng2023sgva}. By leveraging VLP models for grounding tasks, there exists a natural alignment within the cross-modal feature space. Consequently, CLIP-VG \cite{xiao2023clip} adopts a simple architecture comprising two encoders with a fusion encoder and utilizes multiple layers of visual features to facilitate grounding perception. Although the VLP model exhibits certain advantages in realizing grounding transfer, the grounding task necessitates both region-level image perception and semantic understanding of textual logic \cite{chen2025interpreting}. Therefore, it still possesses several limitations that subsequent research endeavors aim to address.

Simultaneously, the VLP model has acquired a comprehensive cross-modal representation from extensive data, which makes it susceptible to catastrophic forgetting if directly fine-tuned with full parameters on small-scale downstream tasks \cite{xiao2024hivg}. Hence, Parameter-Efficient Fine-Tuning (PEFT) \cite{ding2023peft} techniques like LoRA \cite{hu2021lora}, Prompt \cite{jia2022prompt, li2024prompt}, Adapter \cite{gao2024clip-adapter}, \etc, also play a crucial role in facilitating the VLP-based grounding transfer. HiVG \cite{xiao2024hivg} establishes connections between multi-level visual and linguistic features through layer-specific weights and a multi-layer adaptive cross-modal bridge. It introduces a hierarchical low-rank adaptation (HiLoRA) paradigm to hierarchically modulate the visual features of the vanilla CLIP model for achieving SoTA grounding performance. Additionally, several other approaches, such as CRIS \cite{wang2022cris}, RISCLIP \cite{kim2023risclip} \etc, have been proposed to implement RES task transfer based on CLIP.


\vspace{+4pt}
\noindent \textit{\textbf{D. Grounding-oriented Pre-training Methods (From 2020)}}

\textit{\textbf{(a) Region-level grounded pre-training.}} The task of visual grounding is inherently intertwined with detection. However, traditional detection tasks do not encompass the encoding and comprehension of language modalities. Therefore, in 2021, MDETR \cite{kamath2021mdetr} proposed to reformulate the detection task as a modulated detector based on the encoder-decoder architecture of the DETR, thereby achieving the integration of detection and grounding. Inspired by region-phrase correspondences constructed in phrase localization, Li \etal \cite{li2022glip} propose Grounded Language-Image Pre-training (GLIP) to obtain region-level fine-grained cross-modal representation. These serve as the foundation for subsequent multimodal large-scale pre-trained detection models (\eg, GLIPv2 \cite{zhang2022glipv2}), open vocabulary detection frameworks \cite{zareian2021ovrcnn}, \etc. Building upon this progress, MDETR also introduces an experimental branch that diverges from the single-dataset fine-tuning setting, namely the dataset-mixed intermediate pre-training setting, as illustrated in \cref{tab:full_sota_main}. Leveraging the concept of grounded pre-training, Grounding-DINO \cite{liu2023grounding} successfully achieves a unified model for open-set detection and visual grounding. Following this line of research, DINO-X \cite{ren2024dino-x} constructed the Grounding-100M dataset to support large-scale training for grounding-related tasks.

\textit{\textbf{(b) Multi-task pre-training.}} Following the concept of region-level multimodal pre-training, as a fine-grained cross-modal understanding task, visual grounding can also be pre-trained in a multi-task paradigm alongside related tasks (\eg, image captioning, VQA, retrieval, \etc) to obtain more general representations. Representative methods include UniTAB \cite{yang2022unitab}, OFA \cite{wang2022ofa}, UNINEXT \cite{wang2024hipie}, HIPE \cite{yan2023uninext}, ViLBERT \cite{lu2019vilbert}, VL-BERT \cite{su2020vl-bert}, ONE-PEACE \cite{wang2023one-peace}, mPlug \cite{li2022mplug}, \etc. These methods typically employ multi-task learning to capture general cross-modal knowledge, thereby achieving strong region-level understanding capabilities with only limited grounding data. For a detailed overview of this topic, please refer to the survey papers \cite{gan2022vision, wang2023large}.

\begin{figure*}[!t]
	\centering
	\includegraphics[width=0.99\linewidth]{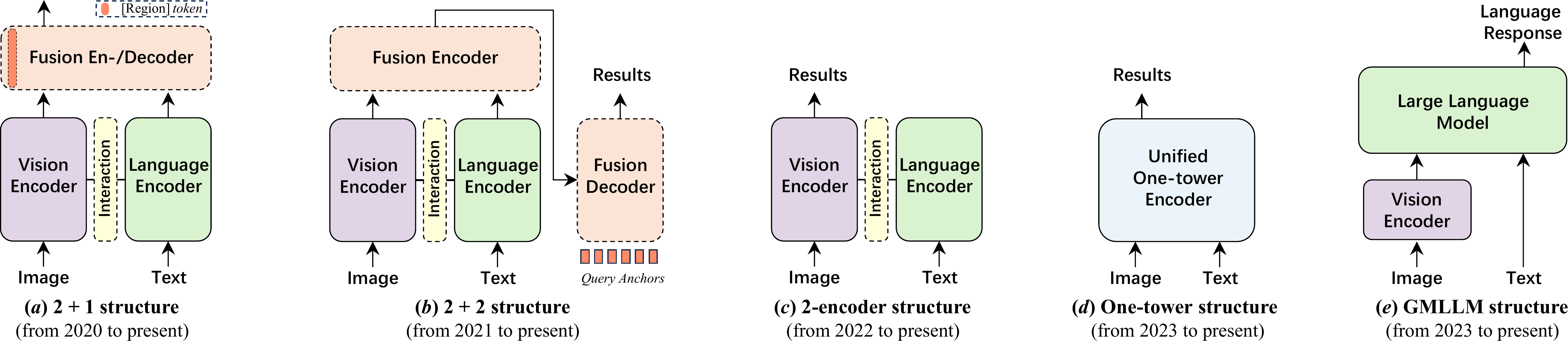}
        \vspace{-4pt}
        \caption{Classification of typical framework architectures for visual grounding when using pre-trained models.}
        \vspace{-12pt}
	\label{fig:model_arch}
\end{figure*}

\vspace{4pt}
\noindent \textit{\textbf{E. Grounding Multimodal LLMs (From 2023)}}

\textbf{\textit{(a) Motivations.}} According to the definition of visual grounding, traditional detector-based and VLP-based approaches encounter several challenges. \textit{Firstly}, most existing grounding methods only adhere to a narrow definition of visual grounding due to their reliance on fixed box regression heads, making it challenging to achieve a generalized grounding (as in \cref{subsec:definition}). \textit{Secondly}, as an open-world setting, visual grounding should support arbitrary language queries; yet conventional approaches are limited by fixed training and testing sets (\eg, RefCOCO/+/g dataset), restricting both object categories and textual content. \textit{Thirdly}, referring and grounding are commonly employed in dialogue scenarios; however, traditional methods perform grounding within only one round. In other words, the current models lack the ability to engage in natural language dialogues while performing grounding. 

With the introduction of OpenAI's GPT-3 and ChatGPT \cite{gpt3, gpt-2, ouyang2022gpt}, LLMs have demonstrated strong AI capabilities through large-scale pre-training. Subsequently, with the release of GPT-4 \cite{achiam2023gpt-4}, general multimodal AI became feasible. In 2023, Meta AI open-sourced an LLM named LLaMA-13B \cite{touvron2023llama}, which surpassed the commercial GPT-3-175B model on most benchmarks and was competitive with the SoTA LLMs such as Chinchilla-70B \cite{hoffmann2022training} and PaLM-540B \cite{chowdhery2023palm}. During this period, LLMs exhibited remarkable progress. Alpaca \cite{taori2023alpaca}, Vicuna \cite{chiang2023vicuna}, and GPT-4-LLM \cite{peng2023gpt4-llm} utilized various machine-generated high-quality instruction examples to enhance the alignment ability of LLMs and achieved impressive performance compared to proprietary LLMS. Subsequently, Liu \etal introduced LLaVA \cite{liu2024llava} as a robust MLLMs baseline by leveraging LLaMA and CLIP in a visual instruction tuning approach, thereby empowering large language models with multimodal capabilities. Consequently, starting from 2023 onwards, extensive efforts have been devoted towards Grounding Multimodal Large Language Models (GMLLMs).

\textbf{\textit{(b) Research status.}} One distinction in utilizing LLM to address the grounding problem lies in the representation of the output bounding box. Shikra \cite{chen2023shikra} pioneer explored GMLLM and conducted experimental verification, demonstrating its superior performance by directly employing coordinate numbers as a textual vocabulary. While KOSMOS-2 \cite{peng2024kosmos-2} is also an earlier work, it primarily builds upon KOSMOS-1 \cite{huang2023kosmos-1} to validate MLLM's capability for zero-shot and few-shot grounding with the aids of instruction tuning \cite{liu2024llava}. In subsequent research, Ferret \cite{you2023ferret} introduced hybrid region representation and implemented open-vocabulary description grounding at free-form shapes and arbitrary granularity through the construction of spatial-aware visual samplers. Ferret-v2 \cite{zhang2024ferret-v2} incorporates multi-scale DINOv2 \cite{oquab2024dinov2} features, enabling grounding and referring with arbitrary resolution via three-stage training. Similarly, Lava-grounding \cite{zhang2024llava-grounding} adopts a comparable model architecture to Ferret but offers more flexibility in input (\eg, click, box, and mark) and output (\eg, text, box, mask, and mark). To reconcile the internal conflict between region-level and image-level visual and language (VL) tasks, LION \cite{chen2024lion} introduces a progressive integration of fine-grained spatial-aware visual knowledge based on Mixture-of-Adapter's structure \cite{chen2022adaptformer} and BLIP-2's Q-former \cite{li2023blip-2} using a three-stage instruction-tuning strategy. Grounding-GPT \cite{li2024groundinggpt} builds upon ImageBind and Q-former frameworks to implement not only visual grounding but also video grounding and audio grounding. Similarly, related MLLM-based methods (\eg, GLaMM \cite{rasheed2024glamm}, LISA \cite{lai2024lisa}, GSVA \cite{xia2024gsva}, UniMLLM \cite{li2024unifiedmllm}, F-LMM \cite{wu2024f-lmm}, VistaLLM\cite{pramanick2024vistallm}, \etc) have emerged in the RES task. In general, these models adopt a multi-stage training strategy (\eg, three-stage \cite{chen2024lion, ma2024groma, zhang2024ferret-v2}) and follow a relatively similar and simple framework, essentially adopting the paradigm illustrated in \cref{fig:model_arch}-(e). Due to space limitations and the recent emergence of similar work, we present part of the other related method (such as MiniGPT4-v2 \cite{chen2023minigpt-v2}, QWen-VL \cite{bai2023qwen-vl}, Groma \cite{ma2024groma}, Lenna \cite{wei2023lenna},  VisCoT \cite{shao2024viscot}, ViGoR \cite{yan2024vigor}, BuboGPT \cite{zhao2023bubogpt}, MiniGPT4 \cite{zhu2024minigpt4}, RegionGPT\cite{guo2024regiongpt}, VistaLLM\cite{pramanick2024vistallm}, VisionLLM \cite{wang2023visionllm}, CogVLM \cite{wang2023cogvlm}, Next-Chat \cite{zhang2024next-chat}, TextHawk \cite{yu2024texthawk},  \etc) in Tab. \textcolor{blue}{A1}.

\textbf{\textit{(c) Techniques for GMLLMs.}} As a type of MLLM, numerous common techniques can be applied to GMLLMs. In Shikra \cite{chen2023shikra}, the author introduces the concept of Grounding Chain-of-Thought (GCOT), which extends the traditional Chain-of-Thought (COT) \cite{li2024multimodal} by incorporating referring reasoning capabilities. For tasks of varying granularities, such as image-level and region-level tasks, LION \cite{chen2024lion} employs a Mix-of-Adapters (MOA) mechanism with a router in the frozen LLM to dynamically integrate visual knowledge acquired from visual branches and LLM adapters. Similarly, LoRA \cite{hu2021lora} has proven effective in models like Next-Chat \cite{zhang2024next-chat} and LISA \cite{lai2024lisa}. Reinforcement learning is also employed to enhance the model's perception of the cross-modal referring reasoning process \cite{chen2025visrl}.

\textbf{\textit{(d) Datasets and benchmarks for GMLLMs.}} A number of current methods present datasets specific to GMLLMs, and we introduce the relevant datasets in Appendix Sec. \textcolor{blue}{3.2}.

\begin{table}[t!]
    \vspace{5pt}
  \setlength\tabcolsep{3pt}
    \footnotesize
    \caption{Part of exemplar work for the five typical structures.}
    \vspace{-5pt}
    \centering
    \resizebox{1.0\columnwidth}{!}{%
    \begin{tabular}{l|llll}
    \toprule
    \textbf{Architectures} &  \multicolumn{4}{c}{Representative grounding work} \\
    \midrule    
    \multirow{2}[1]{*}{\textit{\textbf{2+1 structure}}}  &  TransVG \cite{deng2021transvg}   &  QRNet\cite{qrnet}                   &  CLIP-VG \cite{xiao2023clip}    & UNITER \cite{chen2020uniter}      \\
                                                         &  HiVG \cite{xiao2024hivg}         &  MMCA \cite{yao2024mmca}             &  ReSC \cite{yang2020resc}       & CRIS \cite{wang2022cris}          \\
    \midrule    
    \multirow{2}[1]{*}{\textit{\textbf{2+2 structure}}}  &  MDETR \cite{kamath2021mdetr}     &  DQ-DETR \cite{liu2023dqdetr}        &  D-MDETR \cite{shi2023dynamic}  & G-DINO \cite{liu2023grounding}    \\
                                                         &  RefTR \cite{li2021referring}     &  PolyFormer \cite{liu2023polyformer} &  UniTAB \cite{yang2022unitab}   & OFA \cite{wang2022ofa}            \\
    \midrule    
    \multirow{1}[1]{*}{\textit{\textbf{Two-encoder}}}    &  TransVG++ \cite{transvg++}       &  VG-LAW \cite{vg-law}                & UniQRNet \cite{ye24uniqrnet}    & LAVT \cite{yang2022lavt}          \\
    \midrule    
    \multirow{1}[1]{*}{\textit{\textbf{One-tower}}}      &  OneRef \cite{xiao2024oneref}     &  ONE-PEACE \cite{wang2023one-peace}  &  YORO\cite{ho2023yoro}          & SimVG \cite{dai2024simvg}         \\
    \midrule    
    \multirow{3}[1]{*}{\textit{\textbf{GMLLMs}}}    &  Ferret \cite{you2023ferret}      &  MiniGPT-v2 \cite{chen2023minigpt-v2} & Ferret-v2 \cite{zhang2024ferret-v2}  & Shikra \cite{chen2023shikra}   \\  
            &  G-GPT \cite{li2024groundinggpt}   &  LLaVA-G \cite{zhang2024llava-grounding}  &  QWen-VL \cite{bai2023qwen-vl}  &  Groma \cite{ma2024groma}       \\
            &  Kosmos-2 \cite{peng2024kosmos-2}  &  VisCoT \cite{shao2024viscot}             &  Lenna \cite{wei2023lenna}      &  LION \cite{chen2024lion}       \\
            &  u-LLaVA \cite{xu2023u-llava}      &  VistaLLM \cite{pramanick2024vistallm}    &  Next-Chat \cite{zhang2024next-chat}  &  LISA \cite{lai2024lisa}     \\
    \bottomrule
    \end{tabular}%
    }
    \label{tab:model_arch}%
    \vspace{-5pt}    
\end{table}%

\subsubsection{The Classification of Framework Architectures}
\label{subsubsec:model_arch}

In the last subsection, we present a review of fully supervised methods over the past decade, with a focus on the advancements in technical roadmaps. Notably, since 2020, the widely adopted paradigm of \textit{``pre-training and fine-tuning"} has gained popularity, leading to rapid developments in visual grounding. We provide an overview of model architectures employed in grounding tasks that employ pre-trained models, which can be categorized into five typical types as illustrated in \cref{fig:model_arch} and \cref{tab:model_arch}.

Specifically, \textbf{\textit{(a) the 2+1 structure}} is represented by TransVG \cite{deng2021transvg}, which primarily employs visual and language-independent encoding and subsequently utilizes a fusion encoder for cross-modal feature fusion. This architecture incorporates a special region token to regression grounding results. \textit{\textbf{(b) The 2+2 structure}}, exemplified by MDETR \cite{kamath2021mdetr}, follows the structure of the original DETR \cite{liu2023dqdetr} model. It integrates query anchors to generate grounding boxes, making it more compatible with detection and segmentation tasks. Due to the separation of modal encoding and feature fusion, this architecture can seamlessly adapt to various pre-training paradigms (\eg, Image-Text Matching (ITM), sequence-to-sequence generation, Masked Language Modeling (MLM), \etc) during the pre-training phase. Consequently, it has been widely adopted in the early research on general representation learning (\eg, FIBER \cite{dou2022fiber}, \etc). However, its drawback lies in the excessive number of parameters and high training cost due to the bulky modules. \textbf{\textit{(c) The two-encoder structure}} addresses the issue of parameter redundancy present in structures (a) and (b). By directly discarding the fusion module, these structures achieve higher efficiency. \textbf{\textit{(d) One-tower structures}} like OneRef \cite{xiao2024oneref} eliminates complex integration designs and redundant parameters by utilizing modality-shared feature spaces, thereby achieving both efficiency and promising performance. Similarly, other work such as YORO \cite{ho2023yoro}, ScanFormer \cite{su2024scanformer}, SimVG \cite{dai2024simvg}, \etc, mainly benefit from the pre-trained representation of the one-tower backbone networks, \ie, ViLT \cite{kim2021vilt} and BEiT-3 \cite{beit3}. Finally, \textbf{\textit{(e) GMLLM structure.}} The current GMLLMs essentially follow the paradigm of this structure, which involves encoding visual information and mapping it into the feature space of LLMs to formulate a grounding task as an auto-regressive language task.

\subsubsection{Benchmark Results}
\label{subsubsec:benchmark_result}

\vspace{3pt}
\noindent \textit{\textbf{A. The Four Subdivision Experimental Setting}}

The performance of the representative work since the 2020s is summarized in \cref{tab:full_sota_main}. To ensure a fair comparison, we categorize the experimental results into four typical settings. Specifically, \textit{\textbf{(a)} single dataset fine-tuning with an unimodal close-set detector and language model; \textbf{(b)} single dataset fine-tuning with self-supervised VLP models; \textbf{(c)} intermediate pre-training with mixed datasets;} and \textit{\textbf{(d)} fine-tuning based on GMLLMs.} Furthermore, the third type of setting can be further subdivided based on the intermediate pre-training paradigm, including \textit{\textbf{(i)} intermediate pre-training based on detection supervision (marked with $\dagger$)}, \textit{\textbf{(ii)} intermediate pre-training based on grounding supervision (marked with $*$)}, and \textit{\textbf{(iii)} intermediate pre-training based on multi-task supervision using box-level fine-grained datasets (marked with $\ddagger$)}. It is worth noting that some current methods lack rigor in their experimental comparisons and have not undergone thorough scrutiny during peer review, leading to an unfavorable environment. It is strongly urged that future research should adopt more rigorous classification methods for experimental settings when conducting comparisons.

\vspace{3pt}
\noindent \textit{\textbf{B. Ultimate Performance Prediction on the Three Datasets}}

Even after a decade of development, the RefCOCO/+/g dataset continues to serve as the fundamental dataset for current grounding research. However, as depicted in \cref{tab:full_sota_main} and \cref{fig:perf_trend}-(b), the performance of these three datasets is currently highly crowded. Based on the findings from Ref-L4 \cite{chen2024ref-l4} and CLIP-VG \cite{xiao2023clip}, both the validation and test sets of the RefCOCO/+/g datasets contain numerous errors and challenging grounding examples. Consequently, it is unlikely that the RefCOCO/+/g dataset will achieve a 100$\%$ performance score. Therefore, we aim to predict the performance boundaries of the RefCOCO/+/g dataset to alert future researchers toward proposing more demanding datasets and altering testing benchmarks for grounding evaluation.

The current trend in grounding involves the utilization of increasingly diverse datasets during intermediate pre-training, which makes these models more susceptible to \textbf{\textit{data leaks}}. Based on this, we employ the SoTA model to retrain a new model by incorporating samples from all training, validation, and test sets of the RefCOCO/+/g dataset. Subsequently, we utilize this model to evaluate the performance of the validation and test sets. By adopting such a self-paced curriculum learning approach \cite{xiao2023clip}, it becomes feasible to provide a rough estimation of the upper-performance limit across these three datasets. Specifically, we conducted experiments using OneRef~\footnote[1]{\url{https://github.com/linhuixiao/OneRef}.} under settings (c), with the results presented in \cref{tab:full_sota_main}. It is evident that the current performance gaps are approximately around $5\%\sim10\%$. This indicates an urgent need to propose a new grounding dataset.

\begin{table}[t!]\footnotesize
\setlength\tabcolsep{3.5pt}
\caption{An overview of the weakly supervised methods.}
\vspace{-13pt}
\begin{center}
\resizebox{1.0\columnwidth}{!}{%
    \begin{tabular}{c|c|c|c|c|c}
    \toprule
    \multirow{2}[1]{*}{Methods}    &  \multirow{2}[1]{*}{Venue}  &  V/L        &  Two/one-  &  Flickr  & RefCOCO+  \\
                                   &                             &  Backbone   &  stage     &  test     & val      \\
        \midrule
    \multicolumn{6}{l}{\textit{\textbf{a. Proposal-based Methods.}}}    \\
    GroundR \cite{rohrbach2016grounding}           & ECCV'16    & VGG / LSTM   &  Two  &  28.94  & --     \\
    Xiao \etal \cite{xiao2017weakly}               & CVPR'17    & VGG / LSTM   &  Two  &  N/A    & N/A    \\
    KAC Net \cite{chen2018knowledge}               & CVPR'18    & VGG / LSTM   &  Two  &  38.71  & --     \\
    MATN \cite{zhao2018weakly}                     & CVPR'18    & VGG / LSTM   &  Two  &  33.10  & --     \\
    KPRN \cite{liu2019knowledge}                   & ACMMM'19   & M-R / LSTM   &  Two  &  --     & 35.96  \\
    ARN \cite{liu2019adaptive}                     & ICCV'19    & M-R / LSTM   &  Two  &  37.95  & 32.78  \\
    Align2Ground \cite{datta2019align2ground}      & ICCV'19    & F-R / LSTM   &  Two  &  41.40  & --     \\
    info-ground \cite{gupta2020contrastive}        & ECCV'20    & F-R / BERT   &  Two  &  51.67  & --     \\
    MAF \cite{wang2020maf}                         & EMNLP'20   & F-R / BERT   &  Two  &  44.39  & --     \\
    CCL \cite{zhang2020counterfactual}             & NeurIPS'20 & F-R / BiGRU  &  Two  &  --     & 34.29  \\
    DTWREG \cite{sun2021discriminative}            & TPAMI'21   & F-R / Glove  &  Two  &  --     & 38.91  \\
    NCE-Distill\cite{wang2021improving}            & CVPR'21    & F-R / LSTM   &  Two  &  50.96  & --     \\
    ReIR \cite{liu2021relation}                    & CVPR'21    & F-R / LSTM   &  Two  &  59.27  & --     \\
    EARN \cite{liu2022earn}                        & TPAMI'22   & F-R / LSTM   &  Two  &  38.73  & 37.54  \\
    DRLF \cite{wang2023dual}                       & TMM'23     & F-R / LSTM   &  Two  &  46.46  & --     \\
    Cycle \cite{zhang2023cycle}                    & TIP'23     & F-R / GRU    &  Two  &  64.88  & 37.66  \\
    TGKD \cite{mi2023weakly}                       & ICRA'23    & F-R / Glove  &  Two  &  --     & 40.20  \\
    PSRN \cite{ji2024progressive}                  & TCSVT'24   & F-R / LSTM   &  Two  &  --     & 40.68  \\
    \midrule
    \multicolumn{6}{l}{\textit{\textbf{b. VLP-based WSVG Transfer.}}}    \\
    ALBEF \cite{li2021albef}                       & NeurIPS'21 & ViT / BERT   &  Two  &  --     & 58.46  \\
    X-VLM \cite{zeng2022x-vlm}                     & ICML'22    & Swin / BERT  &  Two  &  --     & 67.78  \\
    CPL \cite{liu2023confidence}                   & ICCV'23    & VGG / CLIP   &  Two  &  46.62  & --     \\
    g++ \cite{shaharabany2023similarity}           & CVPR'23    & VGG / CLIP   &  Two  &  45.56  & --     \\
    RefCLIP \cite{jin2023refclip}                  & CVPR'23    & YLv3 / CLIP  &  One  &  --     & 40.39  \\
    QueryMatch \cite{chen2024querymatch}           & ACMMM'24   & M-F / CLIP   &  One  &  --     & 44.76  \\
    UR \cite{zhang2024universal}                   & TOMM'24    & F-R / CLIP   &  Two  &  --     & 49.37  \\
    PPT \cite{zhao2024ppt}                         & MMM'24     & F-R/X-VLM  &  Two  &  --     & 68.16  \\
    \bottomrule
    \multicolumn{6}{p{9.6cm}}{\rule{0pt}{9pt}{\small Annotation: `M-R', `F-R', `YLv33', `M-F', and `Glove' represents Mask-RCNN \cite{he2017mask}, Faster-RCNN \cite{ren2015faster}, YOLO-v3 \cite{redmon2018yolov3}, Mask2former \cite{cheng2022masked}, and Glove vector \cite{pennington2014glove}, respectively.}}      \\
    \end{tabular}%
}
\end{center}
\label{tab:weakly_sup}
\vspace{-13pt}	
\end{table}


\subsection{Weakly supervised setting}
\label{subsec:weakly_sup}

As defined in \cref{sec:method_survey}, to reduce the dependence on labor-intensive bounding box annotations in fully supervised settings, Weakly Supervised Visual Grounding (WSVG) aims to learn region-query correspondences solely from image-text pairs. This setting has garnered significant attention over the past decade due to its relatively reduced data dependency. As shown in \cref{tab:weakly_sup}, similar to the evolution of full supervision, we categorize existing WSVG approaches into two groups: traditional proposal-based methods and VLP-based WSVG transfer.

\subsubsection{Proposal-based Methods}

As WSVG lacks ground-truth box annotation for supervision during training, an intuitive idea is to utilize an off-the-shelf detector to generate image proposals. This pipeline bears resemblance to the traditional two-stage fully supervised network, thus inspiring most existing methods from 2016 \cite{rohrbach2016grounding} onwards to frame WSVG as a region-text ranking problem within a Multiple Instance Learning (MIL) \cite{karpathy2015deep} framework. The primary challenge in these methods lies in providing effective supervision signals from image-text pairs. To address this issue, researchers have employed various techniques such as sentence reconstruction, Contrastive Learning (CL), relation-aware instance refinement, pseudo-labeling, and one-stage approaches, \etc.

\textit{\textbf{(a) Sentence reconstruction strategies.}}  The reconstruction strategy typically employs an external object detector to generate a set of region proposals from the image and reconstructs the entire query using the proposal with the highest ranking score, thereby establishing matching and reconstruction losses \cite{rohrbach2016grounding, xiao2017weakly, xiong2023client}. GroundR (2016) \cite{rohrbach2016grounding} constructs correspondence by incorporating a visual feature attention mechanism to reconstruct phrases. To enhance the effectiveness of supervision, KAC-Net (2018) \cite{chen2018knowledge} adopts a similar formulation but integrates knowledge of visual consistency and target categories, while Align2Ground (2019) \cite{datta2019align2ground} employs a ranking loss to minimize the distance between relevant image captions and maximize the distance between irrelevant ones. Subsequently, DTWREG (2021) \cite{sun2021discriminative} introduces a discriminative triad and a scalable query parsing strategy, and PSRN (2024) \cite{ji2024progressive} leverages a progressive semantic reconstruction network with a two-level matching-reconstruction process.

\textit{\textbf{(b) Contrastive learning.}} In contrast to sentence reconstruction, CL-based methods construct pairs of positive and negative samples from selected regions and expressions and then compute the InfoNCE loss \cite{oord2018infonce}. For instance, CCL \cite{zhang2020counterfactual} leverages counterfactual CL to develop sufficient contrastive training between counterfactual positive and negative results. NCE-Distill \cite{wang2021improving} utilizes a contrastive paradigm to optimize word-region attention for learning phrase grounding, thereby maximizing the lower bound of mutual information between images and queries. Other methods, such as info-ground \cite{gupta2020contrastive}, Cycle \cite{zhang2023cycle} \etc, utilize similar contrastive modules to achieve superior performance.

\textit{\textbf{(c) Relation-aware instance refinement.}} The utilization of linguistic sentence structure and scene graph is a natural choice for association and parsing in weakly supervised proposals, enabling the refinement of target regions based on spatial relationships. Specifically, MATN \cite{zhao2018weakly} employs a transformation network to search for target phrase locations across the entire image, which are then regularized using precomputed candidate boxes. Moreover, contextual cues have been considered in some work to disambiguate semantics. For instance, ARN \cite{liu2019adaptive} and EARN \cite{liu2022earn} ensure multi-level cross-modal consistency by extracting linguistic and visual cues at entity, location, and context levels. KPRN \cite{liu2019knowledge} further incorporates linguistic context by simultaneously matching subject and target entities. To address the limitation of semantic ambiguity, ReIR \cite{liu2021relation} adopts a weakly supervised learning strategy that focuses on context-aware instance refinement.

\textit{\textbf{(d) Pseudo-labeling.}} In the unsupervised setting, due to the absence of labeling information, Pseudo-Q \cite{jiang2022pseudo} proposes to construct template-based pseudo-labels by leveraging spatial prior information of image context. Motivated by this approach, some studies (\eg, CPL \cite{liu2023confidence}, Lin \etal \cite{lin2024visual}, \etc) have also incorporated pseudo-labels into WSVG settings. CPL \cite{liu2023confidence} introduces a confidence-aware pseudo-labeling method for directly generating pseudo-queries to address the cross-modal heterogeneous gap in the sentence reconstruction process. g++ \cite{shaharabany2023similarity} employs pseudo-labels and localization maps for self-training purposes. DRLF \cite{wang2023dual} incorporates pseudo-queries generated by Pseudo-Q as a warm-start module within a dual reinforcement learning framework explicitly with region-level supervision.

\textit{\textbf{(e) From two-stage to one-stage.}} The aforementioned methods are proposed within a two-stage framework and inevitably encounter various limitations as mentioned in \cref{subsubsec:road_map}. Consequently, recent endeavors \cite{arbelle2021detector, jin2023refclip, chen2024querymatch} have aimed to transition from two-stage to one-stage inference. Specifically, RefCLIP (2023) \cite{jin2023refclip} leverages pre-trained detectors to extract anchor features and utilizes anchor-text matching for selecting target anchors for bounding box decoding. However, the anchor-based framework is impeded by the inability of fragment anchors to represent target information accurately. On the other hand, QueryMatch (2024) \cite{chen2024querymatch} treats WSVG as a query anchor-text matching problem and relies on query features extracted from Transformer-based detectors to represent objects. Subsequently, it employs bipartite matching, where query features can establish one-to-one associations with visual objects.

Although the aforementioned attempts have been made, WSVG still faces challenges in achieving accurate cross-modal grounding capabilities, mainly due to the following issues. \textbf{\textit{Firstly}}, these attempts often rely on a pre-computed set of candidate boxes that contain numerous distractors or background regions, which makes it challenging to identify the correct match. \textbf{\textit{Secondly}}, the candidate boxes typically remain fixed during the learning process due to constraints imposed by external detectors, leading to imprecise grounding. \textbf{\textit{Thirdly}}, these approaches usually represent noun phrases or visual target contexts implicitly by aggregating or encoding predicate triples using attention-based features. Such representations struggle to capture the rich semantics inherent in the relationship between image-sentence pairs, thereby impeding fine-grained cross-modal alignment and introducing ambiguity.


\subsubsection{VLP-based WSVG Transfer}

\textit{\textbf{(a) VLP-aided WSVG.}} Similar to the development of full supervision, researchers aim to enhance WSVG by leveraging the cross-modal alignment capability of the VLP models. To ensure a fair comparison, we separately evaluate these methods in \cref{tab:weakly_sup}. Specifically, on the one hand, methods such as CPL \cite{liu2023confidence}, g++ \cite{shaharabany2023similarity}, RefCLIP \cite{jin2023refclip}, QueryMatch \cite{chen2024querymatch}, UR \cite{zhang2024universal}, VPT-WSVG \cite{lin2024visual}, PPT \cite{zhao2024ppt}, \etc, utilize VLP model's (\eg, CLIP \cite{radford2021clip}, BLIP \cite{li2022blip}, \etc) cross-modal alignment capability to enhance confidence ranking when computing proposal-text similarity. 

\noindent \textit{\textbf{(b) VLP-based WSVG transfer.}} Besides, some VLP models (\eg, ALBEF \cite{li2021albef}, X-VLM \cite{zeng2022x-vlm}, \etc) endeavor to validate their fine-grained alignment capabilities through grounding tasks. However, as coarse-grained VLP lacks direct grounding capability, these methods typically perform cross-modality interaction between input images and text to generate a cross-modality attention map. Subsequently, by overlaying this attention map onto the original image, a cross-modal activation map (\eg, Grad-CAM \cite{Selvaraju2017Gradcam}) is created. Then, the additional detectors are employed to produce candidate boxes in a weakly supervised manner. Finally, the model calculates and ranks these candidate boxes based on the activation map to identify proposals with the highest scores. In these approaches, since VLP models have acquired comprehensive cross-modal representations from large-scale unlabeled data pairs during the pre-training phase, they only require minimal fine-tuning for grounding tasks and can achieve remarkable performance.

\subsection{Semi-supervised setting}
\label{subsec:semi-sup_setting}

As defined in \cref{sec:method_survey}, Semi-Supervised Visual Grounding (SSVG) aims to enhance the model's performance by leveraging limited labeled and unlabeled data. Compared to WSVG, semi-supervised approaches are relatively uncommon. Given the presence of unlabeled data, it is natural to consider employing pseudo-label generation methods for annotating the unlabeled samples (\eg, PQG-Distil \cite{jin2023pseudo}). Additionally, self-paced curriculum learning \cite{xiao2023clip, cascante2021curriculum} or a self-training framework can be utilized to acquire a more robust model from the labeled subset and subsequently refine and filter the unlabeled samples \cite{kang2024actress}. Alternatively, knowledge distillation can be employed to train a stronger teacher model using the labeled subset and then transfer its knowledge into the student model based on the unlabeled data (\eg, PQG-Distil \cite{jin2023pseudo}). Specifically, in \cite{rohrbach2016grounding}, the authors address the challenge of the limited availability of language annotations and bounding boxes by employing an attention mechanism to reconstruct a given phrase for grounding. In LSEP \cite{zhu2021utilizing}, the authors investigate scenarios where objects are without labeled queries and propose a location and subject embedding predictor to generate necessary language embeddings for annotating missing query targets in the training set. Additionally, SS-Ground \cite{chou2022semi} leverages an off-the-shelf pre-trained grounding model to generate pseudo-annotations for region-phrase alignment at multi-scales.

\subsection{Unsupervised Setting}
\label{subsec:unsup_setting}

To further reduce reliance on labor-intensive labeled data, the earlier Unsupervised Visual Grounding (USVG) methods \cite{yeh2018unsupervised, wang2019phrase, shi2023unpaired} have attempted to address this issue by utilizing unpaired image and query based on pre-trained detectors and an extra large-scale corpus. However, the approaches of both image-query and query-box double pairing present challenges. In contrast, Javed \etal \cite{javed2019learning} exploit the presence of semantic commonalities within a set of image-phrase pairs to generate supervisory signals. Pseudo-Q \cite{jiang2022pseudo} proposes template pseudo-label generation with object and attribute detectors, effectively eliminating errors caused by double pairing. Different from Pseudo-Q, CLIP-VG \cite{xiao2023clip} introduces three sources of pseudo-language labels and suggests self-paced curriculum adapting algorithms to strike a balance between reliability and diversity for the taxonomy-limited pseudo-labels in a self-training manner \cite{amini2024self-training}. Other methods, such as Omini-Q \cite{wang2024omni-q} and VG-annotator \cite{ye2024vg-ano}, follow a similar concept of generating pseudo-labels.

\subsection{Zero-shot Setting}
\label{subsec:zero-shot}

\begin{table}[t!]\footnotesize
\caption{A concise overview of methods for zero-shot settings.}
\vspace{-13pt}
\begin{center}
\resizebox{1.0\columnwidth}{!}{%
    \begin{tabular}{c|c|c|c|c}
    \toprule
    \multirow{2}[1]{*}{Methods}    &  \multirow{2}[1]{*}{Venue}  &  Pre-trained &  Two/one-  &  Fine-    \\
                                   &                             &  model &  stage  &  tuning    \\
        \midrule
    \multicolumn{5}{l}{\textit{\textbf{a. Grounding Novel Objects and Unseen Noun Phrases.}}}    \\
    ZSGNet \cite{sadhu2019zsgnet}   & ICCV'19     & None   &  Two-stage  &  Yes    \\
    MMKG   \cite{shi2022mmkg}       & AAAI'22     & None   &  Two-stage  &  Yes    \\
        \midrule
    \multicolumn{5}{l}{\textit{\textbf{b. Open Vocabulary Visual Grounding.}}}            \\
    CLIPREC \cite{ke2023cliprec}    & TMM'23      & CLIP   & One-stage  &  Yes    \\
    Wang \etal \cite{wang2024ovvg}  & Neurcom'23  & CLIP   & Two-stage  &  Yes    \\
    Mi \etal \cite{mi2024zero}      & Neurcom'24  & CLIP   & One-stage  &  Yes    \\
        \midrule
    \multicolumn{5}{l}{\textit{\textbf{c. Finetuning-free for Pre-trained Model with Detected Proposals.}}}       \\
    ReCLIP \cite{subramanian2022reclip} & ACL'22   & CLIP   & Two-stage  &  No     \\
    adapting-CLIP\cite{li2022adapting}  & Arxiv'22 & CLIP   & Two-stage  &  No     \\ 
    ChatRef \cite{sui2023chatref}       & ICLR'23  & GPT-4,GroundingDINO   & Two-stage  &  No     \\
    CPT \cite{yao2024cpt}            & AI Open'24  & VinVL\cite{zhang2021vinvl}  & Two-stage  &  No     \\
    VR-VLA \cite{han2024zero}           & CVPR'24  & CLIP   & Two-stage  &  No     \\
    GroundVLP \cite{shen2024groundvlp}  & AAAI'24  & CLIP,VinVL,ALBEF   & Two-stage  &  No        \\
    MCCE-REC \cite{qiu2024mcce}         & TCSVT'24 & CLIP,Vicuna\cite{chiang2023vicuna}  & Two-stage  &  No     \\
    CRG \cite{wan2024crg}               & ECCV'24  & LLaVA,GroundingDINO   & Two-stage  &  No     \\
    PSAIR \cite{pan2024psair}           & IJCNN'24 & CLIP,GroundingDINO    & Two-stage  &  No     \\
        \midrule
    \multicolumn{5}{l}{\textit{\textbf{d. Direct Grounding for Pre-trained Model without Fine-tuning and Proposals.}}}       \\
    GRILL \cite{liu2023grounding}         & Arxiv'23 &  Train from scratch    & One-stage  &  No     \\
    G-DINO\cite{liu2023grounding}  & ECCV'24  &  Swin,DINO,BERT  & One-stage  &  No     \\
    KOSMOS-2 \cite{peng2024kosmos-2}      & ICLR'24  &  KOSMOS-1\cite{huang2023kosmos-1},CLIP   & One-stage  &  No     \\
    \bottomrule
\end{tabular}%
}
\end{center}
\label{tab:zero-shot}
\vspace{-11pt}	
\end{table}

To further alleviate the data dependency and enhance the model's domain generalization ability beyond the limitations of the training or pre-training set, the zero-shot setting was proposed. As summarized in \cref{tab:zero-shot}, we roughly categorized zero-shot settings into four categories based on existing literature, \ie, grounding novel objects and unseen noun phrases, open vocabulary visual grounding, finetuning-free for pre-trained models with detected proposals, and direct grounding with a pre-trained model.

\subsubsection{Grounding Novel Objects and Unseen Noun Phrases}
\label{subsubsec:traditional_zsg}

Visual grounding differs from the detection task in that the grounding text is not a simple category word, but rather a free-form phrase or sentence.  Additionally, the query text is not limited to a fixed category (\eg, \textit{"the right one."} does not specify the class of the object). Therefore, it becomes challenging to strictly define a zero-shot setting for the grounding task. In 2019, asadhu \etal \cite{sadhu2019zsgnet} first introduced an acceptable zero-shot grounding setting. As shown in \cref{fig:zsg_def}-(b), assuming that the referred subject of the query during training is the base class and the referred subject of the query during testing is a novel class, ZSGNet \cite{sadhu2019zsgnet} divides this novel class into four cases, \ie, (1) \textit{Case 0}: the referred noun in its test set is not included in its training set; (2) \textit{Case 1}: the categories of the referred objects in its test set are not covered by its training set; (3) \textit{Case 2}: the objects semantically close to the referred objects in the test set only appear in the training set; (4) \textit{Case 3}: the objects semantically close to the referred objects in the test set appear not only in the training set but also in the test set. Simultaneously, to facilitate the zero-shot setting, Flick30k Entities and Visual Genome datasets were partitioned into \textit{Flickr-split-0}, \textit{Flickr-split-1}, \textit{VG-split-2}, and \textit{VG-split-3} respectively based on the four cases \cite{sadhu2019zsgnet}. Similarly, in 2023, CLIPREC \cite{ke2023cliprec} continued to reorganize the RefCOCO/+ dataset and built the RefCOCOZ/+ dataset by following the rules of case 0 and case 1 in ZSGNet. Wang \etal \cite{wang2024ovvg} integrated the existing COCO and LVIS datasets by taking 80 COCO classes as base classes while constructing OV-VG and OV-PL datasets with 100 novel classes. In terms of methods, ZSGNet \cite{sadhu2019zsgnet} utilizes a traditional BiLSTM structure along with a two-stage detector. MMKG \cite{shi2022mmkg} constructed a multimodal knowledge graph incorporating external linguistic knowledge and then performed graph reasoning and spatial relationship analysis for grounding the noun phrases. TransCP \cite{tang2023transcp} introduces a context disentangling and prototype inheriting strategy to perceive novel objects.


\subsubsection{Open Vocabulary Visual Grounding}
\label{subsubsec:ovvg}

\begin{figure}[t!]
	\centering
	\includegraphics[width=0.99\linewidth]{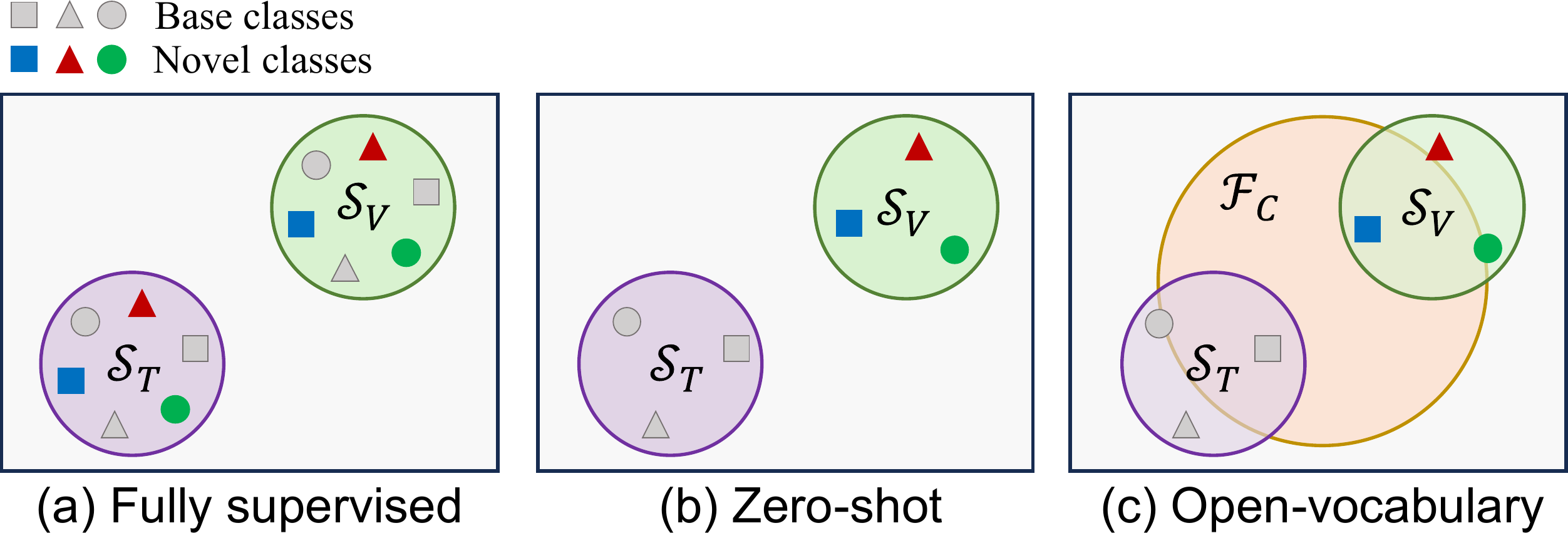}
        \vspace{-6pt}
        \caption{\small Concepts comparison of the (b) Zero-shot Grounding (ZSG)  (\cref{subsubsec:traditional_zsg}) and (c) Open Vocabulary Grounding (OVG) (\cref{subsubsec:ovvg}) with the (a) fully supervised (\cref{subsec:fully_sup}) setting in visual grounding task. The fully supervised setting does not distinguish between basic and novel classes in the training $\mathcal{S}_{T}$ and test $\mathcal{S}_{V}$ sets, while ZSG does. In OVG, the model can ground novel class objects with the help of a large vocabulary feature space $\mathcal{F}_{C}$.}
        \vspace{-14pt}
	\label{fig:zsg_def}
\end{figure}

Open Vocabulary (OV) \cite{wu2024towards} is a special setting in zero-shot learning. Its concept was first proposed in OVR-CNN \cite{zareian2021ovrcnn} and has become a popular setting in the field of detection with the introduction of CLIP. As depicted in \cref{fig:zsg_def}, unlike the traditional definition of zero-shot learning, Open Vocabulary Grounding (OVG or OVVG) refers to the fact that during the pre-training phase, the model may be exposed to a wider range of vocabularies, which may or may not include base classes and novel classes. \textit{However, as described in the previous section, visual grounding itself represents a natural open vocabulary setting because it is trained with long and open-set query text.} Nevertheless, recent work using large-scale pre-trained models (\eg, CLIP) go beyond traditional zero-shot settings' limitations and should be distinguished as OVG for a fair comparison. Specifically, Wang \etal \cite{wang2024ovvg} suggests employing CLIP as the text encoder within the zero-shot framework while leveraging additional training data to enhance generalization on novel classes. CLIPREC \cite{ke2023cliprec} integrates existing detectors for proposal extraction and incorporates CLIP into a graph-based adaptive network to improve perception capabilities toward novel class identification. Moreover, the GMLLMs (\eg, Ferret \cite{you2023ferret}, KOSMOS-2 \cite{peng2024kosmos-2}, GEM \cite{bousselham2024gem} \etc) should be regarded as a natural approach to OVG.

\subsubsection{Finetuning-free for Pre-trained model with Proposals}

This setting bears some resemblance to WSVG, with the key distinction being that weakly supervised methods necessitate grounding training data while the former does not. In these approaches, pre-trained models typically exhibit strong generalization and cross-modal alignment capabilities; however, they lack grounding ability and rely on off-the-shelf proposals provided by existing detectors. Consequently, these zero-shot methods predominantly adopt a two-stage detection-then-matching approach for grounding referred objects. Due to their inability to distinguish between the base class and novel class, these methods are commonly evaluated using the traditional RefCOCO/+/g dataset.

This setting originates from ReCLIP \cite{subramanian2022reclip} in 2021, which employs an NLP language parser and a spatial relation parser to parse sentences. It then utilizes CLIP's cross-modal alignment capabilities to assess the similarity between extracted nouns and proposals. Consequently, the optimal proposal can be obtained without requiring training for grounding purposes. Subsequently, colorful prompt tuning (CPT) \cite{yao2024cpt} proposed reformulating visual grounding as a fill-in-the-blank problem using color-based co-referential markers in both image and text contexts. GroundVLP \cite{shen2024groundvlp} adopts an open vocabulary object detector for detecting proposal boxes and employs GradCAM \cite{Selvaraju2017Gradcam} to capture expressive-related regions within images. These detected proposals are then combined with the region features to ground referred objects effectively. The MCCE-REC \cite{qiu2024mcce} introduces multi-cues cross-modal interaction and contrastive similarity entropy based on richer cues generated by MLLM (\eg, LLaVA \cite{liu2024llava}), ultimately achieving accurate predictions. Similarly, CRG \cite{wan2024crg} leverages classifier-free guidance to assist open-source MLLMs in focusing on specific regions and comprehending visual markers without additional training requirements. VR-VLA \cite{han2024zero} propagates similarity scores from grounded triplets to instances based on ReCLIP; however, this method is fine-tuned using additional datasets and does not strictly adhere to finetuning-free zero-shot principles.

\subsubsection{Direct Grounding without Fine-tuning and Proposals}

These approaches primarily rely on the utilization of fine-grained detection or grounding data during the pre-training stage, enabling the model to possess a certain level of detection or grounding capability even without fine-tuning. Consequently, in zero-shot scenarios, additional detected proposals are no longer necessary for grounding purposes. Such approaches are predominantly observed in large-scale detection models (\eg, GLIP \cite{li2022glip}, Grounding DINO \cite{liu2023grounding}) and MLLMs (\eg, KOSMOS-2 \cite{peng2024kosmos-2}, GRILL \cite{jin2023grill}), which often employ zero-shot grounding as an indicator of the pre-trained model's generalization ability. Therefore, these approaches are typically evaluated directly on the RefCOCO/+/g datasets.

\subsection{Multi-task Setting}
\label{subsec:multi-task_setting}

\subsubsection{REC and REG Multi-task Setting}
\label{subsec:rec_and_reg}

REG \cite{krahmer2012computational} is a classical Natural Language Processing (NLP) problem \cite{winograd1972understanding}. As mentioned in \cref{sec:introduction} and \cref{subsec:realted_domain}, REC originated from the REG task \cite{kazemzadeh2014referitgame}. REC and REG naturally exhibit cycle-consistency constraints. Prior to 2014, research primarily focused on small-scale computer-generated datasets that were not connected to real-world vision systems \cite{mitchell2010natural, van2006building}. In 2016, Mao \etal \cite{refcocog-google}  proposed max-margin Maximum Mutual Information (MMI) joint training, which enables direct generation of expressive text without relying on object or attribute vocabulary. In 2017, Yu \etal \cite{yu2017joint} introduced a joint speaker-listener-reinforcer (SLR) model based on CNN/LSTM for referring expression tasks. They incorporated a reinforcement mechanism and utilized a reward loss to sample more discriminative expression text. Luo \etal \cite{Luo_Shakhnarovich_2017} presented a stochastic mixed incremental cross-entropy comprehension (SMIXEC) approach to leverage learned comprehension models for guiding the generation of improved referring expressions. Liu \etal \cite{liu2017attribute} explores an attribute learning model from visual objects and their paired descriptions, and then incorporates them into both REG and REC branches. Recently, Wang \etal \cite{wang2024cycle} and VLM-VG \cite{wang2025learning} leverage a cycle-consistency learning framework to connect a simple regional image captioner and a Transformer-based grounding model with weight-sharing architecture for large-scale dataset generation and end-to-end pre-training.

\subsubsection{REC and RES Multi-task Setting}
\label{subsec:rec_and_res}

As stated in \cref{subsec:realted_domain}, the REC and RES tasks can be likened to a pair of fraternal twins. In previous studies (\eg, MAttNet (2018) \cite{yu2018mattnet}, MCN (2020) \cite{luo2020mcn}), REC and RES were often discussed concurrently within a multi-task collaborative network by employing two distinct prediction heads based on a shared model. However, with the advancement of pre-training techniques, subsequent research has addressed these tasks separately due to the relatively easier acquisition of a bounding box for grounding compared to segmentation masks in both datasets and result regression \cite{deng2021transvg}. Recent studies (\eg, RefTR (2021) \cite{li2021referring}, SeqTR (2022) \cite{zhu2022seqtr}, PolyFormer (2023) \cite{liu2023polyformer}, VG-LAW (2023) \cite{vg-law}, UniQRnet (2024) \cite{ye24uniqrnet}, EEVG (2024) \cite{chen2024eevg}, M2IF (2024) \cite{qin2024improving}) have followed a similar collaborative and consistency framework and demonstrated that multi-task training involving REC and RES enhances the generalization capabilities of the underlying backbone model and yields improved performance compared to single-task training.

\subsubsection{Grounding with Other Tasks}
\label{subsec:rec_and_other_task}

Numerous studies \cite{yang2022unitab} have demonstrated that the generalization capability of a model can be significantly enhanced through the integration of multiple tasks. In addition to REG and RES, visual grounding can also facilitate numerous other tasks. For instance, grounding plays a supportive role in Grounded VQA \cite{zhu2016visual7w} by enabling object identification before answering questions (\eg, ``\textit{what color is the person on the left wearing?}") \cite{chen2022grounding, khan2022weakly, shrestha2020negative, reich2024uncovering, riquelme2020explaining}. Fukui \etal \cite{fukui2016mcb} propose a multimodal compact bilinear pooling approach for combining multimodal features expressively in both grounding and VQA tasks. Nguyen \etal \cite{nguyen2019multi} introduce a hierarchical multitask learning method tailored to different datasets encompassing multiple tasks (including grounding, VQA, and retrieval). It learns shared visual language representations hierarchically with predictions made at corresponding levels. Furthermore, as described in \cref{subsubsec:road_map}, grounding is often pre-trained alongside other tasks to acquire more generalized representations.

\vspace{-2pt}
\subsection{Generalized Visual Grounding}
\label{subsec:grec}

The definition and evaluation metric of the GVG (or GREC) task are introduced in \cref{subsec:definition} and \cref{subsec:evaluation_metric}, respectively. The counterpart of GREC \cite{he2023grec, xie2024described} is GRES (Generalized RES) \cite{liu2023gres, hu2023beyond, pramanick2024vistallm}, and these two tasks are often studied in conjunction. Compared with traditional grounding methods, GREC demonstrates greater practical potential. However, several challenges have hindered its development, particularly in task formulation, evaluation criterion definition, dataset construction, and output modeling. As a result, no studies had been conducted on this setting prior to 2023. He \etal's analysis \cite{he2023grec} focuses on dataset construction and evaluation criteria for the GREC tasks. Under the defined scope of the GREC, traditional approaches such as single region token regression (\eg, TransVG \cite{deng2021transvg}) or top-1 bounding box-based methods (\eg, MDETR \cite{kamath2021mdetr}) are no longer applicable due to the requirement of returning an uncertain number of multiple grounding boxes. Instead, the number of grounding targets can be indirectly constrained by considering the confidence associated with each box prediction, thus requiring models to utilize an anchor query-based decoder approach. After He \etal's adaptation, customized MCN \cite{luo2020mcn}, VLT \cite{ding2021vlt}, MDETR \cite{kamath2021mdetr}, UNINEXT \cite{yan2023uninext}, RECANTFormer \cite{hemanthage2024recantformer} and SimVG \cite{dai2024simvg} have become capable of handling GREC. To more effectively tackle the varying number of target objects, HieA2G \cite{wang2025hiea2g} introduces an adaptive grounding counter that dynamically determines the number to help select the outputs.

\section{Challenges and Outlook}
\label{sec:future_direction}

\subsection{Challenges}
\label{subsec:challenge}

The present studies are subject to several limitations.

\noindent $\bullet$ \textit{\textbf{Dataset Limitations.}} As previously discussed in this survey, the current grounding datasets face several limitations and challenges. \textit{(a) Firstly}, as shown in \cref{tab:full_sota_main} and \cref{fig:perf_trend}-(b), widely-used datasets such as RefCOCO/+/g have become saturated and are approaching their performance limits. These datasets are derived from the MSCOCO dataset and suffer from limited diversity in object categories, relatively simple textual expressions, and relatively small overall sizes. Similar issues also exist in other datasets, such as Flickr30k, ReferIt, \etc. These limitations hinder the evaluation of models with stronger reasoning capabilities and better generalization performance. \textit{(b) Secondly}, several recently proposed datasets rely on pseudo-labels generated by pre-trained models (\eg, GPT-4 models), which often contain significant noise and often result in suboptimal quality. Research has demonstrated that training models on model-generated pseudo-labels may lead to model poisoning \cite{shumailov2024ai}. \textit{(c) Moreover}, existing datasets predominantly support only single-round reasoning and lack the capacity to represent complex logical expressions (\eg, \textit{grounding ``the fruit richest in vitamin C" in an image that contains multiple types of fruits}). In the era of MLLM-based general-purpose AI, existing datasets are insufficient for supporting multi-round referring dialogues or evaluating GMLLMs. \textit{(d) Fourthly}, for emerging scenarios, such as GREC and multi-image single/multi-object grounding, \etc, current datasets remain limited in both scale and complexity.

\noindent $\bullet$ \textit{\textbf{Task Definition Limitations.}} As discussed in \cref{subsec:definition}, a strong assumption in current grounding research posits that there is and must be only one referring object within an image. This assumption, however, conflicts with real-world scenarios. There remains a need for more inclusive, realistic settings and corresponding evaluation metrics.

\noindent $\bullet$ \textit{\textbf{Video Scenarios.}} Current research on grounding mainly focuses on static images; however, video streams are more practical for applications in surveillance, robotics, and embodied intelligence. While current video grounding research  \cite{wu2025survey} primarily addresses coarse-grained temporal segments, the study of grounding referring objects in video streams is still in its initial stages, particularly concerning datasets, evaluation criteria, and methodologies.

\noindent $\bullet$ \textit{\textbf{Grounding Scaling.}} The current large-scale grounding pre-training faces limitations in two key aspects. \textit{(a) Firstly}, the availability of suitable datasets remains limited. Due to the scarcity of fine-grained box annotation data, the scale of existing grounding training datasets is still relatively small. Although DINO-X \cite{ren2024dino-x} introduced a proprietary Grounding-100M dataset, its size is still significantly smaller compared to widely adopted open-source datasets such as LAION-400M \cite{schuhmann2021laion} and LAION-5B \cite{schuhmann2022laion-5b}, which exceed 400 million samples. The research community currently lacks open-source, high-quality, ultra-large-scale, and fine-grained datasets. \textit{(b) Secondly}, the pre-training paradigm presents challenges. As discussed in \cref{subsec:fully_sup}, potential approaches for achieving large-scale grounding pre-training include phrase-region correspondence-based grounded pre-training \cite{li2022glip} or multi-task pre-training. However, these methods heavily depend on manually annotated fine-grained data, which inherently limits the scalability of grounding pre-training.

\noindent $\bullet$ \textit{\textbf{Application Limitations.}} Current AI is still far from achieving general AI. A notable characteristic is that the current grounding research remains in its nascent stages. Beyond the RefCOCO/+/g dataset, numerous potential applications of grounding have yet to be fully explored.

\subsection{Future Directions}
\label{subsec:future_direction}

In response to the present challenges, some future research directions can be inferred.

\noindent $\bullet$ \textit{\textbf{New Evaluation Benchmarks.}} By analyzing the limitations of existing datasets, we can identify several key features for future datasets. \textit{(a) Large diversity in object categories.} The new dataset must meet the increasing demand for diverse object categories in open-world scenarios. \textit{(b) Large-scale in dataset size.} Compared with the datasets used by VLP models, current grounding datasets remain relatively small in scale, which constrains the upper performance limit of grounding models. \textit{(c) Large diversity in instance scales.} The object instances in existing datasets exhibit relatively uniform scales, which leads to models with limited sensitivity to variations in object size. \textit{(d) Enhanced textual semantic reasoning.} Current datasets rely on relatively simple textual expressions. Future datasets should include samples that require strong logical reasoning for grounding, thereby providing a more comprehensive evaluation of model capabilities. However, the creation of such datasets entails significantly higher annotation costs. \textit{(e) More aligned with the concept of GREC.} As generalized grounding becomes increasingly prevalent, future datasets must be designed to accommodate these scenarios effectively. \textit{(f) Meet more universal grounding scenarios.} As a fine-grained cross-modal task, grounding encompasses a broader range of universal application scenarios compared to classic or generalized VG. These include multi-image single/multi-object grounding, multi-round referring dialogues, and multi-expression simultaneous grounding, \etc.


\noindent $\bullet$ \textit{\textbf{Universal Multi-modal Grounding.}} Universal multi-Modal grounding envisions a unified framework that grounds arbitrary expressions across diverse modalities, environments, and interaction patterns. Future systems may support cross-device grounding (\eg, between mobile phones, AR glasses, and drones), multi-user collaborative grounding, and context-adaptive grounding that incorporates spatial, temporal, auditory, or even tactile signals. Grounding models will need to interpret ambiguous, personalized, or evolving queries in real time, possibly combining vision, speech, gesture, and text simultaneously. Scenarios such as grounding in egocentric video streams, audio-described spatial references, or interactive multi-turn grounding dialogs push the boundaries of current systems. Achieving this vision requires advances in continual learning, knowledge grounding, and reasoning under uncertainty for truly general-purpose multimodal agents.


\noindent $\bullet$ \textit{\textbf{Generalized Video Object Grounding.}} The grounding of natural numbers (not just one) of objects in each frame of a video stream holds extensive application prospects \cite{ding2023mevis}, particularly in intelligent transportation, engineering safety, and other domains. Future research in this field could focus on defining tasks, developing datasets for various industry scenarios, establishing evaluation criteria, and constructing methodologies, all of which present significant research potential.

\noindent $\bullet$ \textit{\textbf{Self-supervised Grounding Pre-training.}} As discussed in \cref{subsec:challenge}, advancing large-scale grounding pretraining in the future will require effort either in the dataset construction or in the pretraining paradigm. \textit{On one hand}, this can be achieved by building a large-scale region-level dataset, followed by implementing grounded pretraining or multi-task pretraining strategies. Alternatively, self-training with pseudo-grounding using CLIP and SAM also presents a feasible approach. \textit{On the other hand}, future research should attempt to explore self-supervised grounding pretraining methods that do not explicitly depend on region-level supervision, enabling the underlying models to achieve precise cross-modal grounding and understanding capabilities. OneRef \cite{xiao2024oneref}  introduced the Masked Referring Modeling (MRefM) self-supervised pretraining paradigm, which eliminates the need for explicit fine-grained supervision by utilizing unsupervised bounding box generation algorithms. Future studies may further investigate region-level self-supervised learning approaches based on pretraining methods such as MAE and contrastive learning \etc.

\noindent $\bullet$ \textit{\textbf{Empower General AI applications with Grounding.}} In addition to the novel grounding applications discussed in Sec. \textcolor{blue}{3} of Appendix (such as high-resolution grounding \cite{ma2024visual}, multi-image visual grounding \cite{xu2024mc-bench}, and referring counting \cite{dai2024ref-count}), grounding technology can facilitate a wide range of tasks and scenarios. For instance, when deployed in security robots and embodied intelligence systems, the challenges of interactive grounding and continuous grounding between robots and humans must be addressed. The technical includes real-time video stream grounding, the modeling of grounding data flows, and the integration of human feedback.

\section{Conclusion}
\label{sec:conclusion}

In this survey, we systematically track and summarize the advancements in visual grounding over the past decade. To the best of our knowledge, this review represents the most comprehensive overview currently available in the field of visual grounding. Specifically, we initially examine the developmental history of visual grounding and provide an overview of essential background knowledge, including fundamental concepts and evaluation metrics. Subsequently, we meticulously organize the various settings in visual grounding and establish precise definitions of these settings to standardize future research. In the dataset section, we compile a comprehensive list of current relevant datasets and conduct a fair comparative analysis. Additionally, we delve into numerous applications and highlight several advanced topics of visual grounding. Finally, we outline the challenges confronting visual grounding and propose valuable directions for future research, which may serve as inspiration for subsequent researchers. This paper is designed to be suitable for both beginner and experienced researchers in visual grounding, serving as an invaluable resource for tracking the latest research developments.




\section*{\Large{Appendices}}
\vspace{4pt}

\setcounter{equation}{0}
\setcounter{table}{0}
\setcounter{figure}{0}
\setcounter{section}{0}
\setcounter{subsection}{0}
\renewcommand{\thetable}{A.\arabic{table}}
\renewcommand{\theequation}{A.\arabic{equation}}
\renewcommand{\thesubsection}{A.\arabic{subsection}}

\renewcommand{\thesection}{A\arabic{section}}
\section{Methods: Supplementary Material}
\renewcommand{\thesection}{A\arabic{section}}
\label{sec:supp_method}

Due to space limitations in the main text, we present the full table of the representative fully supervised methods during the surge stage in \cref{tab:rec_sota}. This table corresponds to Tab. \textcolor{blue}{2} in the main text.


\begin{table*}[t!]
  \setlength\tabcolsep{5pt}
    \footnotesize
    \renewcommand{\thetable}{A\arabic{table}}
    \caption{A performance comparison of representative methods from the new era on RefCOCO/+/g datasets under the fully supervised setting.}
    \vspace{-6pt}
    \centering
    \resizebox{2.04\columnwidth}{!}{%
    \begin{tabular}{c|c|c|c|c|ccc|ccc|cc}
    \toprule
    \multirow{2}[1]{*}{Methods} & \multirow{2}[1]{*}{Venue} & Visual / Language & Intermediate  & Data pair & \multicolumn{3}{c|}{RefCOCO \cite{yu2016modeling}} & \multicolumn{3}{c|}{RefCOCO+ \cite{yu2016modeling}} & \multicolumn{2}{c}{RefCOCOg \cite{refcocog-umd}}  \\
                                &                           & Backbone        & pretrain data & size &   val  & testA & testB       &   val  & testA & testB        &   val  & test               \\
    \midrule    
    \multicolumn{11}{l}{\textit{\textbf{a. Single-dataset fine-tuning setting \textit{w.} unimodal pre-trained close-set detector and language model: \textit{(traditional setting)}}}}        \\
    TransVG \cite{deng2021transvg}     &   ICCV'21     & RN101+DETR / BERT-B    & -- & -- &  81.02  & 82.72  & 78.35  & 64.82  & 70.70  & 56.94  & 68.67  & 67.73    \\
    SeqTR \cite{zhu2022seqtr}          &   ECCV'22     & DN53  / BiGRU          & -- & -- & 81.23  & 85.00  & 76.08  & 68.82  & 75.37  & 58.78  & 71.35  & 71.58    \\ 
    RefTR \cite{li2021referring}      &   NeurIPS'21  & RN101+DETR / BERT-B    & -- & -- & 82.23  & 85.59  & 76.57  & 71.58  & 75.96  & 62.16  & 69.41  & 69.40  \\
    Word2Pix \cite{zhao2022word2pix}   &   TNNLS'22    & RN101+DETR / BERT-B    & -- & -- & 81.20  & 84.39  & 78.12  & 69.74  & 76.11  & 61.24  & 70.81  & 71.34       \\
    QRNet \cite{qrnet}                 &   CVPR'22     & Swin-S \cite{liu2021swin} / BERT-B  & -- & -- & 84.01  & 85.85  & 82.34  & 72.94  & 76.17  & 63.81  & 71.89  & 73.03   \\   
    LADS \cite{su2023lads}             &   AAAI'23     & RN50+DETR / BERT-B    & -- & -- & 82.85  & 86.67  & 78.57  & 71.16  & 77.64  & 59.82  & 71.56  & 71.66 \\
    VG-LAW \cite{vg-law}               &   CVPR'23     & ViT-Det \cite{vit-det} / BERT-B    & -- & -- & 86.06  & 88.56  & 82.87  & 75.74  & 80.32  & 66.69  & 75.31  & 75.95 \\
    TransVG++\cite{transvg++}          &   TPAMI'23    & ViT-Det \cite{vit-det} / BERT-B   & -- & -- & 86.28  & 88.37  & 80.97  & 75.39  & 80.45  & 66.28  & 76.18  & 76.30  \\ 
        \midrule  
    \multicolumn{11}{l}{\textit{\textbf{b. Single-dataset fine-tuning setting \textit{w.} self-supervised vision-language pre-trained model: }}}       \\ 
    CLIP-VG \cite{xiao2023clip} &   TMM'23      & CLIP-B / CLIP-B  & -- & -- & 84.29  & 87.76  & 78.43  & 69.55  & 77.33  & 57.62  & 73.18  & 72.54   \\  
    JMRI \cite{zhu2023jmri} &   TIM'23      & CLIP-B / CLIP-B   & -- & -- & 82.97 & 87.30  & 74.62  & 71.17  & 79.82  & 57.01  & 71.96  & 72.04   \\  
    D-MDETR \cite{shi2023dynamic} & TPAMI'23    & CLIP-B / CLIP-B  & -- & -- & 85.97  & 88.82  & 80.12  & 74.83  & 81.70  & 63.44  & 74.14  & 74.49   \\
    HiVG-B \cite{xiao2024hivg}  &   ACMMM'24      & CLIP-B / CLIP-B     & -- & -- &   87.32    &   89.86    &   83.27    &   78.06    &   83.81    &   68.11    &   78.29    &   78.79    \\
    HiVG-L \cite{xiao2024hivg}  &   ACMMM'24     & CLIP-L / CLIP-L     & -- & -- &  88.14  & 91.09  & 83.71  & 80.10  &  86.77  & 70.53  & 80.78  & 80.25   \\
    OneRef-B  \cite{xiao2024oneref}    &   NeurIPS'24     & BEiT3-B / BEiT3-B    & -- & -- &   88.75    &   90.95    &   85.34    &   80.43    &   86.46    &   74.26    &   83.68    &   83.52    \\  
    OneRef-L  \cite{xiao2024oneref}    &   NeurIPS'24     & BEiT3-L / BEiT3-L    & -- & -- &   92.87    &   94.01    &   90.19  &     87.98  &     91.57  &     83.73  &    88.11  &  89.29   \\
    \midrule       
    \multicolumn{13}{l}{\textit{\textbf{c. Dataset-mixed intermediate pre-training setting:}}}       \\ 
    MDETR$^\dagger$ \cite{kamath2021mdetr}     & ICCV'21   & RN101/RoBERT-B        & GoldG,RefC          & 6.5M  & 86.75  & 89.58  & 81.41  & 79.52  & 84.09  & 70.62  & 81.64  & 80.89  \\  
    YORO$^\dagger$ \cite{ho2023yoro}            & ECCV'22 & ViLT \cite{kim2021vilt} / BERT-B  & GoldG,RefC          & 6.5M  & 82.90  & 85.60  & 77.40  & 73.50  & 78.60  & 64.90  & 73.40  & 74.30  \\
    DQ-DETR$^\dagger$ \cite{liu2023dqdetr}     & AAAI'23   & RN101 / BERT-B        & GoldG,RefC          & 6.5M  & 88.63  & 91.04  & 83.51  & 81.66  & 86.15  & 73.21  & 82.76  & 83.44  \\
    Grounding-DINO-B$^\dagger$                  & ECCV'24  & Swin-T     / BERT-B   & O365,GoldG,RefC     & 7.2M  & 89.19  & 91.86  & 85.99  & 81.09  & 87.40  & 74.71  & 84.15  & 84.94  \\
    Grounding-DINO-L$^\dagger$                  & ECCV'24  & Swin-L     / BERT-B   & G-DINO-L$^*$        & 21.4M & 90.56  & 93.19  & 88.24  & 82.75  & 88.95  & 75.92  & 86.13  & 87.02  \\
    HiVG-B* \cite{xiao2024hivg}        & ACMMM'24  & CLIP-B / CLIP-B       & RefC,ReferIt,Flickr & 0.8M  & 90.56  & 92.55  & 87.23  & 83.08  & 87.83  & 76.68  & 84.71  & 84.69  \\
    HiVG-L* \cite{xiao2024hivg}        & ACMMM'24  & CLIP-L / CLIP-L       & RefC,ReferIt,Flickr & 0.8M  & 91.37  & 93.64  & 88.03  & 83.63  & 88.16  & 77.37  & 86.73  & 86.86  \\
    OneRef-B* \cite{xiao2024oneref}   & NeurIPS'24 & BEiT3-B / BEiT3-B & RefC,ReferIt  & 0.5M   &   91.89 &  94.31 &  88.58  &  86.38  &  90.38  &  79.47  &  86.82  &  87.32  \\
    OneRef-L* \cite{xiao2024oneref}   & NeurIPS'24 & BEiT3-L / BEiT3-L & RefC,ReferIt  & 0.5M   &   93.21 &  95.43 &  90.11  &  88.35  &  92.11  &  82.70  &  87.81  &  88.83  \\
    \midrule       
    UNITER-B$^\ddagger$\cite{chen2020uniter}     & ECCV'20   & UNITER-B / UNITER-B & ALBEF$^*$\cite{li2021albef}  & $\sim$17M & 81.24  & 86.48  & 73.94  & 75.31  & 81.30  & 65.58  & 74.31  & 74.51  \\  
    VILLA$^\ddagger$ \cite{gan2020villa}  &   NeurIPS'20     & VILLA-B / VILLA-B & ALBEF$^*$\cite{li2021albef}  & $\sim$17M    & 81.65 & 87.40 & 74.48 & 76.05 & 81.65 & 65.70 & 75.90 & 75.93  \\
    UniTAB$^\ddagger$ \cite{yang2022unitab}     & ECCV'22   & RN101/RoBERT-B & VG,COCO,\etc & $>$20M  & 88.59  & 91.06 & 83.75  & 80.97  & 85.36  & 71.55  & 84.58  & 84.70   \\
    FIBER$^\ddagger$ \cite{dou2022fiber}              & NeurIPS'22   &  Swin-B / RoBERT-B   & CC,SBU,VG,GoldG,\etc & $\sim$5M & 90.68   &  92.59   & 87.26  & 85.74  & 90.13  & 79.38  & 87.11  & 87.32    \\
    OFA-B$^\ddagger$ \cite{wang2022ofa}         & ICML'22   & OFA-B      / OFA-B    &  unavailable & --  & 88.48  & 90.67  & 83.30  & 81.39  & 87.15  & 74.29  & 82.29  & 82.31   \\
    OFA-L$^\ddagger$ \cite{wang2022ofa}         & ICML'22   & OFA-L      / OFA-L    &  unavailable & --  & 90.05  & 92.93  & 85.26  & 85.80  & 89.87  & 79.22  & 85.89  & 86.55   \\
    mPlug$^\ddagger$ \cite{li2022mplug}       & EMNLP'22    & CLIP-L / BERT-B   & ALBEF$^*$\cite{li2021albef}  & $\sim$17M  & 92.40  & 94.51  & 88.42  & 86.02  & 90.17  & 78.17  & 85.88  & 86.42   \\
    mPlug-2$^\ddagger$ \cite{li2022mplug}       & ICML'23     & Swin-T / BERT-B   & ALBEF$^*$\cite{li2021albef}  & $\sim$17M  & 90.33  & 92.80  & 86.05  &  -- & --  & --  & 84.70  & 85.14   \\
    CyCo$^\ddagger$ \cite{wang2024cycle}                   & AAAI'24   & ViT\cite{dosovitskiy2020image}/ BERT-B & VG,SBU,CC3M,\etc    & $>$120M & 89.47  & 91.87  & 85.33  & 80.40  & 87.07  & 69.87  & 81.31  & 81.04  \\
    \midrule
    \rowcolor{cyan!07}
    \multicolumn{3}{l|}{~~~\textit{\textbf{Predicted ultimate performance (based on OneRef-B model): }}} & RefC,ReferIt  & 0.5M  & 98.69  &  99.08 & 98.57   & 97.93   & 98.50   & 97.34   & 98.14   & 98.48  \\
    \rowcolor{cyan!07}
    \multicolumn{3}{l|}{~~~\textit{\textbf{Predicted ultimate performance (based on OneRef-L model): }}} & RefC,ReferIt  & 0.5M  & 99.01  & 99.10  & 98.95   & 98.51   & 98.92   & 98.76   & 98.94   & 98.98   \\
    \midrule  
    \multicolumn{13}{l}{\textit{\textbf{d. Fine-tuning setting \textit{w.} grounding multimodal large language model (GMLLM): }}}       \\ 
    Shikra-7B \cite{chen2023shikra}   &   arXiv'23     & CLIP-L / Vicuna-7B\cite{chiang2023vicuna}  & L-Inst,RefC,VG,RD,\etc &  $\sim$4M   & 87.01 & 90.61 & 80.24 & 81.60 & 87.36 & 72.12 & 82.27 & 82.19   \\
    Shikra-13B \cite{chen2023shikra}  &   arXiv'23     & CLIP-L / Vicuna-13B\cite{chiang2023vicuna} & L-Inst,RefC,VG,RD,\etc &  $\sim$4M   & 87.83 & 91.11 & 81.81 & 82.89 & 87.79 & 74.41 & 82.64 & 83.16  \\
    Ferret-7B \cite{you2023ferret}  &   ICLR'24     & CLIP-L / Vicuna-7B\cite{chiang2023vicuna} & GRIT \cite{you2023ferret} & $>$8M    & 87.49 & 91.35 & 82.45 & 80.78 & 87.38 & 73.14 & 83.93 & 84.76   \\
    Ferret-13B \cite{you2023ferret}  &   ICLR'24     & CLIP-L / LLaVA-13B & GRIT \cite{you2023ferret} & $>$8M   & 89.48 & 92.41 & 84.36 & 82.81 & 88.14 & 75.17 & 85.83 & 86.34  \\
    Next-Chat \cite{zhang2024next-chat}  &   ICML'24   & CLIP-L / Vicuna-7B\cite{chiang2023vicuna} & L-Inst,RefC,GRIT,...  & $>>$20M  & 88.69  & 91.65 & 85.33 & 79.97 & 85.12 & 74.45 & 84.44 & 84.66  \\
    MiniGPT-v2 \cite{chen2023minigpt-v2}  &   arXiv'23   & CLIP-L / Vicuna-7B\cite{chiang2023vicuna} & L-Inst,RefC,GRIT,...  & $>>$20M  & 88.69  & 91.65 & 85.33 & 79.97 & 85.12 & 74.45 & 84.44 & 84.66  \\
    LLaVA-G \cite{zhang2024llava-grounding}  &   ECCV'24   & CLIP-L,Swin-T / Vicuna-7B & L-Inst,RefC,VG,Flickr,...  & unknown  & 89.16  & --  & --  & 86.18 & -- & -- & 84.82 & --  \\
    G-GPT \cite{li2024groundinggpt}  &   ACL'24     & CLIP-L / Vicuna-7B\cite{chiang2023vicuna} & L-Inst,RefC,VG,Flickr,...  & unknown  & 88.02   & 91.55 & 82.47 & 81.61 & 87.18 & 73.18 & 81.67 & 81.99  \\
    Groma \cite{ma2024groma}  &   ECCV'24     & DINOv2-L / Vicuna-7B\cite{chiang2023vicuna} & L-Inst,RefC,VG,Flickr,...  & unknown  & 89.53  & 92.09 & 86.26 & 83.90 & 88.91 & 78.05 & 86.37 & 86.52  \\
    QWen-VL \cite{bai2023qwen-vl}  &   arXiv'23     & EVA-G / QWen\cite{bai2023qwen} & LAION,GRIT,RefC,...  & $>$1.5B  & 89.36  & 92.26 & 85.34 & 83.12 & 88.25 & 77.21 & 85.58 & 85.48  \\
    VisCoT \cite{shao2024viscot}  &   CoRR'24      & CLIP-L / Vicuna-7B\cite{chiang2023vicuna} & L-Inst,RD,VisCoT,\etc  & $\sim$2.4M  & 87.46  & 92.05 & 81.18 & 91.77 & 94.25 & 87.46 & 88.38 & 88.34  \\
    Lenna \cite{wei2023lenna}  &   arXiv'23     & G-DINO-L / LLaVA-7B\cite{liu2024llava} & L-Inst,G-DINO-L$^*$,...  & $>$21M  & 90.28  & 93.22 & 86.97 & 88.08 & 90.07 & 83.99 & 90.30 & 90.29  \\
    u-LLaVA \cite{xu2023u-llava}  &   arXiv'23     & CLIP-L / Vicuna-7B\cite{chiang2023vicuna} & L-Inst,RefC,COCO,...  & $\sim$4M  & 91.20  & 94.29 & 87.22 & 85.48 & 91.76 & 78.11 & 86.54 & 87.25  \\
    CogVLM-17B \cite{wang2023cogvlm}  &   arXiv'24    & EVA-2\cite{sun2023eva-clip}/Vicuna-7B\cite{chiang2023vicuna} & LION-2B,COYO,... &  $\sim$3B   & 92.76 & 94.75 & 88.99 & 88.68 & 92.91 & 83.39 & 89.75 & 90.79  \\
    Sphinx-2k \cite{wang2023cogvlm}  &   arXiv'24    & EVA-G\cite{sun2023eva-clip}/LLaMA2\cite{touvron2023llama2} & LION-2B,VG,RefC,... &  $\sim$3B   & 91.10 & 92.88 & 87.07 & 85.51 & 90.62 & 80.45 & 88.07 & 88.65  \\
    VistaLLM \cite{pramanick2024vistallm}  &   CVPR'24     & EVA-G\cite{fang2023eva}/Vicuna-7B\cite{chiang2023vicuna} & L-Inst,CoinLt,\etc &  $\sim$4M   & 88.10 & 91.50 & 83.00 & 82.90 & 89.80 & 74.80 & 83.60 & 84.40  \\
    LION-4B \cite{chen2024lion}  &   CVPR'24     & EVA-G\cite{fang2023eva}/FlanT5-3B & VG,COCO,\etc & 3.6M    & 89.73 & 92.29 & 84.82 & 83.60 & 88.72 & 77.34 & 85.69 & 85.63  \\
    LION-12B \cite{chen2024lion} &   CVPR'24     & EVA-G\cite{fang2023eva}/FlanT5-11B & VG,COCO,\etc & 3.6M   & 89.80 & 93.02 & 85.57 & 83.95 & 89.22 & 78.06 & 85.52 & 85.74  \\
    Ferret-v2-7B \cite{zhang2024ferret-v2}  &   COLM'24     & CLIP-L,DINOv2/Vicuna-7B  & GRIT,VQA,OCR,\etc  & unknown   & 92.79 & 94.68 & 88.69 & 87.35 & 92.65 & 79.30 & 89.42 & 89.27   \\
    Ferret-v2-13B \cite{zhang2024ferret-v2}  &   COLM'24     & CLIP-L,DINOv2/Vicuna-13B & GRIT,VQA,OCR,\etc & unknown   & 92.64 & 94.95 & 88.86 & 87.39 & 92.05 & 81.36 & 89.43 & 89.99  \\
    Ferret-v2-13B \cite{zhang2024ferret-v2}  &   COLM'24     & CLIP-L,DINOv2/Vicuna-13B & GRIT,VQA,OCR,\etc & unknown   & 92.64 & 94.95 & 88.86 & 87.39 & 92.05 & 81.36 & 89.43 & 89.99  \\
    DeepSeek-VL2 \cite{zhang2024ferret-v2}  &   arXiv'25     & SigLIP-SO400M \cite{zhai2023siglip} \etc &  OCR,...\etc & unknown   & 95.10 & 96.70 & 92.70 & 91.20 & 94.90 & 87.40 & 92.80 & 92.90  \\
    \bottomrule
    \multicolumn{13}{p{20.85 cm}}{\rule{0pt}{9pt}{\small Annotation: As described in Sec. \textcolor{blue}{3.1.3} of the main text, we divide these methods into four subdivision settings for a relatively fair comparison. In the type of (c), `$\dagger$' indicates the intermediate pre-training based on detection supervision, `$\ast$' indicates the intermediate pre-training based on grounding supervision, `$\ddagger$' indicates the intermediate pre-training based on multi-task supervision with box-level datasets. `...' represents additional datasets that cannot be exhaustively listed due to space limitations. `RefC' represents the mixup of RefCOCO/+/g training data. `G-DINO-L$^*$' denotes `O365,OI,GoldG,Cap4M,COCO,RefC'. Specifically, `GoldG' (proposed in MDETR \cite{kamath2021mdetr}) is a mixed region-level fine-grained dataset created by combining three datasets (Flickr30k \cite{plummer2015flickr30k}, MS COCO \cite{mscoco}, and Visual Genome \cite{krishna2017visualgenome}), along with annotated text data for detection, REC and QGA tasks. It has a size of approximately $6.2$M. `O365' refers to the Object365 \cite{zhou2019objects} dataset, `SBU' stands for SBU caption \cite{ordonez2011sbu}, `VG' here represents Visual Genome \cite{krishna2017visualgenome} dataset, and `OI' stands for OpenImage \cite{kuznetsova2020openimage} dataset. Besides, `ALBEF*' stands for the pre-training dataset used in ALBEF \cite{li2021albef}, which mainly consists of MS COCO \cite{mscoco}, VG \cite{krishna2017visualgenome}, CC3M\cite{sharma2018cc3m}, CC12M\cite{changpinyo2021cc12m}, SBU\cite{ordonez2011sbu}, WebVid-2M\cite{bain2021webvid}, WikiCorpus \cite{devlin2019bert}, \etc. Furthermore, `L-Inst' stands for LLaVA-instruction tuning \cite{liu2024llava} dataset, `RD' stands for Shikra-RD \cite{chen2023shikra} dataset, `LAION' stands for LAION \cite{schuhmann2021laion} dataset, `COYO' stands for COYO-700M \cite{kakaobrain2022coyo-700m} dataset. Since many methods do not directly disclose the amount of data used for intermediate pre-training, the data pair size in the table may be unavailable or statistically inaccurate. \textbf{\textit{It is strongly recommended that future researchers adopt more rigorous experimental settings to ensure fair comparisons and proactively disclose statistics regarding the amount of data used in intermediate pre-training.}}}}      \\
    \end{tabular}%
}
    \label{tab:rec_sota}%
    \vspace{-13pt}    
\end{table*}%

\renewcommand{\thesection}{A}
\renewcommand{\thesubsection}{A.\arabic{subsection}}

\renewcommand{\thesection}{A\arabic{section}}
\section{Datasets and Benchmarks}
\label{sec:datasets_and_benchmarks}

\begin{table}[t!]
  \setlength\tabcolsep{2pt}
  \renewcommand{\thetable}{A\arabic{table}}
  \centering
  \caption{The detailed statistics of the five classical datasets.}
  \vspace{-8pt}
  \resizebox{1.0\columnwidth}{!}{%
  \begin{tabular}{l|cccc|cccc}
  \toprule
  \multirow{2}{*}{Dataset}  &  \multicolumn{4}{c}{Total}     &  \multicolumn{1}{c}{train}   & val   & test(A)   & testB   \\
                            & images  & instances  & queries  & length  &   queries  & queries & queries & queries \\ \midrule
  RefCOCO \cite{yu2016modeling}                    & 19,994 & 50,000    & 142,210  & 3.49  &  120,624  & 10,834  & 5,657 & 5,095 \\
  RefCOCO+\cite{yu2016modeling}                    & 19,992 & 49,856    & 141,564  & 3.58  &  120,191  & 10,758  & 5,726 & 4,889 \\
  RefCOCOg-u \cite{refcocog-umd}                   & 25,799 & 49,822    & 95,010   & 8.47  &  80,512   & 4,896   & 9,602 & -- \\
  RefCOCOg-g* \cite{refcocog-google}                & 26,711 & 54,822    & 104,560  & 8.46  &  85,474   & 9,536   & --    & -- \\
  ReferItGame\cite{kazemzadeh2014referitgame}      & 20,000 & 19,987    & 120,072  & 3.45  &  54,127   & 5,842   & 60,103 & -- \\
  Flickr30k \cite{plummer2015flickr30k}   & 31,783 & 427,000   & 456,107  & 1.59  &  427,193  & 14,433  & 14,481 & -- \\
  \bottomrule
      \multicolumn{9}{p{10.1cm}}{\rule{0pt}{9pt}{\small Annotation: The statistics are based on the dataset employed in TransVG \cite{deng2021transvg}. `*' indicates that the training set of RefCOCOg-g exists data leakage.}}      \\
  \end{tabular}
}
  \label{tab:dataset_st}
  \vspace{-5pt}
\end{table}


\indent \textbf{\textit{Overview:}} As discussed in Sec. \textcolor{blue}{1} of the main text, datasets have a profound impact on the development of grounding. Therefore, after discussing the methods of the last decade, we would like to introduce both existing classical datasets and newly curated ones to facilitate subsequent research.

\subsection{The Datasets for Classical Visual Grounding}
\label{subsec:dataset_for_classical_grounding}

The datasets for classic visual grounding can generally be categorized into two main groups: small-scale fine-tuning datasets and large-scale region-level pre-training datasets.

\subsubsection{Small-scale Fine-tuning Datasets}
\label{subsubsec:small-scale_fine-tuning_datasets}

Among the existing small-scale fine-tuning datasets, RefCOCO/+/g \cite{yu2016modeling, refcocog-umd}, ReferIt \cite{kazemzadeh2014referitgame}, and Flickr30K \cite{plummer2015flickr30k} are the five most widely used datasets; their statistics are presented in \cref{tab:dataset_st}.


\vspace{3pt}
\noindent\textbf{\textit{(a) ReferItGame.}} As introduced in Sec. \textcolor{blue}{1} of the main text, ReferItGame~\cite{kazemzadeh2014referitgame}, proposed by Tamara \etal in 2014, is the first large-scale real-world expression understanding dataset. It belongs to the phrase grounding task, which contains images from SAIAPR12~\cite{escalante2010segmented} and collects expressions through a two-player game. In this game, the first player is presented with an image and an object annotation and requested to write a textual expression that refers to the object. The second player is then shown the same image along with the written expression and asked to click on the corresponding region of the object. If the clicking is correct, both players receive points and swap roles. If it's incorrect, a new image will be displayed.

\begin{figure}[!t]
	\centering
        \renewcommand{\thefigure}{A\arabic{figure}}
	\includegraphics[width=0.95\linewidth]{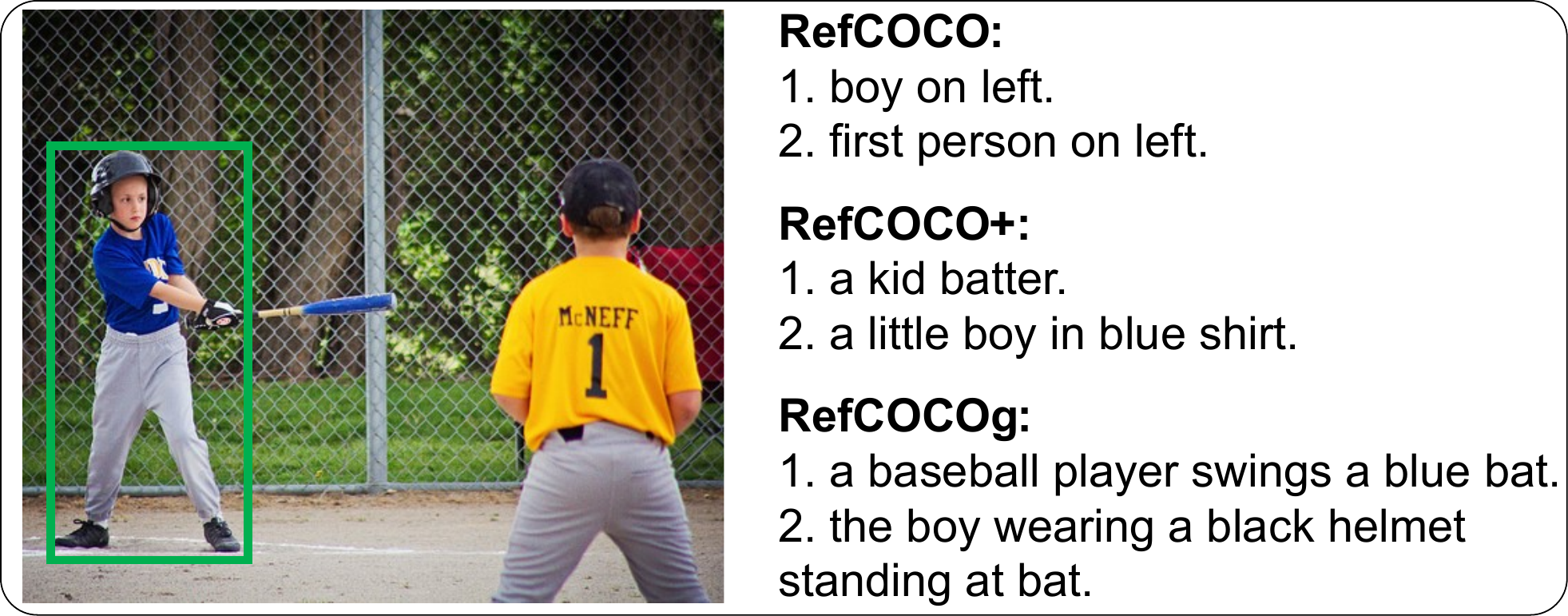}
        \vspace{-7pt}
        \caption{Examples of expression text in RefCOCO/+/g datasets.}
        \vspace{-6pt}
	\label{fig:refcoco/+/g}
\end{figure}

\vspace{3pt}
\noindent\textbf{\textit{(b) Flickr30k Entities.}} The Flickr30k Entities (sometimes short as Flickr30k) dataset \cite{plummer2015flickr30k} was introduced in 2015 by incorporating images from the Flickr30k \cite{young2014image} image dataset. Its primary objective is to construct a fine-grained dataset by establishing region-to-phrase correspondences based on image captions. The subset of data used for phrase grounding has undergone filtration. Nevertheless, the queries within this dataset consist of concise noun phrases extracted from the image captions with a notable amount of noise. As shown in \cref{tab:dataset_st}, the average number of query words is only 1.59. Furthermore, it is common for an image to feature only one object (\eg, a person), and approximately 42$\%$ of the referred regions pertain to individuals and clothing items \cite{plummer2015flickr30k}, thereby eliminating any ambiguity but posing challenges for models to acquire intricate knowledge. Currently, it seems that the fully-supervised work has reached its performance limitations.

\vspace{4pt}
\noindent\textbf{\textit{(c) RefCOCO/+/g.}} The ReferItGame and Flickr30k Entities suffer from a scarcity of instances featuring the same category objects in each image and simplistic and concise text descriptions, which pose a risk of lacking ambiguity \cite{refcocog-google}. Therefore, Mao \etal from Google Inc. proposed the Google Refexp dataset \cite{refcocog-google} in 2016. However, due to \textbf{\textit{data leakage}} in the training set, Nagaraja \etal from the University of Maryland repartitioned the dataset, resulting in a new version known as RefCOCOg-umd \cite{refcocog-umd}, while the original one is usually referred as RefCOCOg-g. In the same year, Yu and Tamara \etal \cite{yu2016modeling}, who also introduced ReferItGame \cite{kazemzadeh2014referitgame} dataset, created RefCOCO and RefCOCO+ datasets based on MS COCO \cite{mscoco} image dataset using a similar two-player game approach. Specifically, there are two test splits called ``testA'' and ``testB'' in RefCOCO/+ \cite{yu2016modeling}. Images in ``testA'' only contain multiple people annotations. In contrast, images in ``testB'' contain all other objects. The RefCOCO dataset does not impose constraints on the usage of location words (\eg, \textit{left, above, middle, first,} \etc) in its expressions, potentially enhancing the model's sensitivity to spatial directions. In contrast, RefCOCO+ restricts the use of such terms, thereby directing attention towards the appearance characteristics of the described object and consequently augmenting textual interest \cite{yu2016modeling}. Additionally, expressions in RefCOCOg~\cite{refcocog-google} are gathered through non-interactive sessions on Amazon Mechanical Turk, resulting in longer and more intricate linguistic structures.

These three datasets (RefCOCO/+/g) laid a solid foundation for grounding research over the past decade. We present examples of ground truth representations from these three datasets in \cref{fig:refcoco/+/g} and provide detailed statistical results summarized in \cref{tab:dataset_st}.

In addition to the datasets mentioned above, several other datasets (\eg, Visual7w \cite{zhu2016visual7w}, GuessWhat?! \cite{de2017guesswhat}, Clevr-ref+ \cite{liu2019clevr-ref+}, Crops-ref \cite{chen2020cops-ref}, Refer360 \cite{cirik2020refer360}, REVERIE \cite{qi2020reverie}, \etc) have received comparatively less attention in the early grounding studies. Due to space limitations, we provide the statistical details of these datasets in \cref{tab:other_ref_datasets}. For more details about these datasets, please refer to the paper \cite{qiao2020referring}.


\subsubsection{Large-scale Region-level Pre-training Datasets}
\label{subsubsec:large-scale_pre_training_datasets}

Recently, to overcome the limitations imposed by category constraints and the limited scale of traditional fine-grained datasets, researchers have developed several new datasets tailored for classic grounding scenarios; their statistics are shown in \cref{tab:other_ref_datasets}.

\noindent \textbf{\textit{(a) Grounding-100M.}} To enhance the model's core open-world detection and grounding capabilities, DINO-X \cite{ren2024dino-x} collected and constructed a large-scale dataset named Grounding-100M. This dataset comprises over 100 million high-quality grounding samples from diverse sources. In this dataset, SAM model \cite{kirillov2023sam} is employed to generate partial pseudo-masks or pseudo-boxes.

\noindent \textbf{\textit{(b) VLM-VG.}} VLM-VG \cite{wang2025learning} is a large-scale grounding dataset created by using generative VLP models to produce regional captions, thereby expanding the scale of visual grounding data.

\begin{table}[t!]\footnotesize
\setlength\tabcolsep{4pt}
\renewcommand{\thetable}{A\arabic{table}}
\caption{The statistics of the other classical grounding datasets.}
\vspace{-13pt}
\begin{center}
\resizebox{1.0\columnwidth}{!}{%
    \begin{tabular}{c|c|c|c|c|c|c}
    \toprule
    \multirow{2}[1]{*}{Datasets}    &  \multirow{2}[1]{*}{Venue}  &   Image   &  Total    &  Total    & Total     & Avg.         \\
                                    &                             &  sources  &  images   &  objects  & queries  & length        \\
    \midrule
    Visual7w \cite{zhu2016visual7w}      &  CVPR'16   & MSCOCO                         & 47,300   & 561,459   & 327,949  &  6.9  \\
    GuessWhat?! \cite{de2017guesswhat}   &  CVPR'17   & MSCOCO                         & 66,537   & 134,073   & 821,899  & 24.9  \\
    TD-SDR \cite{chen2019touchdown}      &  CVPR'19   & Street view                    & 25,575   & --        & 9,326    & 29.8  \\
    Clevr-ref+ \cite{liu2019clevr-ref+}  &  CVPR'19   & CLEVR \cite{johnson2017clevr}  & 99,992   & 492,727   & 998,743  & 22.4  \\
    Crops-ref \cite{chen2020cops-ref}    &  CVPR'20   & COCO/Flickr                    & 75,299   & 1,307,885 & 148,712  & 14.4  \\
    Refer360 \cite{cirik2020refer360}    &  ACL'20    & SUN360                         & 2,000    & 124,880   & 17,137   & 43.8  \\
    REVERIE \cite{qi2020reverie}         &  CVPR'20   & Matterport3D                   & 10,318   & 4,140     & 21,702   & 18.0  \\
    \midrule
    Ground-100M \cite{ren2024dino-x}     & arXiv'24   &  T-Rex2 \cite{jiang2025t-rex2}  &  --     & --       & 100M    & --    \\
    VLM-VG \cite{wang2025learning}       & WACV'25   &  COCO,O365                      & 500K    & 1M       & 16M     & --    \\
    \bottomrule
\end{tabular}%
}
\end{center}
\label{tab:other_ref_datasets}
\vspace{-5pt}	
\end{table}

\begin{table}[t!]\footnotesize
\setlength\tabcolsep{3pt}
\renewcommand{\thetable}{A\arabic{table}}
\caption{The statistics of the datasets for the newly curated universal scenarios.}
\vspace{-13pt}
\begin{center}
\resizebox{1.0\columnwidth}{!}{%
    \begin{tabular}{c|c|c|c|c|c|c}
    \toprule
    \multirow{2}[1]{*}{Datasets}    &  \multirow{2}[1]{*}{Venue}  &   Image   &  Total    &  Total    & Total     & Avg.        \\
                                    &                             &  sources  &  images   &  objects  & queries  & length       \\
    \midrule
    \multicolumn{7}{l}{\textit{\textbf{a. Dataset for Generalized Visual Grounding.}}}    \\
    gRefCOCO \cite{liu2023gres}          & CVPR'23    &  MSCOCO                         & 19,994  & 60,287   & 278,232 & --    \\
    Ref-ZOM \cite{hu2023beyond}          & ICCV'23    &  MSCOCO                         & 55,078  & 74,942   & 90,199  & --    \\
    D$^3$ \cite{xie2024described}        & NIPS'24    &  GRD \cite{wu2023grd_dataset}   & 10,578  & 18,514   & 422     & 6.3   \\
    RefDrone \cite{sun2025refdrone}      & arXiv'25   &  VisDrone\cite{zhu2021detection}  & 8,536   &  63,679   & 17,900   & 9.0    \\
    \midrule
    \multicolumn{7}{l}{\textit{\textbf{b. Representative Datasets and Benchmarks for GMLLMs.}}}     \\
    GRIT$^1$ \cite{peng2024kosmos-2}     & ICLR'24    &  LAION,COYO                     & 90M     & 137M     & 114M    & 4.7    \\
    GRIT$^2$ \cite{you2023ferret}        & ICLR'24    &  VG,O365,...                    & 1.1M    & 678k     & 177k    & --     \\
    HC-RefLoCo \cite{wei2024hc-refloco}  & NIPS'24    &  COCO,O365,...                  & 13,452  & 24,129   & 44,738  & 93.2   \\
    GVC \cite{zhang2024llava-grounding}  & ECCV'24    &  COCO,L-Inst                    & --      & 150K     & --      & --     \\
    Ref-L4 \cite{chen2024ref-l4}         & CVPRW'25   &  RefC,O365                      & 9,735   & 18,653   & 45,341  & 24.2   \\
    \midrule
    \multicolumn{7}{l}{\textit{\textbf{c. Dataset for Other Newly Curated Universal Scenarios.}}}    \\
    GigaGround \cite{ma2024visual}       & CVPR'24    &  PANDA \cite{wang2020panda}     &  3,775  & 61,353   & 61,353  & 14.8  \\
    MC-Bench \cite{xu2024mc-bench}       & arXiv'24   &  COCO,GRD,...                   & 3,345   &  3,202   & 1,514   & 7.2    \\
    \bottomrule
    \multicolumn{7}{p{9.6cm}}{\rule{0pt}{9pt}{\small Annotation: The abbreviations of image sources are the same as in Tab. \textcolor{blue}{A2}.}}    \\
\end{tabular}%
}
\end{center}
\label{tab:new_datasets}
\vspace{-11pt}	
\end{table}



\subsection{Datasets for Generalized Visual Grounding}
\label{subsec:dataset_for_grec}

A GVG dataset needs to support three cases, \ie, (\textit{i}) grounding one target, (\textit{ii}) grounding multi-targets, and (\textit{iii}) grounding no target.

\vspace{3pt}
\noindent  \textbf{\textit{(a) gRefCOCO.}} This dataset is customized by Liu \etal \cite{liu2023gres} for GRES and GREC tasks based on RefCOCO. It contains 278,232 expressions in 19,994 images, of which 800,022 are multi-target, and 32,202 are no-target expressions. In case of grounding multiple targets, some attributes emerge that the RefCOCO dataset does not have, such as \textit{(i)} the use of counting expressions; \textit{(ii)} using a compound sentence structure with ``\textit{without}"; \textit{(iii)} multiple objects are described by different attributes. Besides, the expression uses constrained rules in grounding no-targets (\eg, the expression text can be deceptive but cannot be totally irrelevant to the image). 

\noindent \textbf{\textit{(b) Ref-ZOM.}} It is a GRES dataset constructed by Hu \etal \cite{hu2023beyond} based on COCO, which contains 55,078 images and 90,199 expressions. Among them, there are 56,972 single targets, 21,290 multi-targets, and 11,937 with no targets.

\noindent \textbf{\textit{(c) D$^3$ dataset.}} The D3 (\ie, description detection dataset) \cite{xie2024described} is constructed by Xie \etal using the GRD dataset \cite{wu2023grd_dataset} and based on the COCO protocol. It consists of 10,578 GRD images with 422 expressions, including 316 multi-target expressions and 106 no-target expressions.

\noindent \textbf{\textit{(d) RefDrone.}}  RefDrone \cite{sun2025refdrone} is an REC benchmark designed for drone scenes, which addresses three key challenges: (\textit{i}) scenarios involving multiple targets and no targets; (\textit{ii}) grounding of multi-scale and small-scale objects; (\textit{iii}) reasoning in complex environments with rich contextual.

\subsection{Datasets and Benchmarks for GMLLMs}
\label{subsec:datasets_and_benchmarks_for_gmllms}

GMLLMs differ from traditional VLP-based models in that their training data typically involves multimodal instruction tuning \cite{liu2024llava} and long-text reasoning. With the recent surge in the popularity of GMLLMs, numerous studies have developed their own datasets. Here, we briefly introduce several representative datasets.

\noindent \textbf{\textit{(a) GRIT$^1$.}} GRIT$^1$ (grounded image-text pairs), proposed in KOSMOS-2 \cite{peng2024kosmos-2}, with 114M triplet pairs, is a web-scale dataset built on a subset of image-text pairs from LAION-2B \cite{schuhmann2021laion} and COYO-700M \cite{kakaobrain2022coyo-700m}. It completes the construction by initially generating noun-chunk-bounding-box pairs and then producing referring-expression-bounding-box pairs.

\noindent \textbf{\textit{(b) GRIT$^2$.}} GRIT$^2$ (Ground-and-Refer Instruction-Tuning), proposed in Ferret \cite{you2023ferret}, is a dataset with 1.1M samples, which is based on Visual Genome, Object365, RefC, and other datasets. It is constructed with the help of SAM \cite{kirillov2023sam} and GPT-4’s \cite{achiam2023gpt-4} generation ability. It contains multiple levels of spatial knowledge, including relations, region descriptions, and complex reasoning.

\noindent \textbf{\textit{(c) HC-RefLoCo.}} HC-RefLoCo \cite{wei2024hc-refloco}, as well as HumanRef\cite{jiang2025humanref}, are both the human-centered long-context grounding datasets. The referring text primarily focuses on human-related topics, including appearance, human-object interactions, location, actions, celebrities, OCR, \etc.

\noindent \textbf{\textit{(d) GVC.}} GVC (Grounded Visual Chat) \cite{zhang2024llava-grounding} dataset was proposed in LLAVa-grounding, which is based on COCO image data and LLaVA instruction-tuning data, and uses GPT-4 \cite{achiam2023gpt-4} to generate pseudo labels. The final dataset is obtained through a cycle of self-training and self-calibration.

\noindent \textbf{\textit{(e) Ref-L4.}} The Ref-L4 \cite{chen2024ref-l4} is constructed by integrating the cleaned RefC dataset with the Object365 dataset, serving as a comprehensive REC benchmark for evaluating GMLLMs. Its \textit{``L4"} highlights four critical aspects: a large number of test samples, a large diversity in object categories and instance scales, lengthy referring expressions, and a large vocabulary.

Additionally, several other efforts \cite{yang2025fineCops-ref, jin2025knowdr} have been undertaken to comprehensively evaluate GMLLM. For example, FineCops-Ref\cite{yang2025fineCops-ref} is designed with controllable difficulty levels to accurately assess GMLLM's referring reasoning capabilities across various dimensions, including categories, attributes, and multi-hop relationships.

\subsection{Datasets for Universal Grounding Scenarios}
\label{subsec:dataset_for_universal_grounding}

As a fine-grained cross-modal task, grounding encompasses a broader range of universal application scenarios compared to classic or generalized visual grounding. These include multi-image visual grounding, gigapixel-scale grounding \etc.

\noindent \textbf{\textit{(a) Multi-image visual grounding.}} The goal of multi-image visual grounding is to ground objects referred by text descriptions in multiple images under open-world scenarios. From a universal perspective, multi-image grounding encompasses both single-object grounding and multi-object grounding. Migican \cite{li2025migician} leverages the Chain-of-Thought reasoning capability of MLLMs to extend traditional single-image grounding to the multi-image setting. Additionally, it introduces the MGrounding-630K dataset and establishes the multi-image grounding evaluation benchmark, MIG-Bench. In MC-Bench \cite{xu2024mc-bench}, the authors refer to this task as multi-context visual grounding and construct a set of 2k high-quality, manually annotated samples.

\noindent \textbf{\textit{(b) Gigapixel-scale grounding.}} Giga-Grounding \cite{ma2024visual} is designed to challenge visual grounding in gigapixel-scale scenes. The resolution of the images in this dataset is approximately $25k\times14k$. It mainly deals with high-resolution, large-scale scene understanding and the grounding of multi-hop expressions at both large and small scales of the image.


\renewcommand{\thesection}{A\arabic{section}}
\section{Applications}
\label{sec:application}

\textbf{\textit{Overview:}} As a fundamental cross-modal task, visual grounding not only exhibits strong application values in its own referring comprehension task but also exerts significant influence on other related domains. In addition to tasks such as RES, REG (also known as Grounded Image Captioning \cite{yin2019context}), and Grounded VQA introduced in Sec. \textcolor{blue}{2.4} and Sec. \textcolor{blue}{3.6} of the main text, visual grounding has significantly transformed the research paradigm of numerous tasks by incorporating two modalities of information and exploring cross-modal referring relationships. In this section, we will delve into the broader applications of visual grounding.



\subsection{Grounded Object Detection}
\label{subsec:grounded_object_detection}

Traditional detection tasks \cite{liu2023foregroundness} are constrained by an unimodal design and typically trained on a finite closed set of classes, which essentially involves a classification-based task (\eg, 80 classes in COCO). However, in an open-world scenario, numerous objects can possess unusual labels (\eg, \textit{``syringe"}, \textit{``stingray"}, \etc), and an object may have multiple labels (\eg, \textit{``vaccine"} and \textit{``small vial"} may refer to the same object). Consequently, traditional detection tasks lack competence for such scenes. Grounded Object Detection (GOD) equips unimodal detection tasks with entity region grounding within a multimodality framework. This approach significantly enhances the model's capability for fine-grained semantic alignment during training, enabling the detection model to perceive a broader range of open and diverse objects. In recent studies, numerous grounded detection pre-training approaches (\eg, MDETR \cite{kamath2021mdetr}, GLIP \cite{li2022glip}, GLIPv2 \cite{zhang2022glipv2}, Grounding-DINO \cite{liu2023grounding}, MQ-Det \cite{xu2023mq-det}, \etc) have emerged to address the challenges brought by open-set object detection. The current trend in generalized visual grounding \cite{he2023grec} or described object detection \cite{xie2024described} tends towards amalgamating the detection task with the grounding task.

\subsection{Video Object Grounding}
\label{subsec:video_object_grounding}


\textit{\textbf{Video Object Grounding (VOG)}} \cite{sadhu2020video}, also referred to as \textit{\textbf{Spatio-Temporal Video Grounding}} (\textit{\textbf{STVG}})  \cite{yang2022tubedetr, wasim2024videogrounding, gu2024context}, extends the concept of visual grounding to the video domain. It involves identifying both the temporal segments and spatial locations of objects described by natural language within video sequences. Compared to image-based visual grounding, video object grounding presents additional challenges, including maintaining temporal consistency across frames, handling object occlusions, coping with motion blur, adapting to variations in object appearance, and tracking target objects under partial or full occlusion.

In terms of datasets, STVG primarily encompasses Vid-STG\cite{zhang2020does}, HC-STVG V1 \cite{tang2021human} and HC-STVG V2\cite{tang2021human}. Analogous to the relationship between RES and REC, the counterpart task of STVG is known as \textbf{\textit{Referring Video Object Segmentation (RVOS)}}, and the associated datasets are shared, including Ref-YouTube-VOS \cite{seo2020urvos} and Ref-DAVIS \cite{khoreva2018video}. With the introduction of the concept of generalized visual grounding, Ding et al. proposed MeViS \cite{ding2023mevis}, a dataset that emphasizes the motion attributes of objects and extends beyond the grounding of a single object. Building upon MeViS, MeViSv2 \cite{ding2025mevis} further advances motion reasoning and incorporates no-target expressions. The OmniSTVG \cite{yao2025omnistvg} dataset aims to spatially and temporally localize all objects mentioned in a given textual query within a video.

Existing representative VOG methods can be broadly categorized into three groups: \textit{\textbf{(i)}} Frame-by-frame and online methods. Frame-by-frame approaches (\eg, VOGNet\cite{sadhu2020video}) inherently treat videos as sequences of individual images and perform image-level visual grounding on each frame independently. However, these methods neglect the temporal consistency across frames. To address this limitation, online methods (\eg, OnlineRef \cite{wu2023onlinerefer}) incorporate temporal memory attention and association frameworks to improve temporal coherence. \textit{\textbf{(ii)}} Offline one-stage methods. Although online methods can leverage information from previous frames, they are unable to exploit future frames from a global perspective to constrain the current frame. To overcome this limitation, numerous subsequent approaches (\eg, STVGBert\cite{su2021stvgbert}, TubeDETR\cite{yang2022tubedetr}, \etc) address the VOG problem in an offline manner by considering the entire video sequence. \textit{\textbf{(iii)}} Traditional two-stage methods. Similar to VG, early VOG methods primarily followed a two-stage framework. Some studies (\eg, MAttNet\cite{yu2018mattnet}) extended image-based methods to perform frame-wise grounding and applied post-processing for temporal smoothing. Other approaches typically generate object tracklets across the entire video and then select the one that best aligns with the referring expression. Currently, VOG remains a challenging task due to the complexities of multi-object tracking and ambiguous language references. Consequently, ongoing research continues to explore more effective strategies for spatio-temporal modeling, multimodal fusion, and large-scale pretraining.

It is worth noting that video grounding is a broad concept, aiming to locate relevant video content based on a given textual query. Several related but distinct tasks are often confused with video object grounding and are worth clarifying:  

\textit{\textbf{Video Temporal Grounding (VTG)}}, also known as \textit{\textbf{Video Sentence Grounding (VSG)}}: identifying temporal segments in a video based on a linguistic description.  

\textit{\textbf{Referring Multi-Object Tracking (RMOT)}}: This task involves tracking multiple objects in a video based on textual descriptions referring to specific targets.  

For a more comprehensive understanding of video grounding and its related tasks, readers are encouraged to consult relevant review literature \cite{wu2025survey, zhang2023temporal, ding2025multimodal}.

\subsection{Referring Expression Counting}
\label{subsec:referring_counting}

Traditional counting tasks \cite{guo2024regressor} bear some resemblance to detection tasks as they involve dense objects (\eg, people) counting from unimodal images. However, such indiscriminate counting lacks practicality as it fails to discern the specific information sought by users (\eg, it is more valuable to count \textit{``people in line"} rather than simply count all \textit{``people"}). Consequently, researchers have amalgamated grounding and counting tasks, introducing Referring Expression Counting \cite{dai2024ref-count} or Multimodal Open-world Counting \cite{dai2024ref-count}. To differentiate these tasks from the traditional Referring Expression Comprehension (REC) task, we propose naming them \textbf{\textit{``RefCount"}} (or \textbf{\textit{``GCount"}}). RefCount represents a more challenging and pragmatic task setting compared to conventional unimodal counting, enabling more refined applications.

\subsection{Remote Sensing Visual Grounding}
\label{subsec:remote_sensing_vg}

The Remote Sensing Visual Grounding (RSVG) \cite{ sun2022visual, zhan2023rsvg} is specifically designed for grounding objects referred by a query text on the Remote Sensing (RS) image. RSVG exhibits promising application prospects in various domains, including remote sensing target detection, natural disaster monitoring, agricultural production, search and rescue activities, \etc. Unlike natural scene images, RS images are acquired through satellites and often encounter challenges such as large-scale variations and cluttered backgrounds. Moreover, due to the top-down perspective of RS imagery, the appearance of objects tends to exhibit similar geometric shapes with significant differences from those observed in natural scenes. Consequently, this disparity easily leads to failures in traditional detectors and a domain shift in referring expression texts (\eg, expressions like \textit{``man with red hat"} may not exist while new objects like \textit{``baseball field"} emerge). Given these distinctive characteristics, an effective RSVG model must consider multi-scale information within the image while addressing domain gaps during pre-trained model transfer. Additionally, it is crucial to filter out redundant features to eliminate background clutter effectively. Numerous existing methods \cite{zhan2023rsvg, liu2024rotated, yuan2024rrsis, lan2024language, hang2024regionally, wang2024multi, ding2024visual, li2024language, zhou2024geoground} have proposed to address these challenges, and several RSVG-oriented datasets (\eg, DIOR-RSVG \cite{zhan2023rsvg}, OPT-RSVG \cite{li2024language}, and RefDIOR\cite{lu2025rrsecs}, \etc) have also been proposed.

\subsection{Medical Visual Grounding}
\label{subsec:medical_vg}

Medical Visual Grounding (MVG) \cite{chen2023medical, he2024pfmvg} aims to locate the regions corresponding to medical query phrases in medical images, which is a crucial task in medical image analysis and radiological diagnosis. Similar to RSVG, medical radiology images are typically flat, grayscale, lacking salient object contours, and necessitate specialized knowledge for lesion identification and physiological region recognition. These characteristics significantly differentiate them from natural scene images. Consequently, grounding methods relying on general visual features cannot capture the subtle and specialized attributes required for medical discovery. To address these challenges, researchers constructed the MVG datasets (\eg, MS-CXR \cite{boecking2022making}, ChestX-ray8 \cite{wang2017chestx}, MIMIC-CXR \cite{johnson2019mimic}, \etc) and tailored the common grounding model specifically (\eg, tri-attention context contrastive alignment \cite{chen2023medical}, LLMs \cite{zou2024medrg, luo2024vividmed, yang2023vilam}) for assisting in medical diagnosis \cite{chen2023medical, zou2024medrg, he2024pfmvg}, as well as grounded medical report analysis and generation \cite{luo2024vividmed, yang2023vilam}.

\subsection{3D Visual Grounding}
\label{subsec:3d_visual_grounding}

3D Visual Grounding (3DVG), proposed in 2020 \cite{chen2020scanrefer, achlioptas2020referit3d}, is designed to ground a semantically-specific 3D region corresponding to language queries from three-dimensional (3D) scenes. Unlike those of 2D images, 3D scenes \cite{li2025seeground} are typically represented as intricate and unordered point clouds that capture more comprehensive spatial and depth information. This introduces unique challenges and complexities due to the increased dimensionality and the need for geometric and semantic interpretation. The development of 3DVG is closely related to both 2D VG and 3D detection techniques, with its technical roadmap undergoing a similar transformation from a two-stage process to a one-stage process \cite{liu2024survey}. However, unlike traditional VG methods, 3DVG employs dedicated 3D-based feature extractors, encoders, and bounding box regression heads. For further details on this topic, please refer to paper \cite{liu2024survey}.

\subsection{Speech Referring Expression Comprehension}
\label{subsec:SREC}

Speech Referring Expression Comprehension (SREC) was introduced in 2024, and it aims to ground a target region in an image through spoken language. As an extension of VG, SREC enhances interaction in real-world multimodal systems, especially for voice-based agents, AR/VR applications, and robotic interfaces. Building upon the RefCOCO/+/g dataset, the CSRef \cite{huang2024csref} framework utilized a speech generation tool and constructed three face-centric SREC datasets, named sRefFACE/+/g \cite{huang2024csref}. Additionally, a contrastive semantic alignment module was proposed to align speech and text modalities. This approach enables direct grounding within the speech modality, thereby avoiding information loss that occurs in indirect methods that require converting speech into textual transcripts. Nevertheless, SREC still encounters challenges such as the scarcity of datasets, low speech recognition accuracy, and the difficulty of explicitly aligning semantic representations between speech and visual modalities.

\subsection{Robotic and Multimodal Agent Systems}
\label{subsec:robotic_and_multi-agent}

Referring and grounding are pervasive in various domains. In addition to the applications in detection, counting, remote sensing, medical diagnosis, 3DVG, and video scenarios, a broader utilization of grounding can be observed in robotics and multi-agent systems. For instance, integrating grounding with Visual Language Navigation (VLN) \cite{anderson2018navigation} enables robots to locate targets \cite{liu2023fdo-calibr, xiao2019dynamic} and plan paths efficiently \cite{li2019integrated, jain2023ground}. Moreover, combining grounding with robotic arm manipulation facilitates machine grasping capabilities (\eg, HiFi-CS \cite{bhat2024hifi}). Similarly, Ferret-UI \cite{you2024ferret-ui} and Ferret-UI 2 \cite{zhangheng2025ferret-ui2} leverage MLLMs as agents to navigate user interface screens through flexible input on mobile phones by implementing grounding operations. By integrating grounding abilities into AI's multi-agent systems, it becomes possible to achieve cross-modal referring human-machine dialogue (\eg, Shikra \cite{chen2023shikra}) and effectively enhance the general intelligence of robots.

\subsection{Industrial Applications}
\label{subsec:industrial_applications}

Visual Grounding is increasingly adopted in practical industrial systems due to its ability to interpret free-form language and localize corresponding visual targets. In automated inspection, VG enables defect detection via natural queries like “\textit{cratched surface on the left panel}”, reducing reliance on rigid rule-based systems. In logistics and warehousing, VG assists in locating items based on spoken or written commands (\eg, “\textit{blue container behind the second shelf}”), improving pick-and-place efficiency. Smart surveillance systems integrate VG to detect violations such as “\textit{vehicles parked in no-parking zones}” or “\textit{person crossing outside the crosswalk}”. In equipment maintenance, technicians can use VG-powered AR glasses to highlight components by referring expressions like “\textit{loose valve near the red pipe}”. These applications demonstrate VG’s value in enhancing automation, safety, and human-machine interfaces across diverse industrial domains.

\renewcommand{\thesection}{A\arabic{section}}
\vspace{-2pt}
\section{Advanced Topics}
\label{sec:advanced_topics}
\vspace{-2pt}

\textbf{\textit{Overview:}} Several commonly encountered techniques for visual grounding are independent of specific experimental settings and frequently employed in various scenarios. For this reason, we will discuss some of the representative topics individually in this chapter. In the following contexts, we will briefly introduce NLP language parsers  (\cref{subsec:nlp_parser}), spatial relations and graph neural networks (\cref{subsec:graph_neural_network}), and modular grounding (\cref{subsec:modular_grounding}), \etc.


\begin{figure}[t!]
	\centering
	\includegraphics[width=1.0\linewidth]{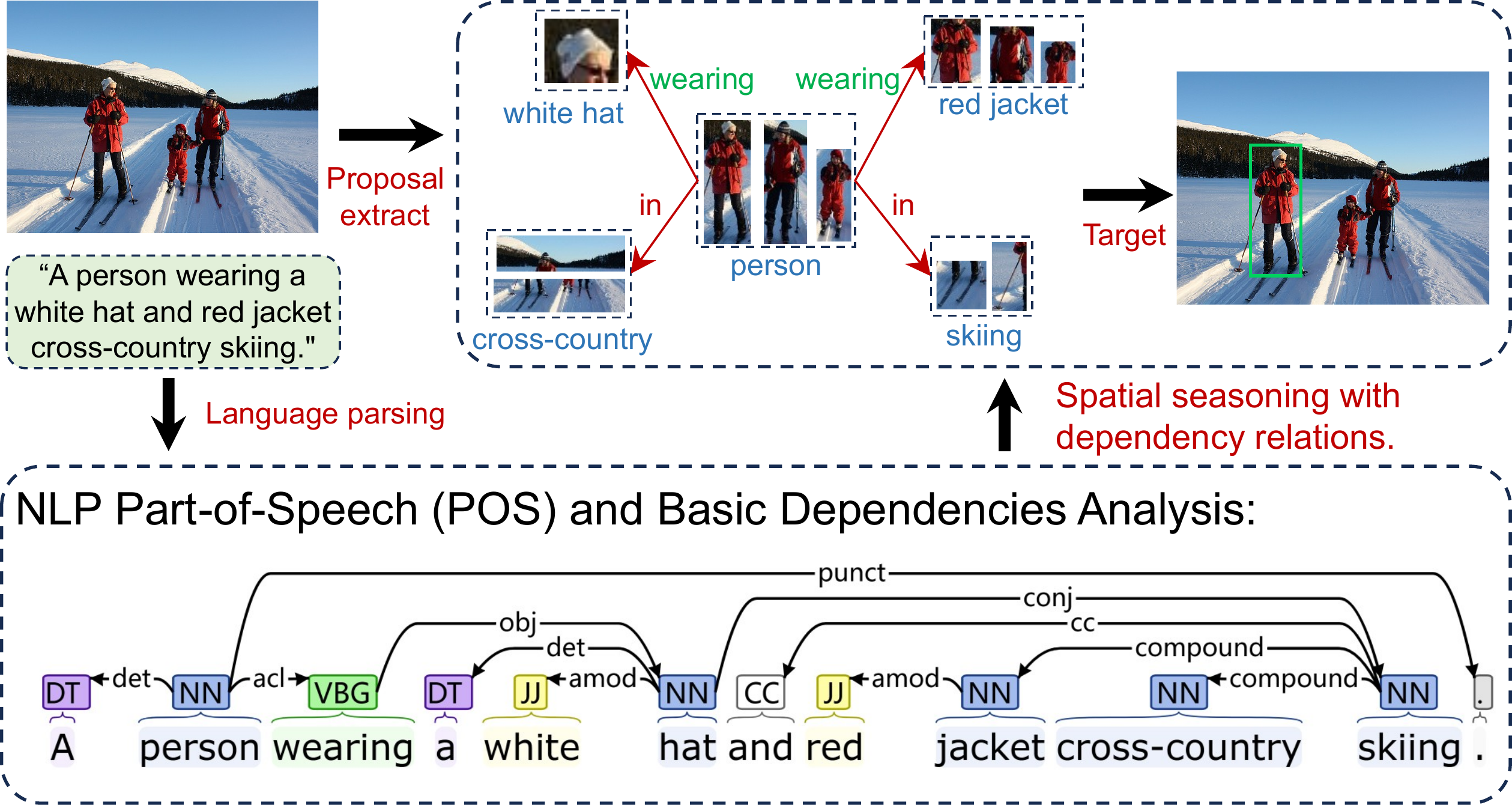}
        \vspace{-16pt}
        \renewcommand{\thefigure}{A\arabic{figure}}
        \caption{Illustration of the language structure parsing. The results are obtained by Stanford CoreNLP \cite{chen2014fast} (\url{https://corenlp.run/}).}
        \vspace{-10pt}
	\label{fig:nlp_parser}
\end{figure}

\vspace{-3pt}
\subsection{Language Structure Parsing in Visual Grounding}
\label{subsec:nlp_parser}

As depicted in \cref{fig:nlp_parser}, language sentences inherently contain structured information that can be readily distinguished, such as subject, predicate, object, and attribute words. In the context of visual grounding, the object to be grounded often corresponds to the subject of a textual expression. Therefore, researchers naturally consider leveraging structured prior information from language to aid grounding. Specifically, by parsing sentence structures (as illustrated in \cref{fig:nlp_parser}), one can ascertain the dependency relationships between textual entities. Consequently, the target region can be determined by matching proposals with textual entities and utilizing spatial prior relation knowledge (\eg, \textit{`A is to the left of B'}, indicating that the horizontal coordinate of object A's bounding box should be smaller than that of object B). 

Following this intuitive and straightforward principle, NLP language parsing tools (\eg, SpaCy \cite{honnibal2015spacy, vasiliev2020natural}, Stanford CoreNLP \cite{chen2014fast}, Stanza \cite{qi2020stanza}, NLTK \cite{bird2006nltk}, OpenNLP \cite{ schmitt2019replicable}, Gensim \cite{ srinivasa2018natural}, Keras \cite{socher2013parsing}, \etc) and spatial prior relations (\eg, ReCLIP \cite{subramanian2022reclip}, Pseudo-Q \cite{jiang2022pseudo}) have been extensively employed across various visual grounding scenarios over the past decade. For instance, in fully supervised setting, researchers employ these tools to construct scene graphs for reasoning during grounding processes (\eg, GroundNet \cite{cirik2018groundnet}, NMTree \cite{liu2019learning}, RVGTree \cite{hong2019rvg-tree}, \etc). Besides, modular grounding can be achieved with assistance from language parsing (\eg, CMN \cite{hu2017modeling}). Moreover, NLP parsing tools prove valuable in generating pseudo-language labels (\eg, CLIP-VG \cite{xiao2023clip}, \etc) under unsupervised grounding settings and constructing new datasets (\eg, ARPGrounding \cite{zeng2024investigating}, \etc). Additionally, weakly supervised approaches (\eg, DTWREG \cite{sun2021discriminative}, \etc) and zero-shot approaches (\eg, ReCLIP \cite{subramanian2022reclip}, MMKG \cite{shi2022mmkg}, \etc) more heavily rely on NLP parsing tools due to the limited availability of accurate ground truth boxes.

\subsection{Spatial Relations and Graph Neural Networks}
\label{subsec:graph_neural_network}

The objective of two-stage visual grounding is essentially to determine the referred region of an expression text within multiple proposals by considering attribute and relative relation constraints. The intricate relationships among these proposals can be represented as a scene relation graph. Graph-based methods \cite{chen2024survey, li2023towards} enable the consideration of target-object relationships and mitigate the ambiguity associated with single descriptions. In particular, within graph neural networks, nodes can highlight relevant targets, while edges are utilized to identify relationships present in textual expressions. Over the past decade, numerous studies have emerged employing relation-based or graph-based techniques for visual grounding tasks (\eg, LGRANs \cite{wang2019neighbourhood}, DGA \cite{yang2019dynamic}, CMCC \cite{liu2020cmcc}, CMRIN \cite{yang2019cmrin}, MMKG \cite{shi2022mmkg}, CLIPREC \cite{ke2023cliprec}, and others \cite{zheng2024resvg}). These studies utilize graph neural networks to acquire grounding reasoning within the visual context by exploiting linguistic structures present in expressions. By harnessing graph attention mechanisms to establish associations and capture supporting cues within graphs, these models significantly enhance visualization and interpretability of the reasoning process involved in visual grounding.



\subsection{Modular Grounding}
\label{subsec:modular_grounding}

Neural Modular Networks (NMNs) \cite{andreas2016neural} were initially proposed for VQA tasks, aiming to decompose a question into multiple components and dynamically assemble several sub-networks to compute an answer. In the traditional CNN-LSTM era, grounding methods often employed a single LSTM to encode the entire textual expression, disregarding the distinctions between different information provided in the text. The fundamental concept of modular grounding is to decompose the text into distinct components and match each component with its corresponding visual region through a modular network, enabling one-step reasoning. CMN \cite{hu2017modeling} proposes compositional modular networks, which consist of a grounding module and a relational module that parse expression text into subjects, relations, and objects using soft attention. MAttNet \cite{yu2018mattnet} introduces the modular attention network, which decomposes expression texts into three components: subject appearance, location, and relationship with other objects. It does not rely on additional NLP parsers but automatically parses expressions based on soft attention mechanisms. Subsequently, matching scores are calculated for three vision modules to measure compatibility between objects and expressions. Furthermore, erasing-based training of modular networks \cite{yu2018mattnet} as well as the application in weak supervision \cite{fang2019modularized} are also investigated subsequently. However, the advent of the BERT model's advanced capabilities in effectively perceiving language semantics has gradually diminished the prominence of NMN in mainstream research.

\ifCLASSOPTIONcaptionsoff
  \newpage
\fi



{
\bibliographystyle{IEEEtran}
\bibliography{IEEEabrv,ref_full}
}


\vfill
\end{document}